\documentclass{article}

\usepackage[final]{neurips_data_2021}

\usepackage[utf8]{inputenc} 
\usepackage[T1]{fontenc}    
\usepackage{url}            
\usepackage{booktabs}       
\usepackage{amsfonts}       
\usepackage{nicefrac}       
\usepackage{microtype}      
\usepackage{xcolor}         

\usepackage{amsmath,amsfonts,bm}

\def\eqref#1{equation~\ref{#1}}

\def\1{\bm{1}}

\DeclareMathAlphabet{\mathsfit}{\encodingdefault}{\sfdefault}{m}{sl}
\SetMathAlphabet{\mathsfit}{bold}{\encodingdefault}{\sfdefault}{bx}{n}

\DeclareMathOperator*{\argmax}{arg\,max}
\DeclareMathOperator*{\argmin}{arg\,min}

\usepackage{minitoc}
\usepackage[toc,page,header]{appendix}

\usepackage{amssymb,mathtools}
\usepackage{enumitem}

\usepackage{graphicx}
\usepackage{subfigure}

\usepackage{algorithm}
\usepackage{amssymb}
\usepackage{mathtools}
\usepackage{cancel}
\usepackage{booktabs}
\usepackage{comment}
\usepackage{tikz}
\usepackage{asypictureB}
\usepackage{bm}
\usepackage{caption}
\usepackage{wrapfig}
\usepackage[utf8]{inputenc} 
\usepackage[T1]{fontenc}    
\usepackage{url}            
\usepackage{booktabs}       
\usepackage{amsfonts}       
\usepackage{nicefrac}       
\usepackage{microtype}      
\usepackage[utf8]{inputenc} 
\usepackage[T1]{fontenc}    
\usepackage{url}            
\usepackage{amsfonts}       
\usepackage{nicefrac}       
\usepackage{graphicx}
\usepackage{booktabs,multirow} 
\usepackage{amsmath}
\usepackage{makecell}
\usepackage[export]{adjustbox}

\usepackage{verbatim} 

\usetikzlibrary{backgrounds}
\usetikzlibrary{calc}
\usetikzlibrary{fit}
\usetikzlibrary{positioning}
\usepackage{enumitem}
\usepackage[acronym,smallcaps,nowarn]{glossaries}
\usepackage{soul}
\usepackage[font={small}]{caption}
\usepackage{todonotes}
\usepackage{multicol}
\usepackage[noend]{algpseudocode}
\usepackage{float,pbox}
\usepackage{pgfplots}
\pgfplotsset{compat=1.16}
\usepackage{setspace}
\usepackage{todonotes}

\definecolor{mydarkblue}{rgb}{0,0.08,0.45}
\newcommand\blfootnote[1]{%
  \begingroup
  \renewcommand\thefootnote{}\footnote{#1}%
  \addtocounter{footnote}{-1}%
  \endgroup
}

\usepackage[colorlinks=true,
    linkcolor=mydarkblue,
    citecolor=mydarkblue,
    filecolor=mydarkblue,
    urlcolor=mydarkblue]{hyperref}
\usepackage[capitalize,nameinlink]{cleveref}

\definecolor{mygray}{gray}{0.4}

\newcommand{\g}[2]{#1\textsubscript{\textcolor{mygray}{$\pm$#2}}}

\DeclarePairedDelimiter{\norm}{\lVert}{\rVert}

\newcommand{\highlight}[1]{\colorbox{blue!10}{#1}}

\newacronym{SDI}{sdi}{Structural Discovery from Interventions}
\newacronym{NIMs}{nims}{Neural Interventional Models}
\newacronym{NIM}{nim}{Neural Interventional Model}

\title{Systematic Evaluation of Causal Discovery in Visual Model Based Reinforcement Learning}

\author{%
  Nan Rosemary Ke \textsuperscript{*,1,2} 
   Aniket Didolkar\textsuperscript{*,3} 
   Sarthak Mittal \textsuperscript{3} 
   Anirudh Goyal \textsuperscript{3}\\ 
   {\bf Guillaume Lajoie \textsuperscript{3}
   Stefan Bauer \textsuperscript{6}
   Danilo Rezende \textsuperscript{2}}\\
  {\bf
   Yoshua Bengio \textsuperscript{3,$\dagger$}
    Michael Mozer \textsuperscript{5}
   Christopher Pal \textsuperscript{1,4} }\\ 
    }

\begin{document}
\blfootnote{ \textsuperscript{*} Authors contributed equally, \textsuperscript{1} Mila, Polytechnique Montr\'eal, \textsuperscript{2} Deepmind, \textsuperscript{3} Mila, Polytechnique Montr\'eal, \textsuperscript{4} Element AI, \textsuperscript{5} Google AI, \textsuperscript{6}Max Planck Institute for Intelligent Systems, \textsuperscript{$\dagger$} CIFAR Senior Fellow
Corresponding authors: \texttt{rosemary.nan.ke@gmail.com}
}

\maketitle

\begin{abstract}
Inducing causal relationships from observations is a classic problem in machine learning. Most work in causality starts from the premise that the causal variables themselves  are observed. However, for AI agents such as robots trying to make sense of their environment, the only observables are low-level variables like pixels in images. To generalize well, an agent must induce high-level variables, particularly those which are causal or are affected by causal variables. A central goal for AI and causality is thus the joint discovery of abstract representations and causal structure. However, we note that existing environments for studying causal induction are poorly suited for this  objective because they have complicated task-specific causal graphs which are impossible to manipulate parametrically (e.g., number of nodes, sparsity, causal chain length, etc.). In this work, our goal is to facilitate research in learning representations of high-level variables as well as causal structures among them. In order  to systematically probe the ability of methods to identify these variables and structures, we design a suite of benchmarking RL environments. We evaluate various representation learning algorithms from the literature and find that explicitly incorporating structure and modularity in models can help causal induction in model-based reinforcement learning.

\end{abstract}

\section{Introduction}

Deep learning methods have made immense progress on many reinforcement learning (RL) tasks in recent years. However, the performance of these methods still pales in comparison to human abilities in many cases. Contemporary deep reinforcement learning models have a ways to go to achieve robust generalization \citep{nichol2018gotta}, efficient planning over flexible timescales \citep{silver2012compositional}, and long-term credit assignment \citep{osband2019behaviour}.  Model-based methods in RL (MBRL) can potentially mitigate this issue \citep{schrittwieser2019mastering}. These methods observe
sequences of state-action pairs, and from 
these observations are able to learn a self-supervised model of the environment. With a well-trained world model, these algorithms can
then simulate the environment and look ahead to future events to establish better value estimates,
without requiring expensive interactions with the environment \citep{sutton1991dyna}. Model-based methods can thus be
far more sample-efficient than their model-free counterparts when multiple objectives are to be achieved in the same environment. However, for model-based approaches to be successful, the learned models must capture relevant mechanisms that guide the world, i.e., they must discover the right causal variables and structure.  
Indeed, models sensitive to causality have been shown to be robust and easily transferable \citep{bengio2019meta,ke2019learning}. As a result, there has been a recent surge of interest in learning causal models for deep reinforcement learning \citep{de2019causal,dasgupta2019causal,nair2019causal,goyal2019recurrent, goyal2020inductive, rezende2020causally,wang2021alchemy,scholkopf2021toward}. Yet, many challenges remain, and a systematic framework to modulate environment causality structure and evaluate models' capacity to capture it is currently lacking, which motivates this paper.

What limits the use of causal modeling approaches in many AI tasks and realistic RL settings is that most of the current causal learning literature presumes abstract domain representations in which the cause and  effect variables are explicit and given \citep{pearl2009causality}. Methods are needed to automate  the inference and identification of such causal variables (i.e. \textit{causal induction}) from low-level state representations (like images).
Although one solution is manual labeling, it is often impractical and in some cases impossible to manually label all the causal variables.
In some domains,  the causal structure may not be known. Further, critical causal variables may change from one task to another, or from one environment to another. And in unknown environments, one ideally aims for an RL agent that could induce the causal structure of the environment from observations and interventions. 

In this work, we seek to evaluate various model-based approaches  parameterized to exploit structure of environments purposfully designed to modulate causal relations. We find that modular network architectures appear particularly well suited for causal learning. Our conjecture is that causality can provide a useful source of inductive bias to improve the learning of world models.

{\bfseries \itshape Shortcomings of current RL development environments, and a path forward.}  Most existing RL environments are not a good fit for investigating causal induction in MBRL, as they have a single fixed causal graph, lack proper evaluation and have entangled aspects of causal learning. For instance, many tasks have complicated causal structures as well as unobserved confounders. These issues make it difficult to measure progress for causal learning. As we look towards the next great challenges for RL and AI, there is a need to better understand the implications of varying different aspects of the underlying causal graph for various learning procedures.

\begin{figure}
 \vspace{-3\baselineskip}
    \centering
    \includegraphics[scale=0.22]{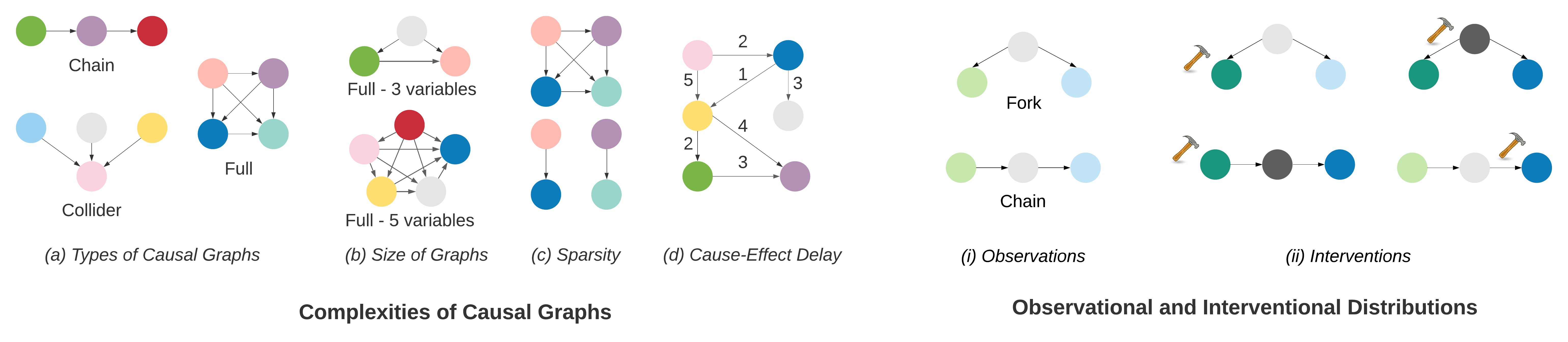}
    \vspace{-0.5\baselineskip}
    \caption{ (a)-(d): Different aspects contributing to the complexity of causal graphs. (i), (ii): Difference between observational and interventional data. In RL setting, actions are interventions in the environment. The hammer denotes an intervention. Intervention on a variable not only affects its direct children, but also all reachable variables. Variables impacted by the intervention have a darker shade.  }
    \label{fig:causal_aspects}
    \vspace*{-1.75\baselineskip}
\end{figure}

Hence, to systematically study various aspects of causal induction (i.e., learning the right causal graph from pixel data), we propose a new suite of environments as a platform for investigating inductive biases, causal representations, and learning algorithms. The goal is to disentangle distinct aspects of causal learning by allowing the user to choose and modulate various properties of the ground truth causal graph, such as the structure and size of the graph, the sparsity of the graph and whether variables are observed or not (see \Cref{fig:causal_aspects} (a)-(d)). 
We also provide evaluation criteria for measuring causal induction in MBRL that we argue help measure progress and facilitate further research in these directions.
We believe that the availability of standard experiments and a platform that can easily be  extended to test different aspects of causal modeling will play a significant role in speeding up progress in MBRL. 

\begin{wrapfigure}{r}{0.55\textwidth}
    \includegraphics[width=.55\textwidth]{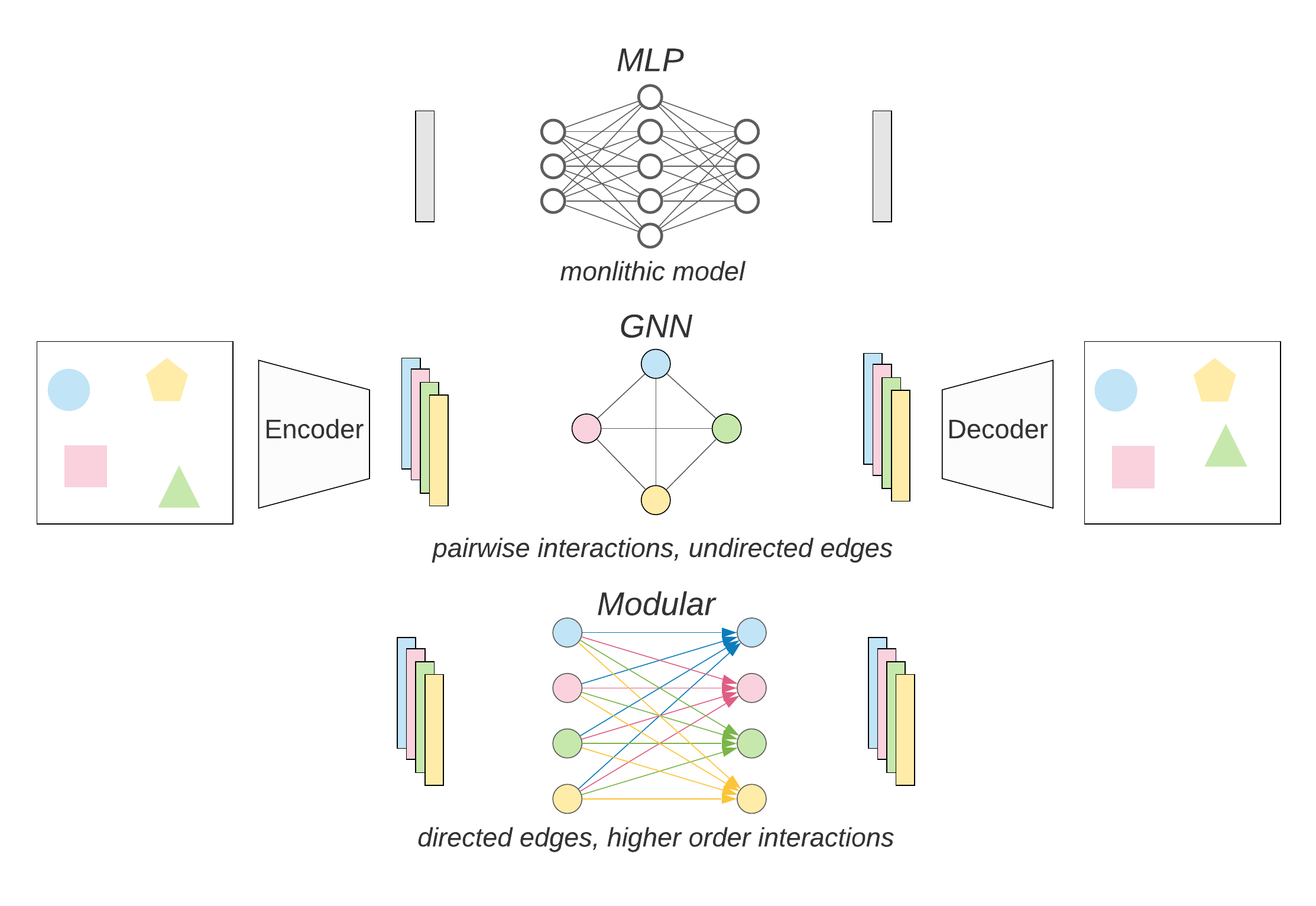}
    \vspace*{-1.5\baselineskip}
    \caption{All models have 3 components: \textit{encoder}, \textit{decoder} and \textit{transition model}. The transition models can either be monolithic, modular model or graph neural networks (GNNs). Monothlic models don't have explicit structure. GNNs have factorized representation of variables. Modular models have factorized representation of both variables and directed edges to potentially model causal relationships, e.g. $A$ causing $B$.}
    \label{fig:various_models}
    \vspace*{-2\baselineskip}
\end{wrapfigure}

{\bfseries \itshape Insights and causally sufficient inductive biases.} Using our platform, we investigate the impact of explicit structure and modularity for causal induction in MBRL. We evaluated two typical of monolithic models (autoencoders and variational autoencoders) and two typical models with explicit structure: graph neural networks (GNNs) and modular models (shown in \Cref{fig:various_models}). Graph  neural networks (GNNs) have a factorized representation of variables and can model undirected relationships between variables.  Modular models also have a factorized representation of variables, along with directed edges between variables which can model directed relationship such as $A$ causing $B$, but not the other way around. We investigated the performance of such
structured approaches on learning from causal graphs with varying complexity, such as the size of the graph, the sparsity of the graph and the length of cause-effect chains (\Cref{fig:causal_aspects} (a) - (d)).

The proposed environment gives novel insights in a number of settings. Especially, we found that even our naive implementation of modular networks can scale significantly better compared to other models (including graph neural networks). This suggests that explicit structure and modularity such as  factorized representations and directed edges between variables help with causal induction in MBRL. We also found that graph neural networks, such as the ones from \citet{kipf2019contrastive} are good at modeling pairwise interactions and significantly outperform monolithic models under this setting. However, they have difficulty modeling complex causal graphs with long cause-effect chains, such as the chain graph (demonstration of chain graphs are found  in \Cref{fig:causal_aspects} (i)). Another finding is that evaluation metrics such as likelihood and ranking loss do not always correspond to the performance of these models in downstream RL tasks.

\vspace{-2mm}
\section{Environments for causal induction in model-based RL}
\vspace{-2mm}

Causal models are frequently described using graphs in which the edges represent causal relationships. In
these \emph{structural causal models}, the existence of a directed edge from $A$ to $B$ indicates 
that intervening on $A$ directly impacts $B$, and the absence of an edge indicates no direct interventional impact (see \Cref{causal_background} for formal definitions).

In parallel, world models in MBRL describe the underlying data generating process of the environment by modeling the next state given the current state-action pair, where the actions are interventions in the environment. 
Hence, learning world models in MBRL can be seen as a causal induction problem. Below, we first outline how a collection of simple causal structures can capture real-world MBRL cases, and we propose a set of elemental environments to express them for training. Second, we describe precise ways to evaluate models in these environments.

\subsection{Mini-environments: explicit cases for causal modulation in RL}
\begin{figure}
\vspace{-3\baselineskip}
    \centering
    \includegraphics[scale=0.3]{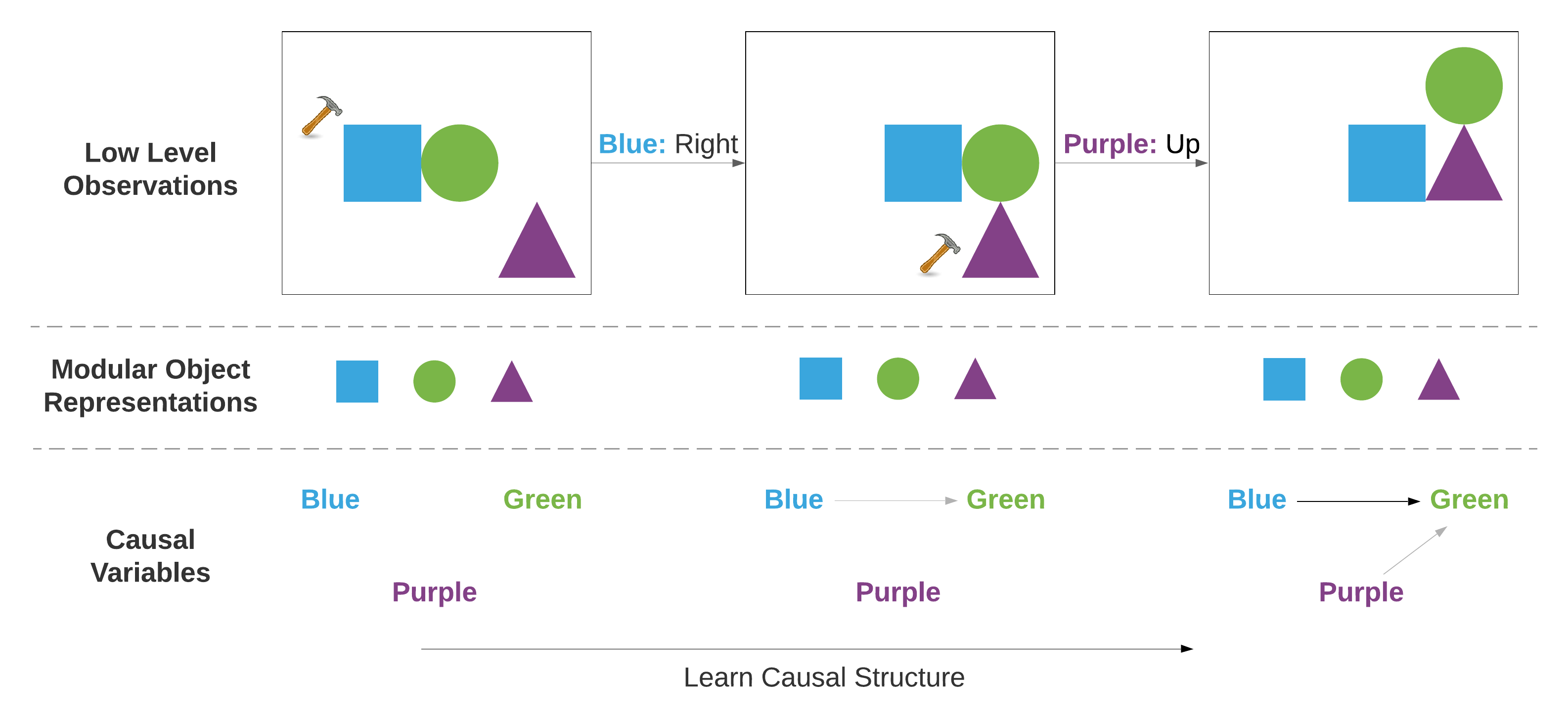}
    \vspace{-0.5\baselineskip}
    \caption{Illustration of the key features of the suite. Environments have objects that interact according to the underlying causal graph which can be based on a subset of objects' properties. An efficient model should be able to infer the high level causal variables from raw pixel data and learn the underlying causal graph through interactions between these high level causal variables.}
    \label{fig:main}
    \vspace*{-1.5\baselineskip}
\end{figure}
\vspace{-2mm}

The ease with which an agent learns a task greatly depends on the structure of the environment's underlying causal graph.
For example, it might be easier to learn causal relationships in a collider graph ( see \Cref{fig:causal_aspects}(a))  where  all interactions are pairwise, meaning that an intervention on one variable $X_i$ impacts no more than one other variable $X_j$, hence the cause-effect chain has a length of at most 1. However, causal graphs such as full graphs (see \Cref{fig:causal_aspects} (a)) can have  more complex causal interactions, where intervening on one variable impacts can impact up to $n-1$ variables for graphs of size $n$ (see \Cref{fig:causal_aspects}). 
Therefore, one important aspect of understanding a model's performance on causal induction in MBRL is to analyze how well the model performs on causal graphs of varying complexity.

Impotant factors that contribute to the complexity of discovering the causal graph are the \textit{structure}, \textit{size}, \textit{sparsity of edges} and \textit{length of cause-effect} chains of the causal graph (\Cref{fig:causal_aspects}). Presence of \textit{unobserved variables} also adds to the complexity.  The size of the graph increases complexity because  the  number of possible graphs grows super-exponentially with the \textit{size of the graph} \citep{eaton2007exact,peters2016causal,ke2019learning}. The \textit{sparsity of graphs} also impacts the difficulty of learning, as observed in \citep{ke2019learning}. Given graphs of the same size, denser graphs are often more challenging to learn. Futhermore,  the \textit{length of the cause-effect} chains can also impact learning. We have observed in our experiments, that graphs with shorter cause-effect lengths such as  colliders (\Cref{fig:causal_aspects} (a)) can be easier to model as compared to chain graphs with longer cause-effect chains. Finally,  \textit{unobserved variables} which commonly exist in the real-world can greatly impact learning, especially if they are confounding causes (shared causes of observed variables). 

Taking these factors into account, we designed two suites of (toy) environments:
 the \underline{\textit{physics environment}} and the \underline{\textit{chemistry environment}}, which we discuss in more detail in the following section. They are designed with a focus on the underlying causal graph and thus have a minimalist design that is easy
 to visualize. 
 \vspace{-1mm}
 \subsubsection{Physics environment: Weighted-block pushing}
\begin{figure}
    \centering
    \includegraphics[scale=0.25]{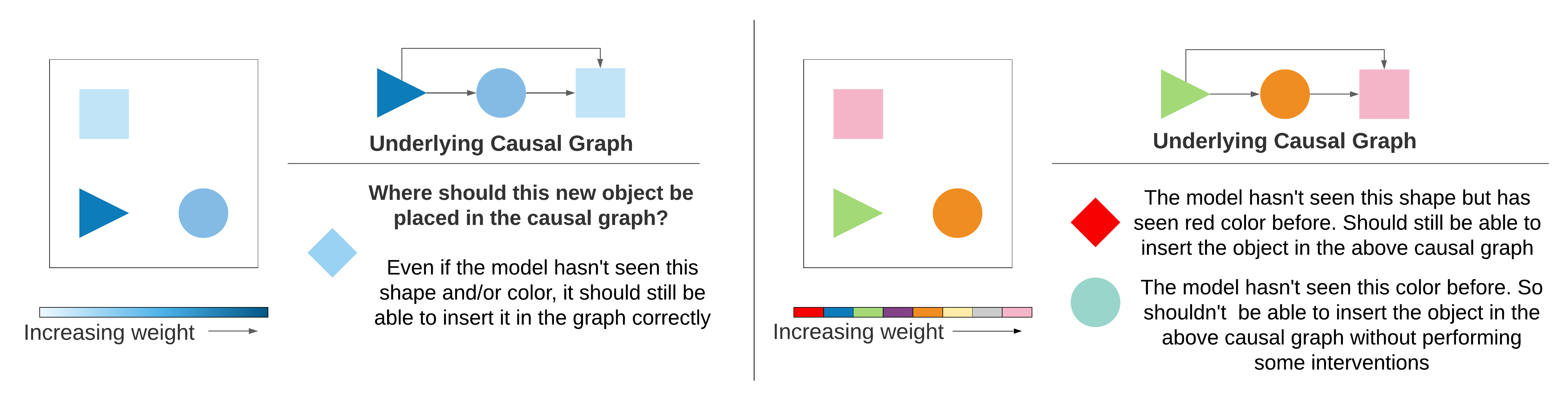}
    \vspace*{-1\baselineskip}
    \caption{Demonstration of the weighted-block pushing environment (left: observed, right: unobserved) along with the feasible generalizations that the setup provides.}
    \label{fig:weighted_blocks}
    \vspace*{-1.5\baselineskip}
\end{figure}

The physics environment simulates very simple physics in the world. It consists of blocks of different, unique weights. The rule for interaction between blocks is that heavier objects can push lighter ones. Interventions ammount to move a particular block, and the consequence depends on whether the block next to it (if present) is heavier or lighter.  For an accurate world model, inferring the weights becomes essential. Additionally, one can allow the weight of the objects to be either observed through the intensity of the color, or unobserved, leading to two environment settings described below. The underlying causal graph is an acyclic tournament, shown in \Cref{fig:weighted_blocks}. 
For more details about the setup, please refer to \Cref{appendix:physics}.

\textit{Fully observed setting}. In the fully observed setting, all objects are given a particular color and the weight of each block is represented by the intensity of the color. 
Once the agent learns this underlying causal structure, it does not have to perform interventions on new objects in order to infer they will interact with the others.

\textit{Unobserved setting}. In this setting, the weight of each object is not directly observable by its color.
The agent thus needs to interact with the object in order to understand the order of weights associated with the blocks. In this case, the weight of objects needs to be inferred through interventions. We consider two sub-divisions of this setting -  \textit{FixedUnobserved} where there is a fixed assignment between the shapes of the objects and  their weights and \textit{Unobserved} where there is no fixed assignment between the shape and the weight, hence making it a more challenging environment.  We refer the reader to \Cref{appendix:physics_complexity} for details.
\vspace{-1mm}
\subsubsection{Chemistry environment}

\begin{wrapfigure}{r}{0.55\textwidth}
    \vspace{-2.0\baselineskip}
    \includegraphics[width=.55\textwidth]{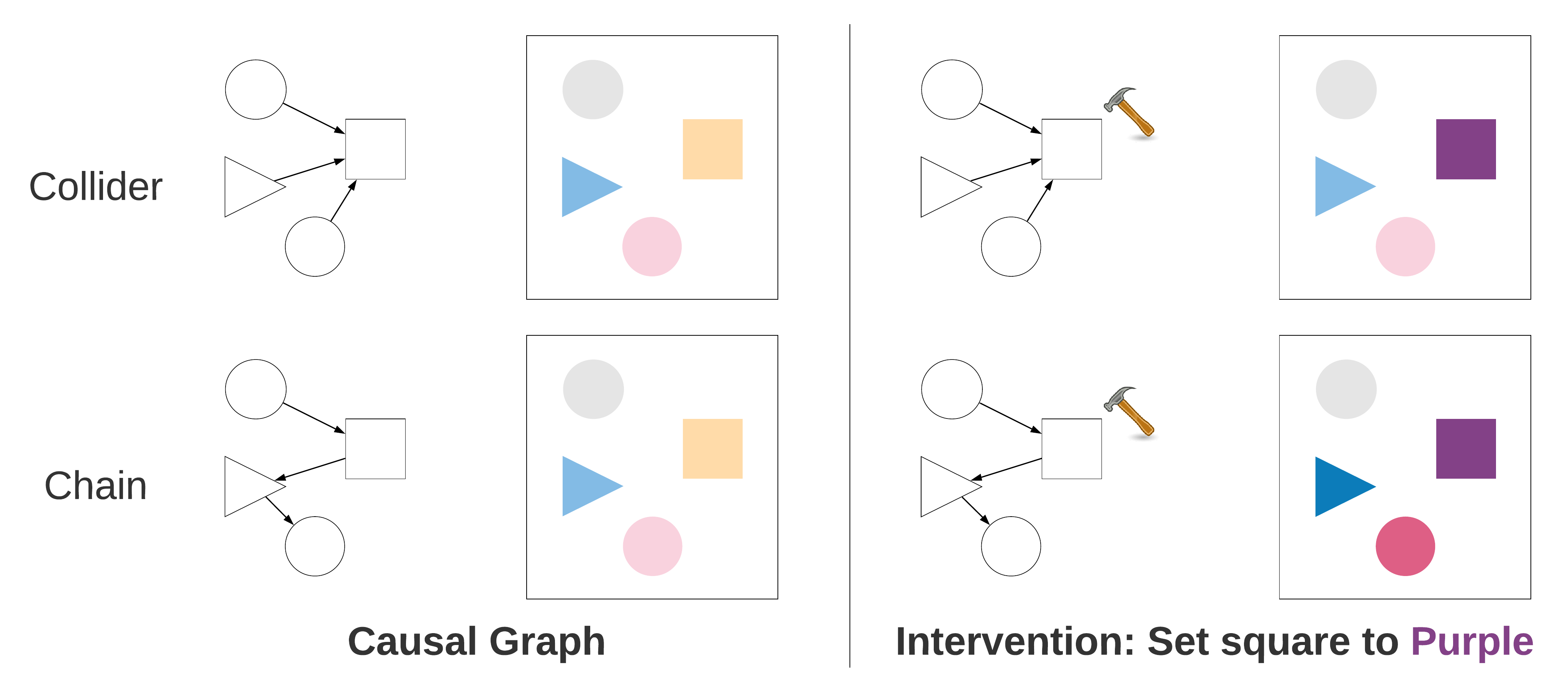}
    \caption{Demonstration of the vanilla chemistry environment (left: ground truth causal graph and a sample from it - same sample shown to demonstrate the affect of interventions, right: the affect of interventions and how far they affect based on underlying causal graph)}
    \label{fig:chem_env_demo}
    \vspace{-1\baselineskip}
\end{wrapfigure}

The chemistry environment enables more complexity in the causal structure of the world by allowing arbitrary causal graphs. This is depicted by simple chemical reactions, where the state of an element can cause changes to another variable's state. 
The environment consists of a number of objects whose positions are kept fixed and thus, uniquely identifiable.

The interactions between different objects take place according to the underlying causal graph which can either be a randomly generated DAG, or specified by the user. An interaction  consists of changing the color (state) of a variable. At this point, the color of all variables affected by this variable (according to the causal graph) can change. Interventions change a block's color
unconditionally, thus cutting the graph edge linking it with its parents in the
graph. All transitions are probabilistic and defined by conditional probability tables (CPTs). A visualization of the environment can be found in \Cref{fig:chem_env_demo}.

This environment allows for a complete and thorough testing of causal models as there are various degrees of complexities which can be easily tuned such as: (1) Complexity of the graph: We can test any model on many different graphs thus ensuring that a models performance is not only limited to a few select graphs. (2) Stochasticity: By tuning the skewness of the probability distribution of each object we can test how good is a given model in modelling data uncertainty. In addition to this we can also tune the number of object or the number of colors to test whether the model generalizes to larger graphs and more colors. A causally correct model should be able to infer the causal relationships between observed objects, as well as their respective color distribution and its dependence on a causal parent's distribution.
\vspace{-1mm}
\subsection{Evaluating  causal models}\label{sec:evaluation}
In much of the existing literature, evaluation of learned causal models is based on the structural difference between the learned graph and the ground-truth graph \citep{peters2016causal,zheng2018dags}. However, this may not be applicable for most deep RL algorithms, as they do not necessarily learn an explicit causal structure \citep{dasgupta2019causal,ke2020amortized}. 
Even if a structure is learned, it may not be unique as several variable permutations can be equivalent, introducing an additional evaluation burden.

Another possibility is to exhaustively evaluate models on all possible intervention predictions and all environment states, a process that quickly becomes intractable even for small environments.
 
We therefore propose a few evaluation methods that can be used as a surrogate metrics to measure the model's performance on recovering the correct causal structure.

\textit{Predicting Intervention Outcomes.}
While it may not be feasible to predict all intervention outcomes in an RL environment, we propose that evaluating predictions on a subset of interventions provides an informative evaluation.
Here, the test data is collected from the same environment used in training, ensuring a single underlying causal graph. Test data is generated from new episodes that are unseen during training. All interventions (actions) in the test episodes are randomly sampled and we evaluate the model's performance on this test set.

\textit{Zero Shot Transfer.} Here, we test the model's ability to generalize to unseen test environments, where the environment does not have exactly the same causal graph as training, but training and test causal graphs share some similarity.

For example, in the {\it observed} Physics environment, a model that has learned the underlying causal relationship between color intensity and weight would be able to generalize to new variables with a novel color intensity.

\textit{Downstream RL Tasks.} Downstream RL tasks that require a good understanding of the underlying causal graph of the environment are also good metrics for measuring the model's performance. For example, in the \textit{physics environment}, we can provide the model with a target configuration in the form of some specific arrangement of blocks on a grid and the model needs to perform actions in the environment to reach the target configuration. Models that capture causal relationships between objects should achieve the target configuration more easily (as it is can predict intervention outcomes). For more details about this setup, please refer to \Cref{appendix:reward_prediction}.

\textit{Metrics.} We also evaluate the learned models on ranking metrics in the latent space as well as reconstruction-based metrics in the observation space \citep{kipf2019contrastive}. In particular we measure and report Hits at Rank 1 (H@1), Mean Reciprocal Rank (MRR) and Reconstruction loss for evaluation in standard as well as transfer testing settings. We report these metrics for 1, 5 and 10 steps of prediction in the latent space (refer \Cref{ranking_metrics}).
\vspace{-2mm}
\section{Models}\label{models}
\vspace{-2mm}
A large variety of neural network models have been proposed as world models in MBRL. These models can roughly be divided into two categories: \textit{monolithic models} and models that have \textit{structure} and \textit{modularity}. \textit{Monolithic models} typically have no explicit structure (other than layers). Some typical monolithic models are Autoencoders and Variational Autoencoders \citep{kingma2013auto,rezende2014stochastic}. Conversely, \textit{structured} models have explicit architecture built into (or learned by) the model. Examples of such models are ones based on graph neural networks \citep{battaglia2016interaction, van2018relational, kipf2019contrastive,veerapaneni2020entity} and modular models \citep{ke2020amortized, goyal2019recurrent,mittal2020learning, goyal2020object}. We picked some commonly used models from these categories and evaluated their performance to understand their ability for causal induction in MBRL.

To disentangle the architectural biases and effects of different training methodologies, we trained all the models on both likelihood based and contrastive losses, respectively. All models share three common components: \textit{encoder}, \textit{decoder} and \textit{transition model}. We follow a similar training procedure as in \citet{ha2018world,kipf2019contrastive}.
Details of the architectures as well as the training protocols and losses can be found in \Cref{appendix:model_details}.

\vspace{-2mm}
\subsection{Monolithic Models}
We evaluate causal induction on two commonly used monolithic models: multilayered autoencoders and variational autoencoders. We follow a similar setup as in \citet{ha2018world}. 
These models do not have strong inductive biases other than the number of layers used. 
\vspace{-2mm}
\subsection{Modular and Structured Models}
Several forms of structure can be included in neural networks, including \textit{modularity}, \textit{factorized variables}, and \textit{directed rules}. 

Taking the three factors into account, we consider two types of structured models in our paper, \textit{graph neural networks} (GNN) and so called \textit{modular networks}. 
Graph neural networks (GNN) \citep{gilmer2017neural, tacchetti2018relational, battaglia2018relational, kipf2019contrastive} is a widely adopted relational model that have a factorized representation of variables and  models  pairwise interactions between objects while being permutation invariant. In particular, we consider the C-SWM model \citep{kipf2019contrastive}, which is a state-of-art GNN used for modeling object interactions. Similar to most GNNs, the C-SWM model learns factorized representations of different objects but for modelling dynamics it considers all possible pairwise interactions, and hence the transition model is monolithic (i.e., not a modular transition model).

Modular networks on the other hand are composed of an initial encoder that factorizes inputs (images), and then a \textit{modular transition model} (MTM) - $M$. This internal model is tasked to create separate factored representations for each objects in the environment, while taking into account all other objects' representations. This model also learns interactions between objects. The rules learned here are \textit{directed rules}. 
\vspace{-2mm}
\section{Experiments}
Our experiments seak to answer the following questions: (a) Does  explicit structure and modularity help for causal induction in MBRL? If so, then what type of structures provide good inductive bias for causal induction in MBRL? (b) How do  different objective functions (likelihood or contrastive) impact learning? 
(c) How do different models scale to complex causal graphs? 
(d) Do prediction metrics (likelihood and ranking metrics) correspond to better downstream RL performance? (e) What are good evaluation criteria for causal induction in MBRL? 

We report the performance of our models on both the Physics and the Chemistry environments, and refer the readers to \Cref{appendix:model_details} for implementation details.. All models are trained using the procedure described in \Cref{appendix:training_procedure} and are evaluated based on \textit{ranking} and \textit{likelihood metrics} on $1,5$ and $10$ step predictions. For the Chemistry environment, we evaluate the models on causal graphs with varying complexity, namely - \textit{chain}, \textit{collider} and \textit{full} graphs. These graphs vary in \textit{the sparsity of edges} and the \textit{length of cause-effect chains}. For the Physics environment, we evaluate the model in the fully observed setting as well as the unobserved setting.
\begin{figure}
\vspace{-1\baselineskip}
    \centering
    \includegraphics[width=15cm]{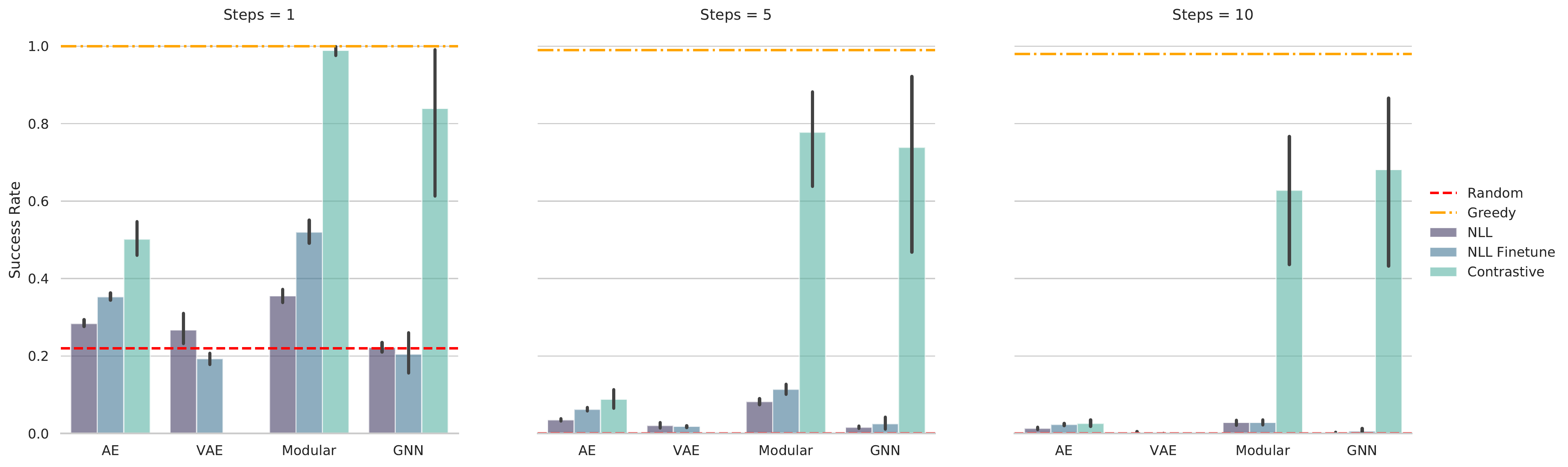}
\caption{Success Rate \textit{(higher is better)} for different models and training losses for 1, 5 and 10 step prediction for the Fixed Unobserved Physics environment setting with 5 objects. Here, (a) Random stands for a random policy, (b) greedy is the policy with best greedy actions, (c) NLL are models trained in 2 stages: pretraining the encoder/ decoder, following by only training the transition model, (d) NLL with finetune are models in 3 stages: pretraining the encoder/ decoder, following by only training the transition model and then finetuning the encoder, decoder and transition models together. (e) Contrastive are models trained using a contrastive loss.  The GNN and Modular models trained on constrastive loss significantly outperform the monolithic models (autoencoders and VAE). The margin significantly increases as the number of steps to reach the goal increase, suggesting that models with explicit structure and modularity have a much better understanding of the world.}
\label{fig:physics_fixedunobserved_3obj_rl_main}
\vspace*{-1.5\baselineskip}
\end{figure}
\vspace{-2mm}
\subsection{Explicit structure and causal induction}

We found that for both the Physics and the Chemistry environments, models with explicit structure outperform monolithic models on both prediction metrics and downstream RL performances. In particular, models with explicit structure (GNNs and modular models) scale better to graphs of \textit{larger size} and \textit{longer cause-effect chains}. 

The Physics environment has a complex underlying causal graph (full graph: refer \Cref{fig:causal_aspects} (a)). We found that GNNs performed well in this environment with 3 variables. They achieved good prediction metrics  (\Cref{fig:physics_observed_3obj}) and high RL performance (\Cref{fig:physics_observed_3obj_rl}) even at longer timescales. However, their performance drops significantly on environments with 5 objects both in terms of prediction metrics (\Cref{fig:physics_observed_5obj}) and RL performance (\Cref{fig:physics_observed_5obj_rl}). We also see in \Cref{fig:physics_observed_5obj,fig:physics_observed_5obj_rl} that modular models scale much better compared to all other models, suggesting that they hold an advantage for \textit{larger} causal graphs. Further, modular models and GNNs when evaluated on zero shot settings outperform monolithic models by a significant margin (\Cref{fig:physics_observed_3obj_0shot,fig:physics_observed_5obj_0shot} and \Cref{tab:physics_observed_3obj_0shot,tab:physics_observed_5obj_0shot}).

For the chemistry environment, we find that modular models outperform all other models for almost all causal graphs in terms of both prediction metrics (\Cref{fig:chemistry_nll_finetune}) and RL performance (\Cref{fig:chem_reward}).  This is especially true on more complex causal graphs, such as \textit{chain} and \textit{full} graphs which have long cause-effect chains. This suggests that modular models scales better to more complex causal graphs.

Overall, these results suggest that structure, and in particular modularity, help causal induction in MBRL when scaling up to larger and more complex causal graphs. The performance comparisons on modular networks and C-SWM \citep{kipf2019contrastive} suggest that both factorized representation of variables and directed edges between variables can help for causal induction in MBRL.

\subsection{Complexity of the Underlying Causal Graph}
There are several ways to vary complexity in a causal graph: \textit{size of the graph}, \textit{sparsity of edges} and \textit{length of cause-effect chain} (\Cref{fig:causal_aspects}). Increasing the size of the graph significantly impacts all models' performances. We evaluate models on the Physics environments with 3 objects (\Cref{fig:physics_observed_3obj}) and 5 objects (\Cref{fig:physics_observed_5obj}) and find that increasing the number of objects from 3 to 5 has a significant impact on performance. Modular models achieve over $90$ on ranking metrics over 10-step prediction for 3 objects while for 5 objects, they achieve only $50$ (almost half the performance on 3 objects). A similar pattern is found in almost all models. Another factor impacting complexity of the graph is the \textit{length of cause-effect chain}.We see that collider graphs are the easiest to learn, with modular models and autoencoders significantly outpeforming all other models (\Cref{fig:chemistry_nll_finetune}). This is because the collider graph has short pair-wise interactions, i.e, intervention on any node in a collider graph can impact at most one other node. Chain and full graphs are significantly more challenging because of longer cause-effect chains. For a chain or a full graph of $n$ nodes, an intervention on the $k^{th}$ node can impact all the subsequent $(n - k)$ nodes. Modeling interventions on chain and full graphs require modeling more than pairwise relationships, hence, making it much more challenging. We find that modular models slightly outperform all other models on these graphs. 

\subsection{Prediction Metrics and RL Performance}
As discussed in \Cref{sec:evaluation}, there are multiple evaluation metrics based on either prediction metrics or RL performance. The performance of the model on one metric may not necessarily transfer to another. 
We would like to analyze if this is the case for the models trained under various environments. We first note that while the ranking metrics were relatively good for most models on physics environments, most of them only did slightly better than a random policy on downstream RL, especially on larger graphs (Figures \Cref{fig:physics_observed_3obj} - \ref{fig:physics_fixedunobserved_5obj} and \Cref{tab:physics_observed_3obj} - \ref{tab:physics_fixedunobserved_5obj} for ranking metrics; \Cref{fig:physics_observed_3obj_rl} - \ref{fig:physics_fixedunobserved_5obj_rl} and \Cref{tab:physics_observed_3obj_rl} - \ref{tab:physics_fixedunobserved_5obj_rl} for downstream RL). 
 \Cref{fig:physics_observed_scatter_3,fig:physics_observed_scatter_5,fig:chemistry_scatter_5} show scatter plots for each pair of losses, with one loss on each axis.
While there is some correlation between ranking metric and RL performance (Modular and GNN; \Cref{fig:physics_observed_scatter_3}), we did not find this trend to be consistent across models and environment settings. We feel that these results give further evidence of need to evaluate on RL performance.

\subsection{Training objectives and learning}
Likelihood loss and contrastive loss \citep{oord2018representation,kipf2019contrastive} are two frequently used objectives for training world models in MBRL. We trained the models under each of these objective functions to understand how they impact learning. In almost all cases, models with explicit structure (modular models and GNNs) trained on contrastive loss perform better in terms of ranking loss compared to those trained on likelihood loss (refer to \Cref{fig:physics_observed_3obj} - \ref{fig:physics_fixedunobserved_5obj}). We don't see a very clear trend between training objective and downstream RL performance but we do see a few cases where contrastively trained models performed much better than others (refer to \Cref{fig:physics_fixedunobserved_3obj_rl_main,fig:physics_observed_3obj_rl,fig:physics_fixedunobserved_3obj_rl,fig:physics_fixedunobserved_5obj_rl} and \Cref{tab:physics_observed_3obj_rl,tab:physics_fixedunobserved_3obj_rl,tab:physics_fixedunobserved_5obj_rl}). 

For other key insights and experimental conclusions on different environments, we refer the readers to \Cref{appendix:physics_insight} for the physics environment and \Cref{appendix:chemistry_insight} for the chemistry environment.

\section{Related work}
\vspace{-2mm}
\textit{Video Prediction and Visual Question Answering.} There exist a number of video prediction \citep{yi2019clevrer,baradel2019cophy} and visual question answering \citep{johnson2017clevr} datasets that also make use of a blocks world for visual representation. Though these datasets can appear visually similar to ours at first glance, they lack two essential ingredients for systematically evaluating models for causal induction in MBRL. The first is that they do not allow active interventions and hence make it challenging for evaluating model-based reinforcement learning algorithms. Another key point is that these environments do not allow one to systematically perturb different aspects of causal graphs, hence, preventing to systematically study the performances of  models for causal induction.

\textit{RL Environments. }    There exist several benchmarks for multi-task learning for robotics (Meta-World \citep{yu2019meta} and RLBench \citep{james2020rlbench}) and for  video gaming domain (Arcade Learning Environment, CoinRun \citep{cobbe2018quantifying}, Sonic Benchmark \citep{machado2018revisiting}, MazeBase \citep{nichol2018gotta} and BabyAI \citep{chevalier2018babyai}). However, as mentioned earlier, these benchmarks do not allow one to systematically control different aspects of causal models (such as the structure, the sparsity of edges and the size of the graph), hence making it difficult to systematically study causal induction in MBRL. The Alchemy \citep{wang2021alchemy} environment, which was released earlier this year, moves a step towards causal induction for meta-RL. Though the environment allows for some level of control of the underlying causal structures of the environment, it still does so in a limited way.  

\textit{Block World.} The AI community has been using the ``blocks world'' for decades as a testbed for various AI problems, including learning theory \citep{winston1970learning},   natural language \citep{winograd1972understanding}, and planning \citep{fahlman1974planning}. Block world allows to easily vary different aspects of the underlying causal structure, and also allow interventions to be performed on many high level variables of the environment giving rise to a large space of tasks which have well-defined relations between them.

\section{Discussions and conclusions}
In our work, we  focus on studying various model-based approaches for causal induction in model-based RL. We  highlighted the  limitations of existing benchmarks and introduced a novel suite of environments   that can help  measure progress and facilitate research in this direction. We evaluated various models under many different settings and discuss the essential problems and challenges in combining both fields i.e ingredients, that we believe are common in the real world, such as modular factorization of the objects and interactions of objects governed by some unknown rules. Using a proposed evaluation framework, we demonstrate that structural inductive biases are beneficial to learning causal relationships and yield significantly improved performances in learning world models. We hope that our work helps to facilitate future work for understanding causal relationships in model-based reinforcement learning.

\section*{Acknowledgements}
The authors would like to acknowledge the support of the following agencies for research funding and computing support: NSERC, Compute Canada, the Canada Research Chairs, CIFAR. We would also like to thank the developers of Pytorch for developments of great frameworks. We would like to thank Dmitriy Serdyuk for useful feedback and discussions.

\bibliographystyle{plainnat}
\bibliography{neurips}
\clearpage

\appendix

\onecolumn

\addcontentsline{toc}{section}{Appendix} 
\part{Appendix}

\section{Dataset Documentation}
We open-source our environment, data, code and instructions on how to run them \footnote{\url{https://github.com/dido1998/CausalMBRL}}. We do not provide the data as a downloadable file, it can be generated using the instructions in the repository. We provide instructions to reproduce the results of our benchmark experiments. The data is provided under the MIT license. We bear all responsibility in-case the dataset leads to any violation of rights. The metadata for the dataset can also be found in the given github repository. 

\textbf{Intended Use}. The intended use of this dataset is for causal learning research in model-based RL. We hope that this dataset can help to speed up discovery of novel methods that can learn causal relations in RL environments. 

\textbf{Reading the Data}. The data is generated and stored in HDF5 format. \footnote{\url{https://www.hdfgroup.org/solutions/hdf5/}}, it can be accessed using the h5py python package \footnote{\url{https://pypi.org/project/h5py/}}. We provide the code for reading the data.

\section{A short review to Structured Causal Models}\label{causal_background}

\paragraph{Causal modeling.}A Structural Causal Model (SCM) \citep{peters2017elements}  over a finite number $M$ of random variables $X_i$ is a function that maps from the jointly-independent noise $N_i$ and  parents (direct causes) $X_{pa(i,C)}$ of $X_i$ to $X_i$. The matrix  $C \in \{0,1\}^{M \times M}$ represents the adjacency matrix (structure) of the graph, such that $c_{ij} = 1$ if node $i$ has node $j$ as a parent (equivalently, $X_j \in X_{pa(i,C)}$; i.e. $X_j$ is a direct cause of $X_i$). 
\begin{align}\label{eq:structuralassignment}
	X_i &:= f_i(X_{pa(i,C)}, N_i)\;, \quad \forall i \in \{0,\ldots,M-1\}
\end{align}
Causal structure discovery is the recovery of ground-truth $C$ from observational and/or interventional studies.

\paragraph{Interventions.} An intervention on a variable $X_i$ changes the function  $f_i$ that maps from the causal parents of $X_i$ and the independent noise ($(X_{pa(i,C)},N_i)$) to $X_i$. There are several common types of interventions available \citep{eaton2007exact}:
\textit{No intervention:} only observational data is obtained from the ground truth model. \textit{Perfect:} the value of a single or several variables is fixed and then ancestral sampling is performed on the other variables. \textit{Imperfect:} the conditional distribution of the variable on which the intervention is performed is changed.  All our experiments are performed with perfect interventions (aka. setting the state of a variable to a particular value, for example location or color), as they are the most common type of interventions in RL.

\section{Ranking based Evaluation}\label{ranking_metrics}
Apart from standard reconstruction loss, we also provide ranking results based on the evaluation metrics followed by \cite{kipf2019contrastive}. Given observations at two different time steps, these metrics capture how close is the predicted transition in the embedding space to the embedding of the true observation obtained through the true environment transitions. Here the notion of closeness is defined as ranking from a large buffer of states under euclidean norm.
\subsection{Hits at Rank 1 (H@1)}
This score is 1 for a particular example if the predicted state representation is nearest to the encoded true observation and 0 otherwise. Thus, it measures whether the rank of the predicted representation is equal to 1 or not, where ranking is done over all reference state representations by distance to the true state representation. We report the average of this score over the test set.
\subsection{Mean Reciprocal Rank (MRR)}
This is defined as the average inverse rank, i.e, MRR $= \frac{1}{N} \sum_{n=1}^N \frac{1}{\text{rank}_n}$ where $\text{rank}_n$ is the rank of the $n^{th}$ sample of the test set where ranking is done over all reference state representations.

\section{Reward Prediction Evaluation}\label{appendix:reward_prediction}
Below, we provide the methodology of training the reward predictor and doing evaluation based on it as well as further implementation details relevant to our particular set of environments.
\subsection{Methodology}
For downstream RL evaluation, we consider learning a reward predictor and then performing planning based on taking greedy actions in the direction of immediate highest reward (inspired from \cite{cobra}). For our tasks, the reward is a function of the next state and the target state but not the action. For example, in physics environment the reward is the average distance between the objects in their current configuration and a target configuration. Similarly, for chemistry environment it is the number of color matches between the current state and the target state.

More concretely, we learn a reward predictor function (parameterized by a single layered MLP) that takes as input the current state as well as the target state of the world and tries to predict the reward for the current state. This reward predictor is learned in a supervised way and all the other weights (encoder, decoder, transition models) are kept fixed during this training. Thus, it is only possible to learn a good reward predictor if the encoder model captures the important aspects of the objects from the raw image.

Given the current encoded state of the world, we consider all possible actions and transitions according to them in the latent space (using the learned transition model). After the transition, we use the learned reward predictor to predict the reward for the (new state, target state) pair. This gives us the immediate reward obtained from each action. Having obtained those rewards, our policy is to just greedily take the action that gives us the best immediate reward. Note that in our reward setting (dense and/or partial rewards) this is typically a good policy as can be seen in the oracle (greedy) performance (where we take actions according to the true reward).

For training, we consider the supervised $L_1$ loss optimized using the Adam Optimizer -
\begin{align*}
    \mathcal{L}_{Reward\;Predictor}(\theta) &= \norm{f_\theta(s_t, s_{target}) - r(x_t, x_{target})}_1 \\
    s_t &= \text{Encoder}(x_t) \\
    s_{target} &= \text{Encoder}(x_{target}) \\
\end{align*}
where $r(\cdot, \cdot)$ is the true reward function. 

For evaluation, we consider the true final reward as well as the success rate obtained under policy $\pi$ where $\pi$ is implicitly defined using the learned reward function $f_\theta$ as follows -
\begin{align*}
    \pi(s_t, s_{target}) &= \argmax_{a\in \mathcal{A}} f_\theta(\text{Transition}(s_t, a), s_{target})
\end{align*}
We leave the formulation of training a value function estimator using a TD-learning objective as an important future work.

\subsection{Implementation Details}\label{appendix:reward_prediction_implementation}
For all the environments, when training a reward predictor we consider a starting state of the environment and the state of the environment obtained after doing 10 random actions. Given the starting state and the target state, we use the dense reward obtained in the configuration to act as the supervision signal for training of the reward predictor model.

For physics environment, we consider the reward to be the average distance of objects from their target configurations. Whereas, for the chemistry environment we consider the number of partial matches between the two states as the reward function.

For evaluation on downstream RL tasks, for $k^{th}$ step prediction, we consider targets that are generated from $k$ random actions in the environment. We also report baseline performances of a random policy as well as an optimal policy. For the physics environment, we set the optimal policy to be the one step greedy policy based on the true reward while for the chemistry environment, we consider the same actions that led to the target configuration to be the optimal policy. Note that since the chemistry environment is stochastic, the same actions may not lead to the same state. Hence any loss in performance even after performing optimal actions is due to the data uncertainty that arises due to the stochasticity. 

\section{Model setups and training procedure}\label{appendix:model_details}
\subsection{Model Based Experiments}
For our model based experiments, we consider four models that encode different inductive biases -
\begin{itemize}
    \item Autoencoders (AE) - Monolithic model that compresses everything into a single entity.
    \item Variational Autoencoders (VAE) - Similar to Autoencoders but with regularization to stay close to a prior distribution in latent space.
    \item Modular Model (Modular) - Has a separate representation for each object and can be used to capture interactions between multiple sets of objects.
    \item Graph Neural Networks (GNN) - Also has an object-wise representation but can capture only pairwise interactions between objects.
\end{itemize}

Each model has an encoder-decoder model as well as a transition model. The encoder-decoder model is aimed at inferring the high level causal variables from raw pixel data whereas the transition model is tasked with controlling how the encoded state transitions based on the actions taken. We build all our models on the architectural backbone provided by~\cite{kipf2019contrastive}.

The encoder model is a convolutional neural network followed by a 3-layered MLP (\Cref{tab:conv_encoder}). It outputs a single representation in case of monolithic models and an object-wise representation (i.e. separate for each object) in case of modular networks and graph neural networks. 

The decoder model (if used - refer \Cref{appendix:training_procedure}) takes either a single representation (in case of monolithic models) or object-wise representations (in case of modular networks / GNNs) and outputs an image as close as possible to the input image. The structure of the decoder is detailed in \Cref{tab:conv_decoder}.

We follow the \textit{medium} encoder-decoder structure followed by \cite{kipf2019contrastive}. For embedding dimension, we use a fixed embedding dimension of 32 per object where the number of objects are specified by the environment description. For example, if we have 3 objects in the environment, then the embedding dimension of Autoencoder based models is 96 while it is 32 per object for Modular/GNN models.

Mathematically, given an observation $x_t$, the encoder maps the observation to its latent representation $s_t$ which is either monolithic or modular. Further, the decoder (if used) maps the latent representation back to the input space.
\begin{align*}
    s_t &= \text{Encoder} (x_t) \\
    \hat{x}_t &= \text{Decoder} (s_t)
\end{align*}

Each architecture also has a transition model to model how a particular action affects the state of the world. Based on the current state of the world and an action taken, the transition model predicts the next state of the world. For monolithic models (AE and VAE), the transition model is a 3-layered MLP. For GNN, it is a graph neural network with only one node-to-edge and one edge-to-node information propagation, that is, it encodes only pairwise interactions. For modular models, it is a separate MLP for each object, that allows it to encode higher order interactions between multiple objects.

Mathematically, the transition (prediction of next state) from a given state $s_t$ based on an action $a_t$ can be shown as -
\begin{align*}
    \hat{s}_{t+1} &= \text{Transition}(s_t, a_t)
\end{align*}

\begin{table}[]
    \centering
    \begin{tabular}{|c|c|c|c|}
    \hline
    Type & channels & activation & stride \\
    \hline
    Conv2D $9 \times 9 $ & 512 & Leaky Relu & 1 \\
    BatchNorm2D & - & - & - \\
    Conv2D $5 \times 5$  & $M$(number of objects) & Sigmoid & 5 \\
    \hline
     
    \end{tabular}
    \caption{Architecture of the encoder used for the world models.}
    \label{tab:conv_encoder}
\end{table}

\begin{table}[]
    \centering
    \begin{tabular}{|c|c|c|c|}
    \hline
    Type & channels & activation & stride \\
    \hline
    Linear & 512 & Relu & - \\
    Linear &  512 & Relu & - \\
    Linear & $M \times 10 \times 10$ & - & - \\
    ConvTranspose2D $5 \times 5$ & 512 & Relu & 5 \\
    BatchNorm2D &- & -& -\\
    ConvTranspose2D $9 \times 9$ & 50 & - & 1 \\
    
    \hline
     
    \end{tabular}
    \caption{Architecture of the decoder used for the world models.}
    \label{tab:conv_decoder}
\end{table}

\subsection{Training Details}\label{appendix:training_procedure}
We consider two methods of training for all our baseline models -
\begin{itemize}
    \item Negative Log Likelihood \textit{(NLL)}
    \item Contrastive Loss \textit{(Decoder Free)}
\end{itemize}

For the models trained using NLL, we perform training in 3 stages. First, we do \textit{pretraining} where only the encoder and decoder are trained to reconstruct the given image. Second, we learn the \textit{transition} where the encoder and decoder are fixed and the transition function is trained to optimally predict the next state given the current state and action. Finally, we do \textit{finetuning} where we train both the encoder-decoder model as well as the transition model on combined objectives of reconstructing the current images, reconstructing the images in next step as well as doing correct transitions in the latent space.

For the reconstructions, we use the binary cross entropy loss (BCE loss) while for the transitions, we use the mean squared error loss (MSE loss). 

Mathematically, given the current observation $x_t$, the action taken $a_t$ and the next observation obtained $x_{t+1}$, we first encode both the observations into the latent space as -
\begin{align*}
    s_t &= \text{Encoder} (x_t) \\
    s_{t+1} &= \text{Encoder} (x_{t+1})
\end{align*}
We then perform a transition from the current step using the transition model as well as use the decoder to perform reconstructions based on the current encoded state as well as the predicted state -
\begin{align*}
    \hat{s}_{t+1} &= \text{Transition} (s_t, a_t) \\
    \hat{x}_t &= \text{Decoder} (s_t) \\
    \hat{x}_{t+1} &= \text{Decoder} (\hat{s}_{t+1})
\end{align*}
Given these variables, the \textit{pretraining}, \textit{transition training} and the \textit{finetuning} can be characterized as -
\begin{align*}
    \textit{Pretraining : }&\argmin_{\text{Encoder}, \text{Decoder}} \text{BCE}(x_t, \hat{x}_t) \\
    \textit{Transition : }&\argmin_{\text{Transition}} \text{MSE}(s_{t+1}, \hat{s}_{t+1}) \\
    \textit{Finetuning : }&\argmin_{\text{Encoder, Decoder, Transition}} \text{BCE}(x_t, \hat{x}_t) + \text{MSE}(s_{t+1}, \hat{s}_{t+1}) + \text{BCE}(x_{t+1}, 
    \hat{x}_{t+1})
\end{align*}

For models trained with contrastive loss, we follow the same setup as in \cite{kipf2019contrastive}. In this setup we don't use a decoder and instead learn everything in encoded state end-to-end. Mathematically, this can be described as the following -
\begin{align*}
    \textit{Contrastive Training }: \argmin_{\text{Encoder, Transition}} &H + \max(0, \gamma - \tilde{H}) \\
    H &= \text{MSE}(\hat{s}_{t+1}, s_{t+1}) \\
    \tilde{H} &= \text{MSE}(\tilde{s}_{t+1}, s_{t+1}) \\
    \tilde{s}_{t+1} &: \text{ Negative state obtained from random shuffling of batch}
\end{align*}
We train each stage for 100 epochs using Adam optimizer ~\citep{adam} with a learning rate of 5e-4 and batch size 512.

\section{Physics Environment}\label{appendix:physics}
\subsection{Detailed setups}\label{appendix:physics_details}
We provide an environment which consists of objects of different shapes and potentially different colors. Each object has a unique weight associated with it and only heavier objects can push lighter ones. This induces an acyclic tournament causal graph with sparse two-way interactions between the objects, which form the nodes of the graph.

More precisely, the physics environment with $M$ objects (eg. 3) and colormap $C$ (eg. blues) can be considered as the set $\{o_i = \{s_i, w_i, c_i, p_i\} \mid i = 1 \text{ to } M\}$ where $o_i$ denotes the $i^{th}$ object which is characterized by its position $p_i$, its shape $s_i$, its color $c_i$ and its weight $w_i$. An edge exists from $o_i$ to $o_j$ if and only if $w_i > w_j$. We consider the weight of each object to be unique, thereby getting rid of cycles. The specifics of the environment are determined by how the shape, color and weight of an object are related. For our experimentation, we consider two different settings which are outlined below. However, we emphasize that the physics environment is not limited to just these specifications and can be easily extended to form more complicated relationships between the three properties.

\subsection{Identity of Objects}
\label{appendix:physics_complexity}
Since we are proposing RL environments, we need to make sure that the mapping from the action space to the object space is well defined and observable / learnable. Here, we briefly discuss that it is the case in the settings of the physics environment proposed in this paper. We also discuss that in the \textit{Unobserved} environment this mapping can be very hard to learn and for this reason, we proposed another variant known as \textit{FixedUnobserved} environment.

Our mapping from action space to object space is such that given an initialization of the environment, the first action dimension always corresponds to the heaviest object. Similarly, the second to the second heaviest and so on.

Now, in the \textit{Observed} environment case, the heaviest object is also the darkest object in the scene so it is relatively easy for a model to infer the action to object mapping once it has learned the fact that intensity of color represents the weight of the object.

On the other hand, in the \textit{Unobserved} case, the colors of the objects are sampled without replacement from a larger set of colors. For example, consider a 3 object environment with the set of colors to be red < green < orange < yellow where the ordering defines the ordering of the weight. Then if in one initialization has the colors (red, green, yellow) then here the first action dimension corresponds to the color red. However, another initialization of the same environment can be (green orange, yellow) and then the first action dimension would correspond to the green object. Thus, for a model to learn the action to object mapping, it has to learn this global ranking of colors. We found that this was typically hard for the models to do.

To alleviate the above complexity, we consider another setting \textit{FixedUnobserved} where we keep the shapes of the objects fixed and unique. Here, there is an additional constraint that apart from the colors following a global ordering of weights, the unique shapes also follow a global ordering of weights and hence, this creates an easily learnable mapping.

\subsection{All variables are observed}\label{appendix:physics_observed}
In this setting, we consider all the objects to be of the same color but different shades, eg. different shades of the color blue. The weight of each object is a monotonic function of its color intensity, meaning that darker objects are heavier.

Mathematically, given a colormap $C$ (single color; continuous in intensity of the color), $c_i \in [0,1]$ denotes the intensity of the color $C$ for object $i$ (1 being darkest; 0 lightest). Moreover, the weight of that object is given by $w_i = g(c_i)$, where g is a strictly monotonic function. Thus, darker objects are given heavier weights and thus can push lighter objects.

This setting easily allows for zero shot generalization since a model that has been trained on a subset of shades of a particular color can generalize to do well across different shades of the same color. Moreover, the shape of an object here is a distractor since the dynamics of the objects are only controlled by their colors.

\subsection{Some variables are unobserved}\label{appendix:physics_unobserved}
In this setting, all objects are of distinct discrete colors drawn from a discrete colormap $c$. Each color is associated with a unique weight and here, too, heavier objects can push lighter ones but not vice versa.

Mathematically, given a colormap $C$ (multiple discrete options), $c_i \in C$ denotes the color for object $i$ such that $c_i \neq c_j \;\forall i \neq j$. Moreover, the weight of that object is given by $w_i = g(h_i)$, where g is an injective function and $g : C \to \mathbb{R}$.

This setting does not allow for zero shot generalization in the colors since whenever a new color is introduced, the agent will have to perform interventions on it to infer its place in the graph. However, similar to the observed case, the shapes of the objects act as distractors since the dynamics is only controlled by the colors.

\subsection{Unobserved Variables but Fixed Shapes}\label{appendix:physics_fixedunobserved}
In this setting, all objects are of distinct discrete colors and shapes where the set of shapes is kept constant across different episodes. Here, the weight of an object can be reflected either from its shape or its color. For example, the lightest object in the episode will always be of a fixed unique shape and it will always have the lightest color (where lightest color is defined according to the order on the color in the colormap - eg. red < blue < green)

This setting does not allow for zero shot generalization in either the colors or the shapes since whenever a new color or shape is introduced, the agent will have to perform interventions on it to infer its place in the graph.

\subsection{Experimental Results}
\label{appendix:physics_insight}
We perform experiments on a wide range of settings for the underlying causal graph for the physics environment. We categorize our findings below -

\begin{figure}
    \centering
    \includegraphics[width=15cm]{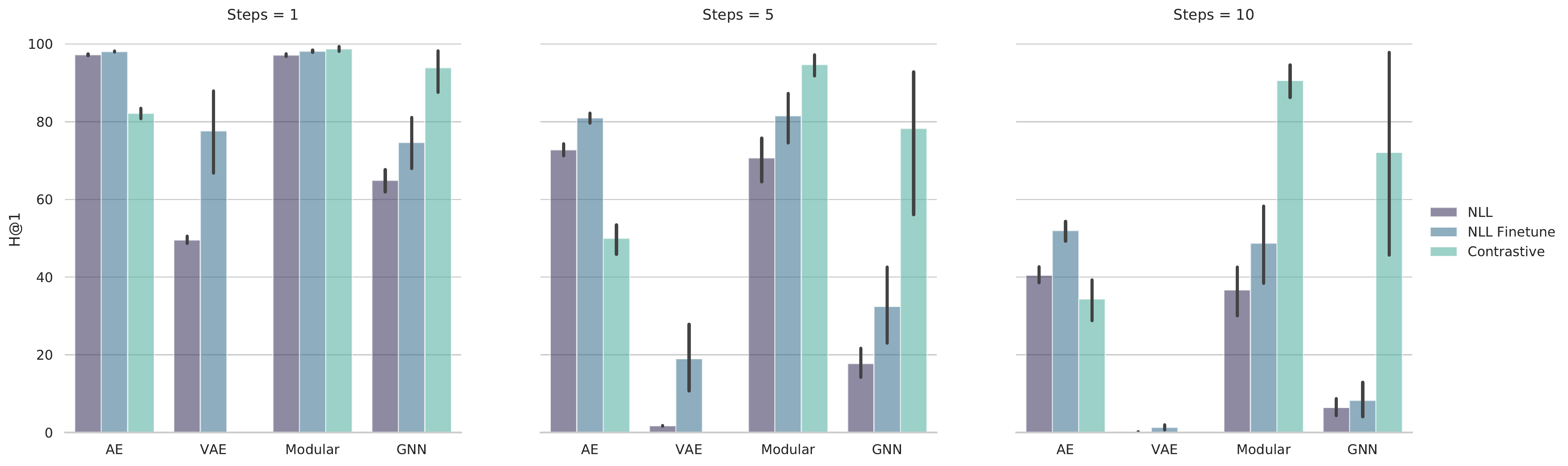}
    \includegraphics[width=15cm]{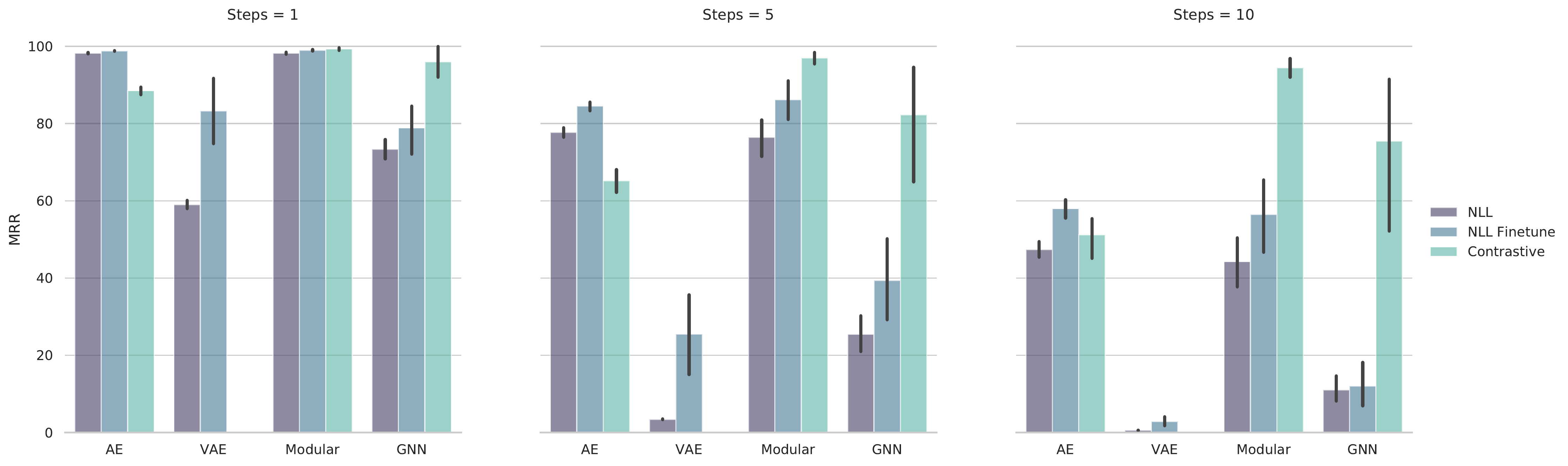}
    \includegraphics[width=15cm]{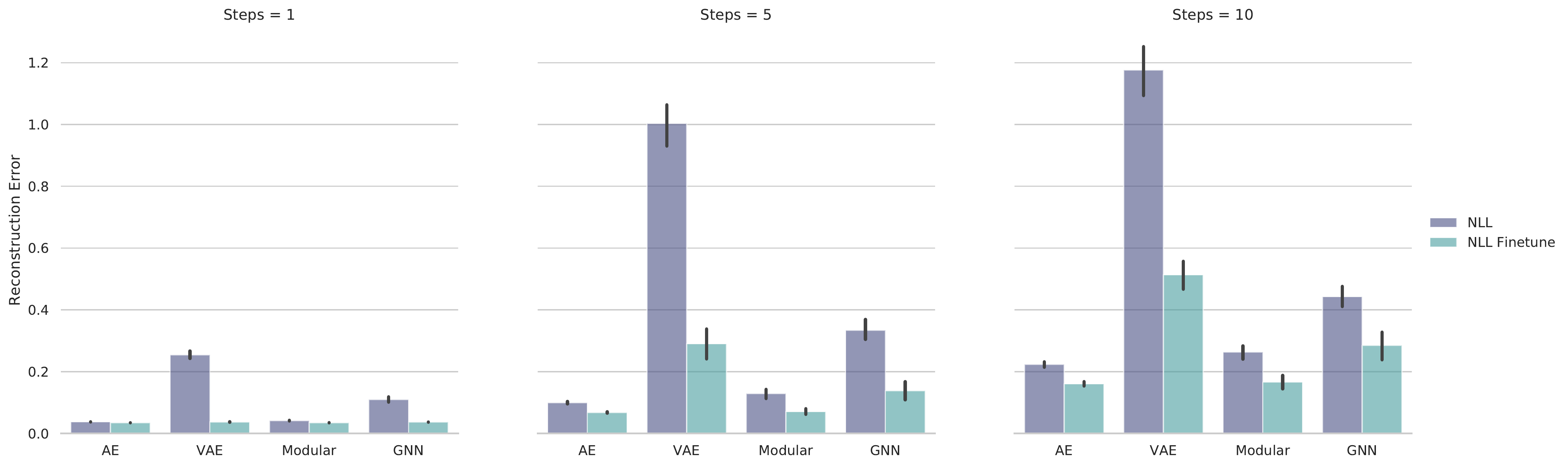}
    \caption{Hits at Rank 1 (H@1), Mean Reciprocal Rank (MRR) \textit{(higher is better)} and Reconstruction Error \textit{(lower is better)} for different models and training losses for 1, 5 and 10 step prediction for the Observed Physics environment setting with 3 objects.}
    \label{fig:physics_observed_3obj}
\end{figure}

\begin{figure}
    \centering
    \includegraphics[width=15cm]{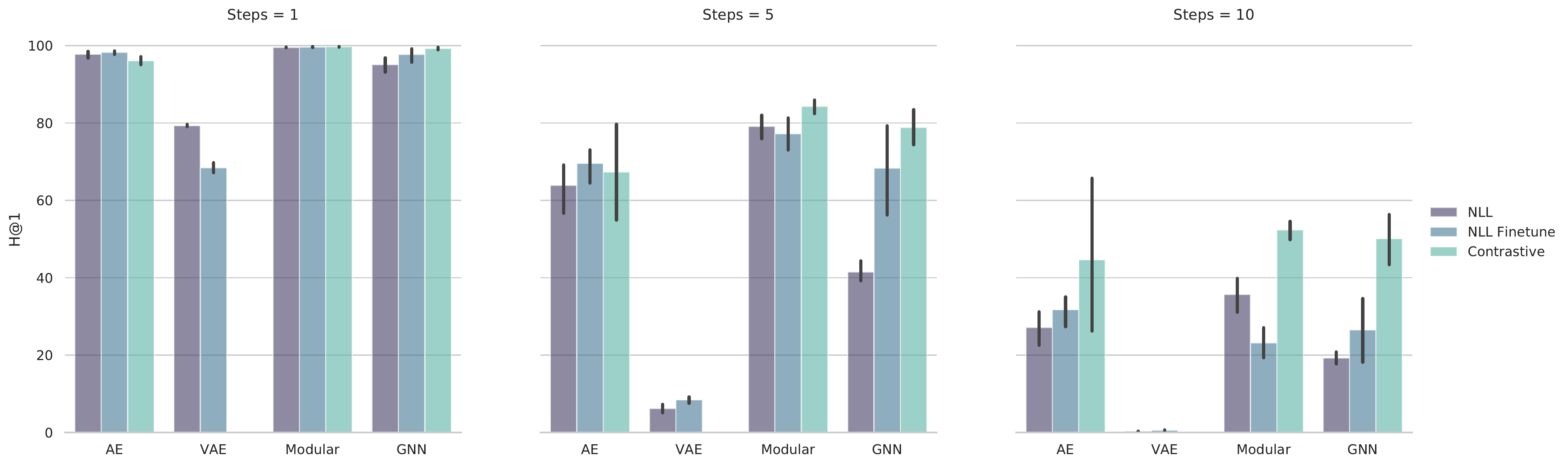}
    \includegraphics[width=15cm]{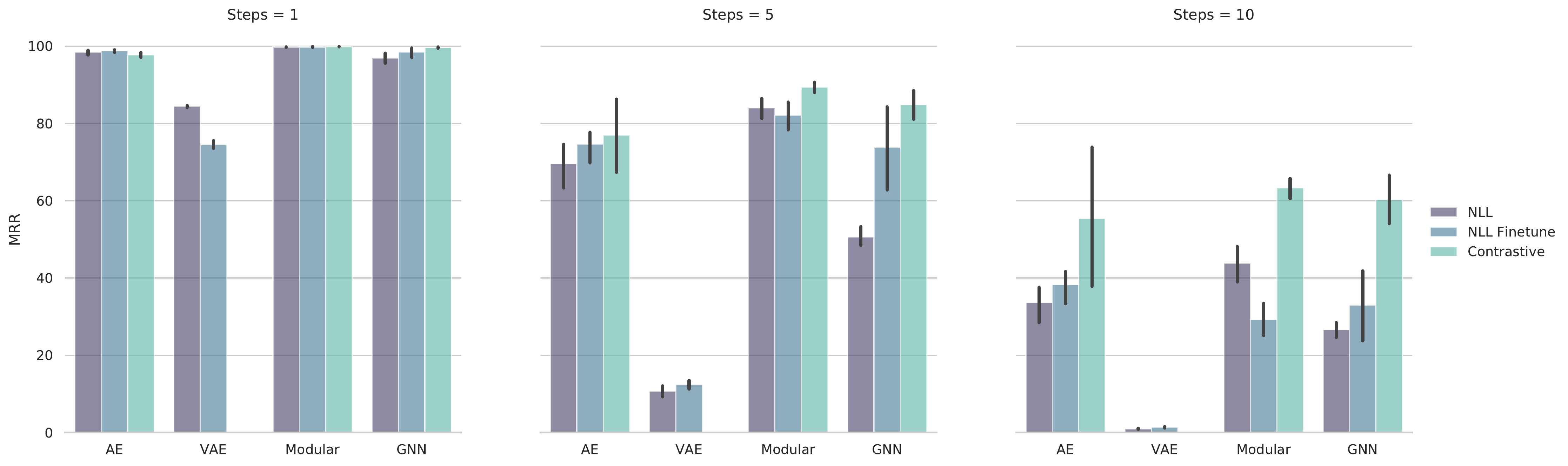}
    \includegraphics[width=15cm]{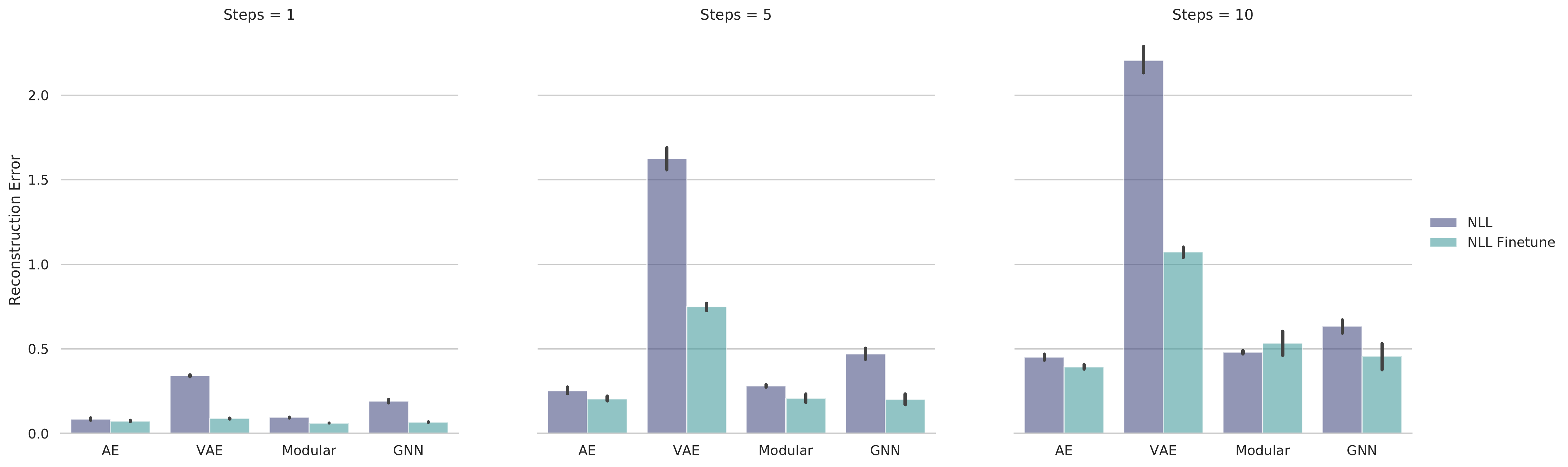}
    \caption{Hits at Rank 1 (H@1), Mean Reciprocal Rank (MRR) \textit{(higher is better)} and Reconstruction Error \textit{(lower is better)} for different models and training losses for 1, 5 and 10 step prediction for the Observed Physics environment setting with 5 objects.}
    \label{fig:physics_observed_5obj}
\end{figure}

\begin{figure}
    \centering
    \includegraphics[width=15cm]{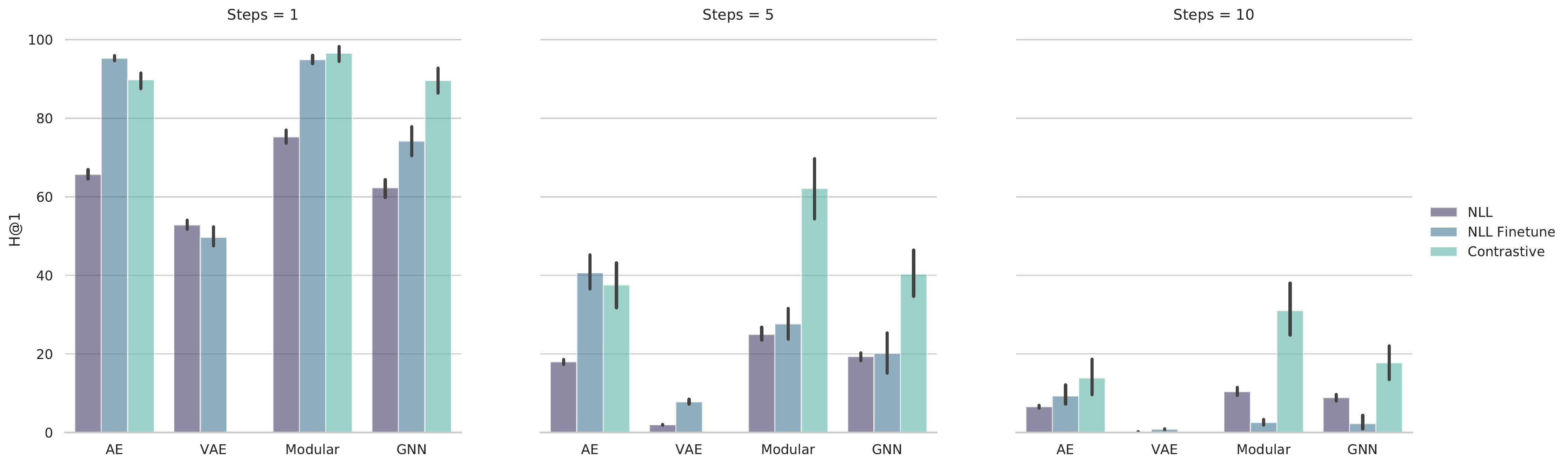}
    \includegraphics[width=15cm]{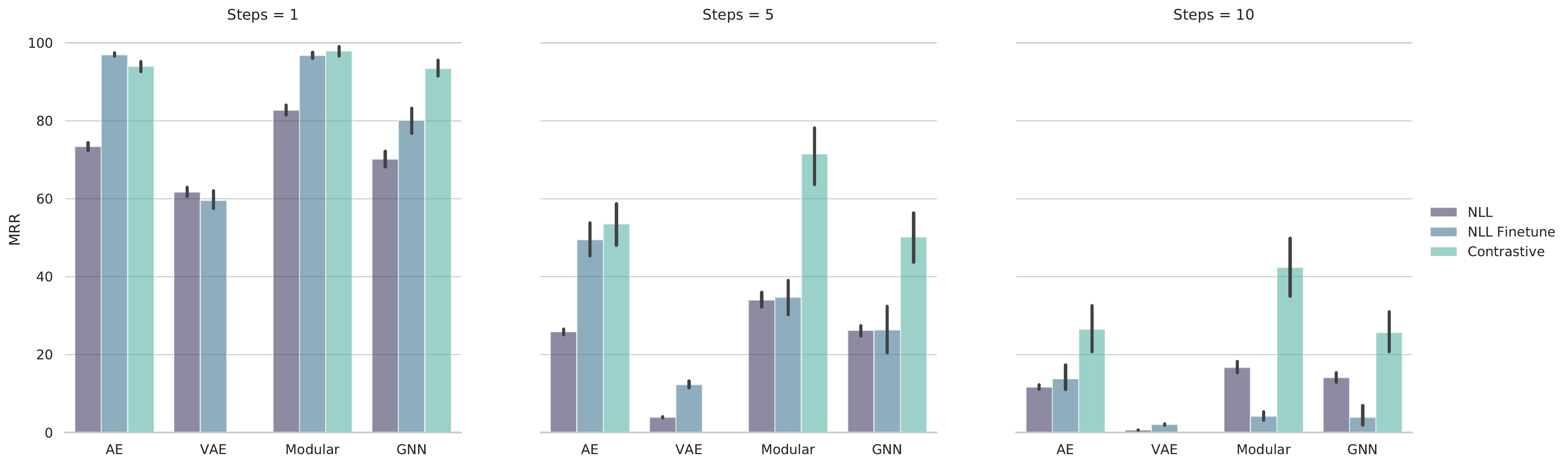}
    \includegraphics[width=15cm]{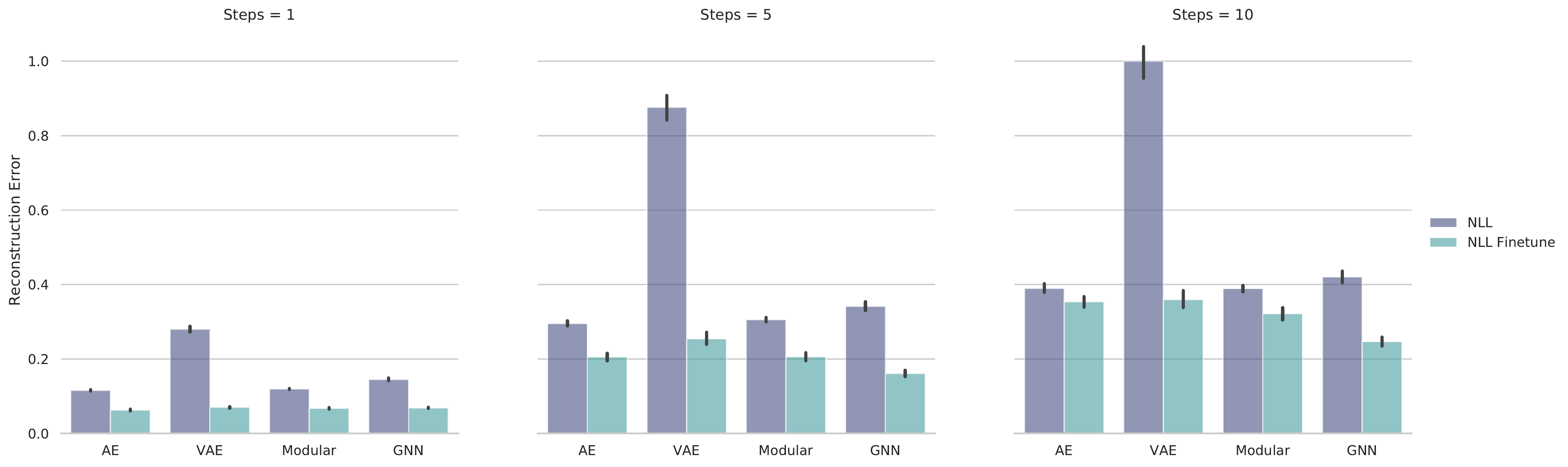}
    \caption{Hits at Rank 1 (H@1), Mean Reciprocal Rank (MRR) \textit{(higher is better)} and Reconstruction Error \textit{(lower is better)} for different models and training losses for 1, 5 and 10 step prediction for the Unobserved Physics environment setting with 3 objects.}
    \label{fig:physics_unobserved_3obj}
\end{figure}

\begin{figure}
    \centering
    \includegraphics[width=15cm]{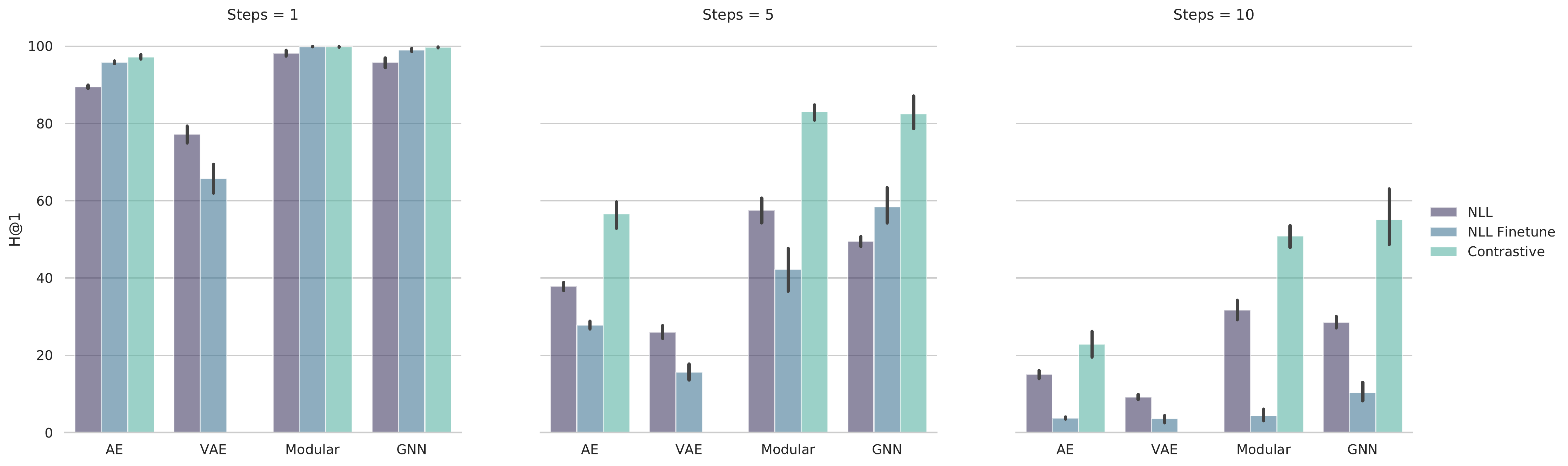}
    \includegraphics[width=15cm]{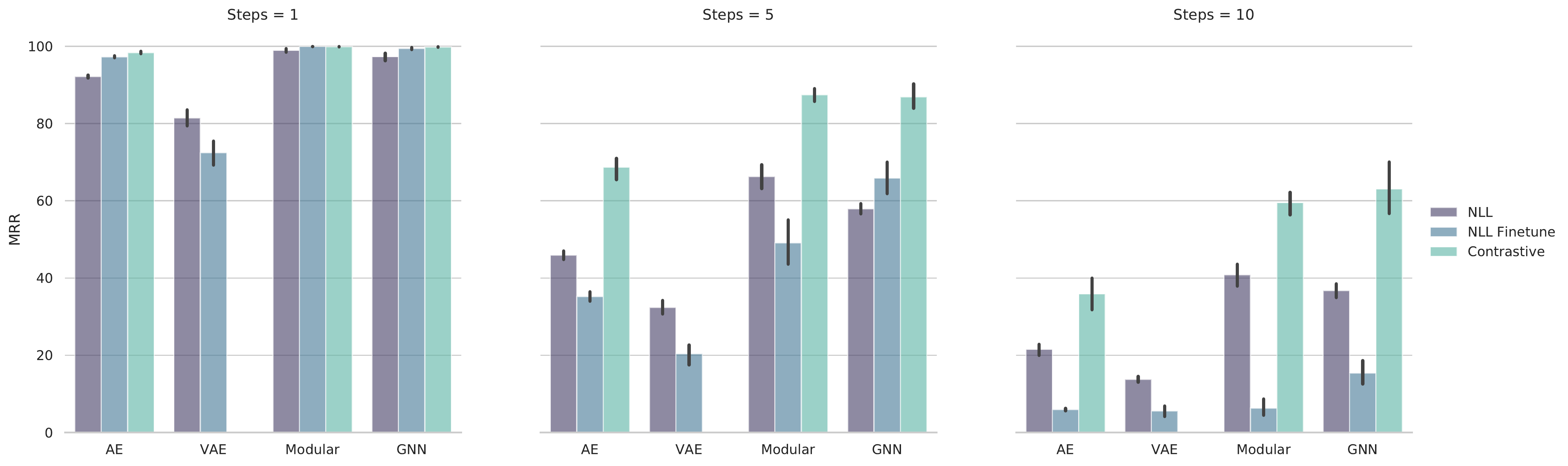}
    \includegraphics[width=15cm]{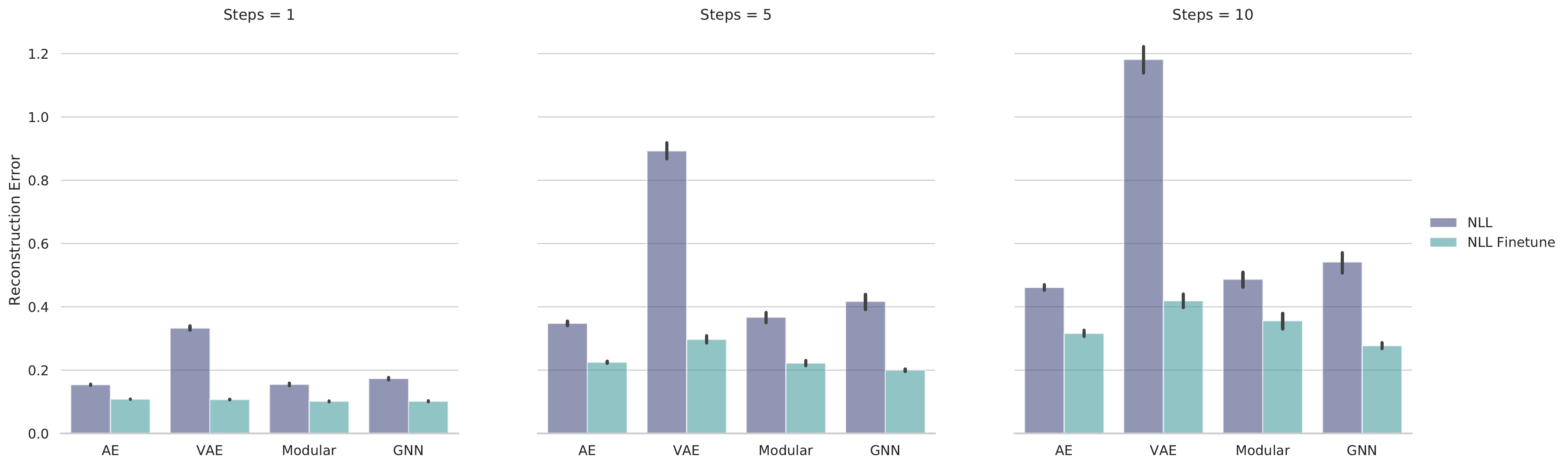}
    \caption{Hits at Rank 1 (H@1), Mean Reciprocal Rank (MRR) \textit{(higher is better)} and Reconstruction Error \textit{(lower is better)} for different models and training losses for 1, 5 and 10 step prediction for the Unobserved Physics environment setting with 5 objects.}
    \label{fig:physics_unobserved_5obj}
\end{figure}

\begin{figure}
    \centering
    \includegraphics[width=15cm]{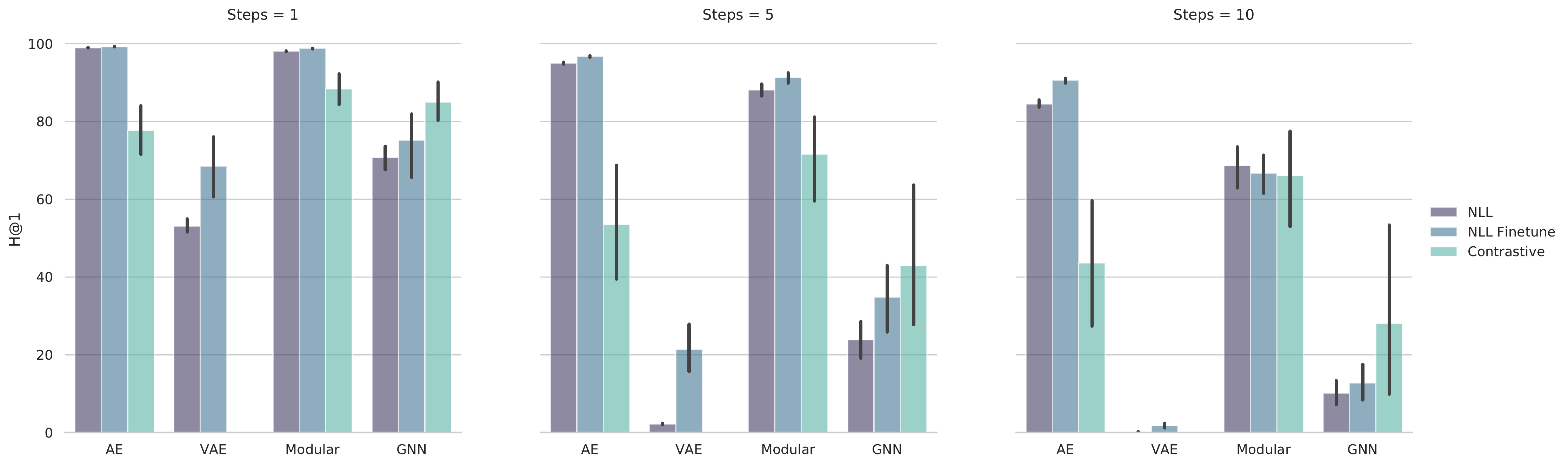}
    \includegraphics[width=15cm]{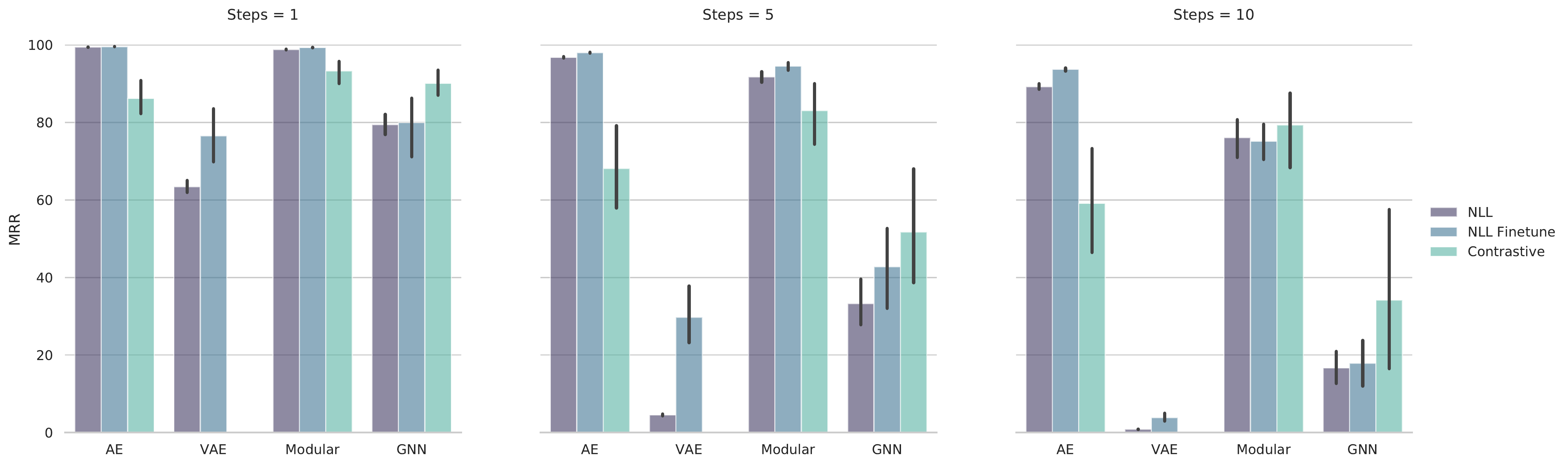}
    \includegraphics[width=15cm]{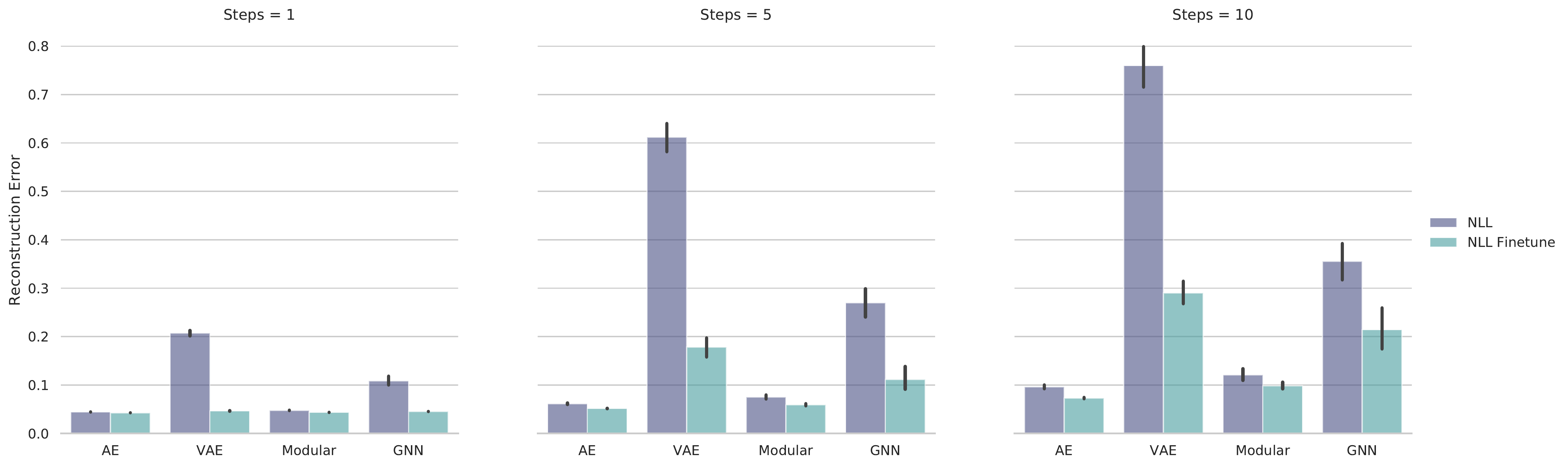}
    \caption{Hits at Rank 1 (H@1), Mean Reciprocal Rank (MRR) \textit{(higher is better)} and Reconstruction Error \textit{(lower is better)} for different models and training losses for 1, 5 and 10 step prediction for the FixedUnobserved Physics environment setting with 3 objects.}
    \label{fig:physics_fixedunobserved_3obj}
\end{figure}

\begin{figure}
    \centering
    \includegraphics[width=15cm]{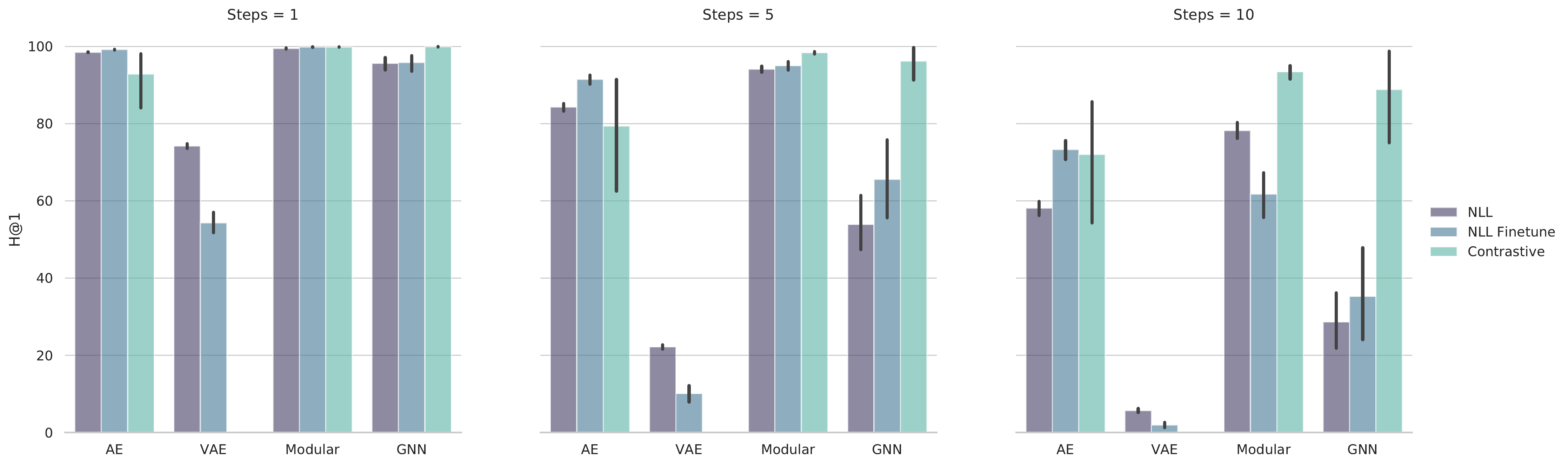}
    \includegraphics[width=15cm]{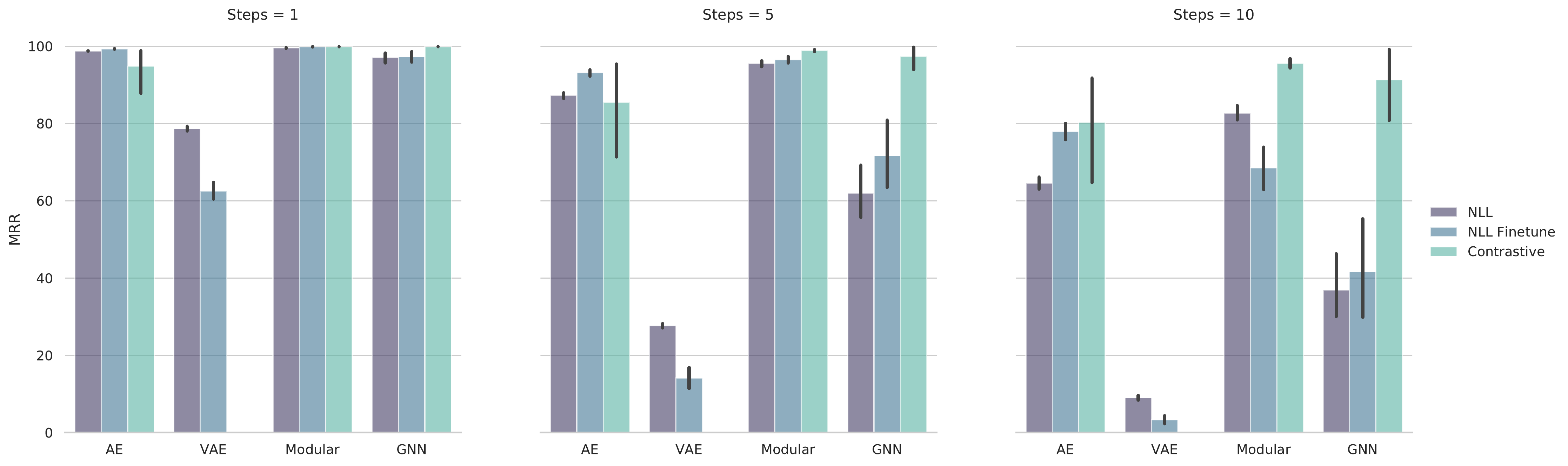}
    \includegraphics[width=15cm]{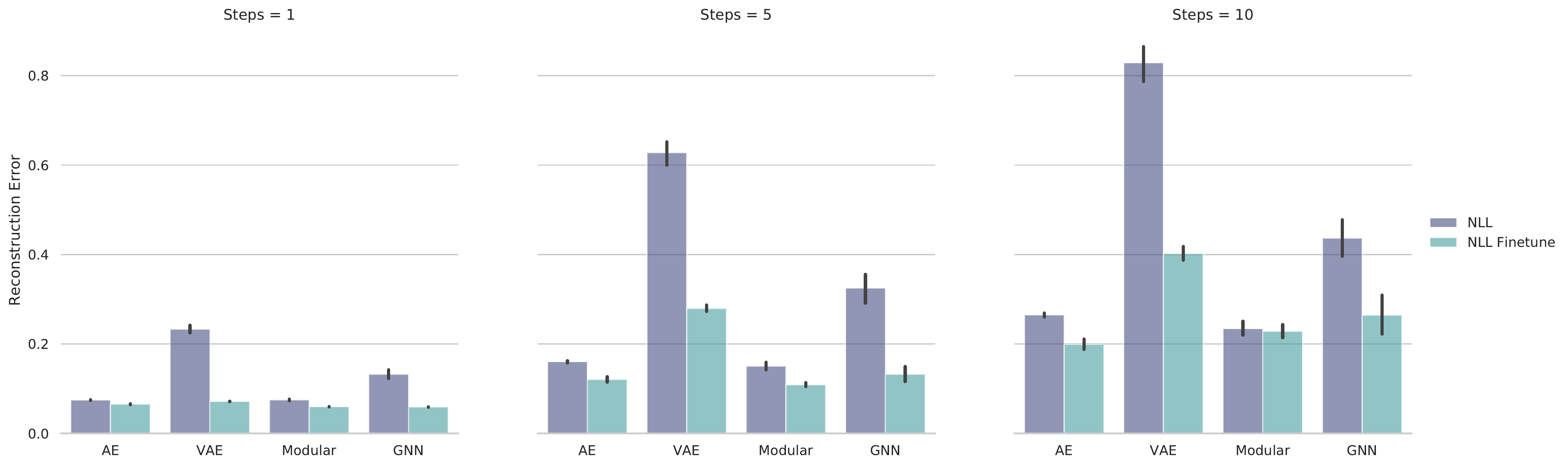}
    \caption{Hits at Rank 1 (H@1), Mean Reciprocal Rank (MRR) \textit{(higher is better)} and Reconstruction Error \textit{(lower is better)} for different models and training losses for 1, 5 and 10 step prediction for the FixedUnobserved Physics environment setting with 5 objects.}
    \label{fig:physics_fixedunobserved_5obj}
\end{figure}

\begin{figure}
    \centering
    \includegraphics[width=15cm]{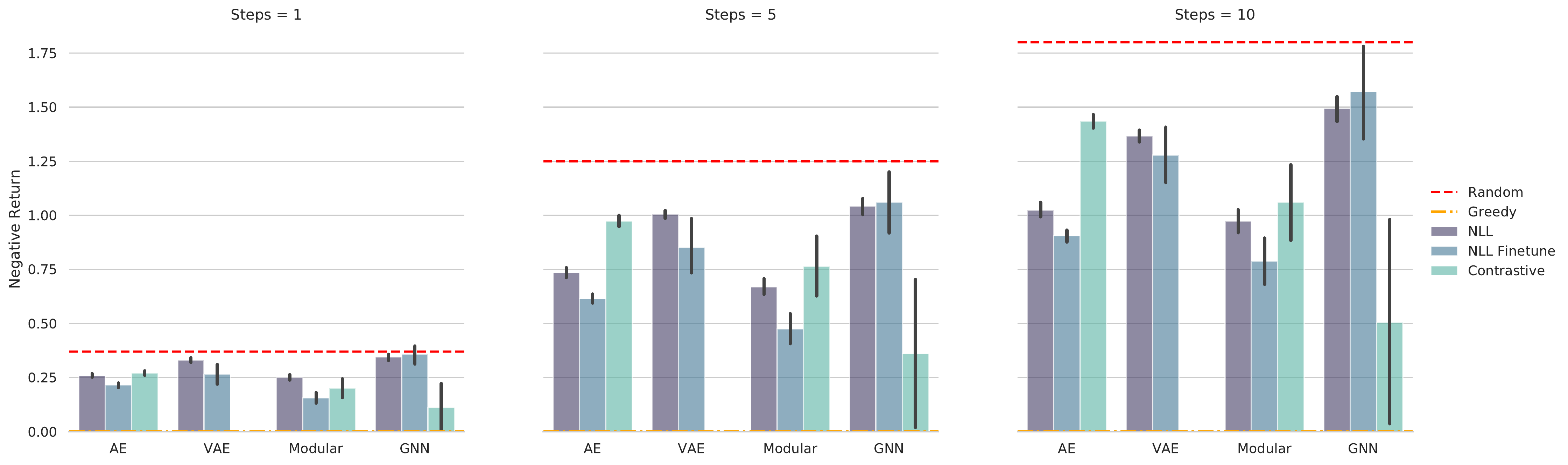}
    \includegraphics[width=15cm]{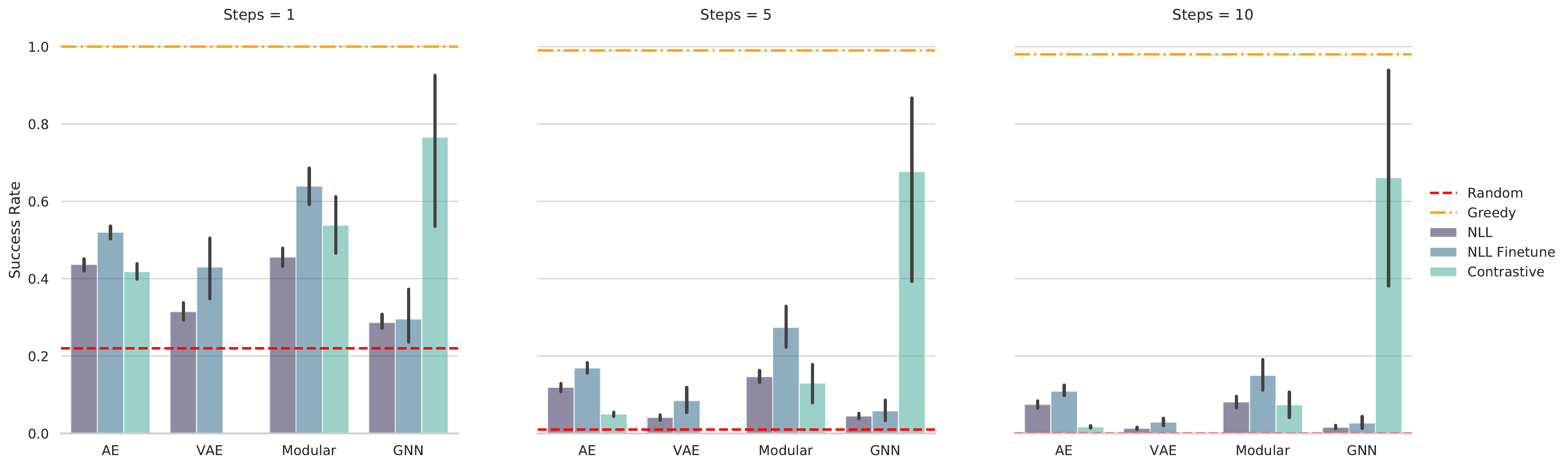}
\caption{Negative Return \textit{(lower is better)} and Success Rate \textit{(higher is better)} for different models and training losses for 1, 5 and 10 step prediction for the Observed Physics environment setting with 3 objects.}
\label{fig:physics_observed_3obj_rl}
\end{figure}

\begin{figure}
    \centering
    \includegraphics[width=15cm]{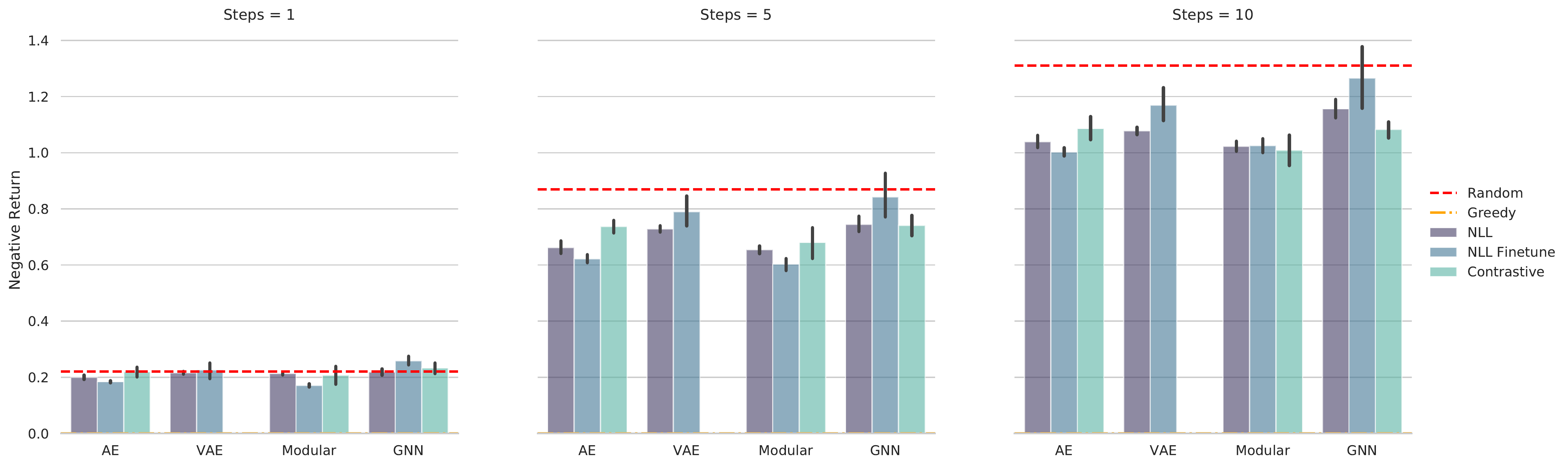}
    \includegraphics[width=15cm]{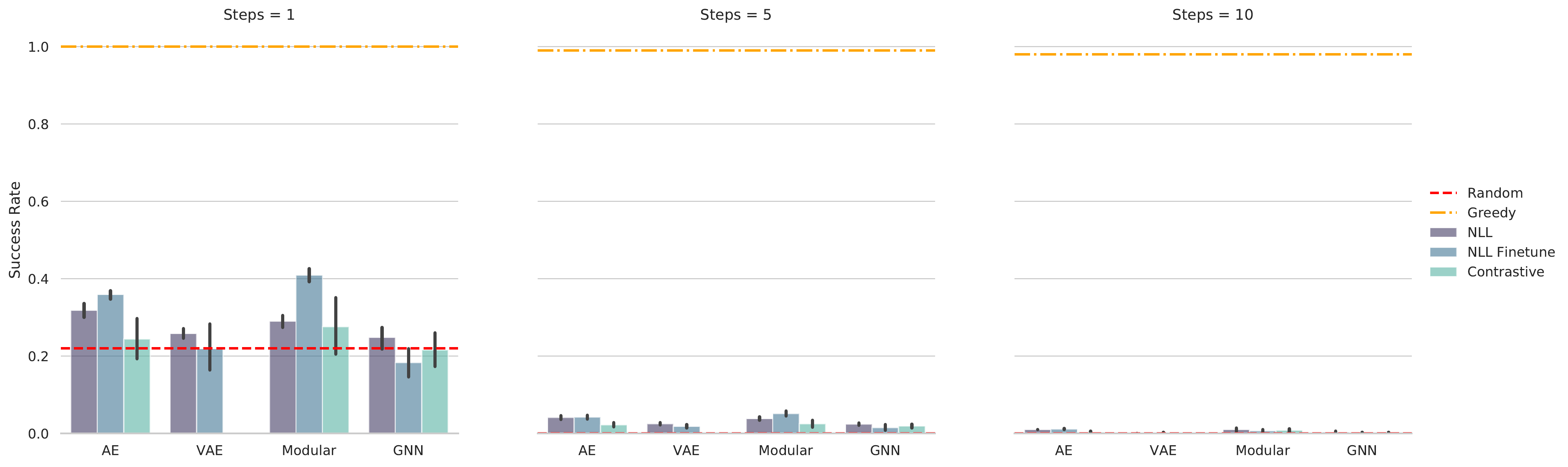}
\caption{Negative Return \textit{(lower is better)} and Success Rate \textit{(higher is better)} for different models and training losses for 1, 5 and 10 step prediction for the Observed Physics environment setting with 5 objects.}
\label{fig:physics_observed_5obj_rl}
\end{figure}

\begin{figure}
    \centering
    \includegraphics[width=15cm]{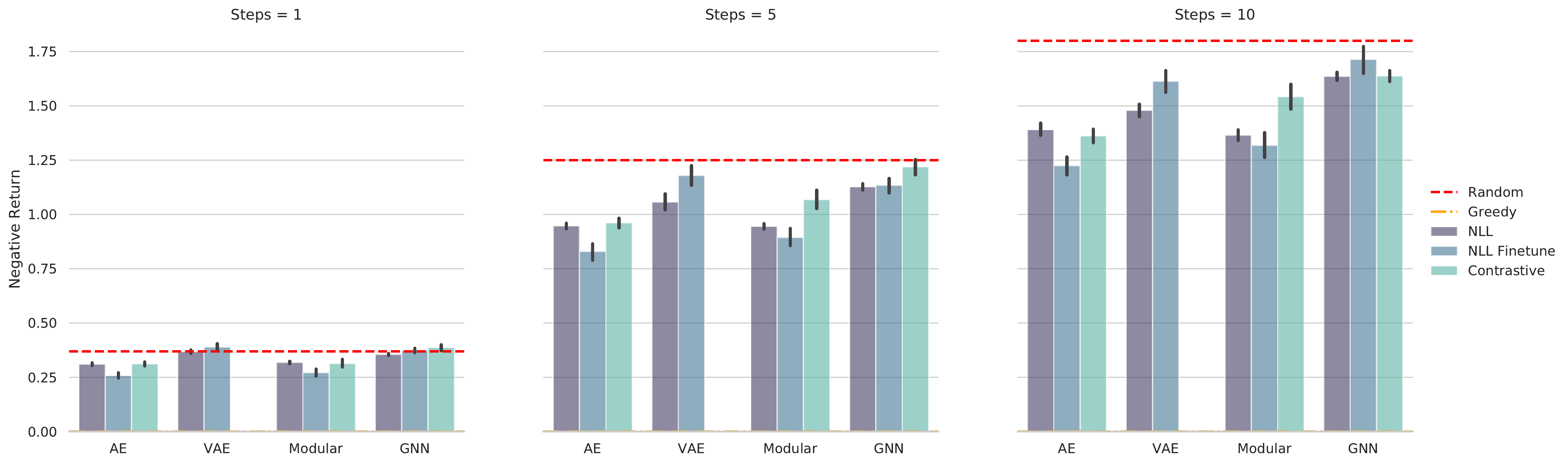}
    \includegraphics[width=15cm]{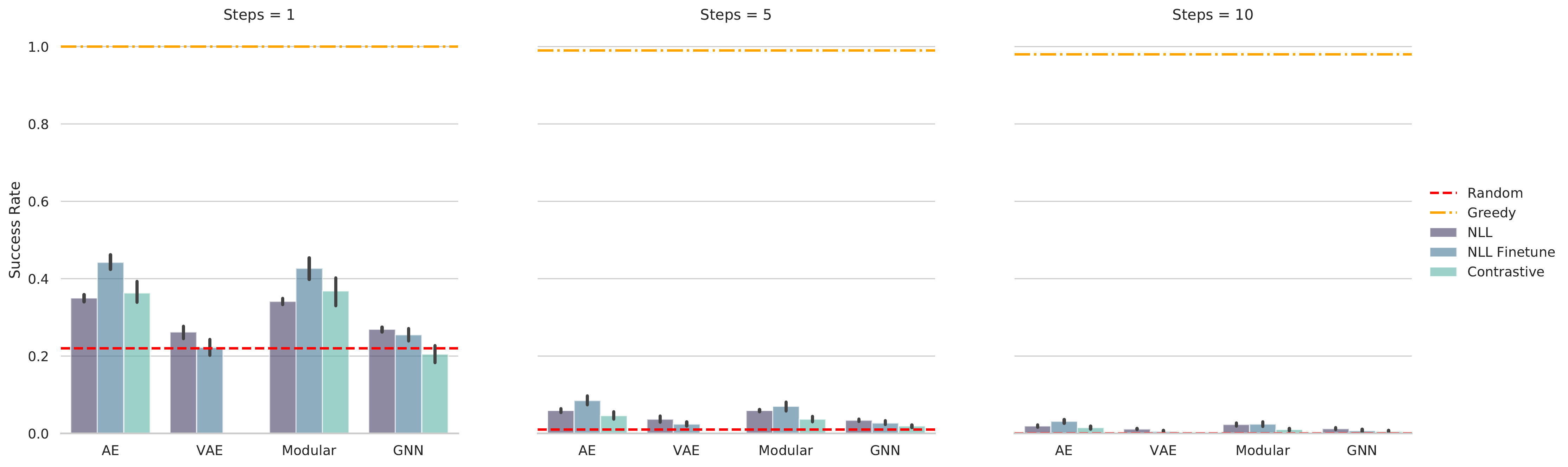}
\caption{Negative Return \textit{(lower is better)} and Success Rate \textit{(higher is better)} for different models and training losses for 1, 5 and 10 step prediction for the Unobserved Physics environment setting with 3 objects.}
\label{fig:physics_unobserved_3obj_rl}
\end{figure}

\begin{figure}
    \centering
    \includegraphics[width=15cm]{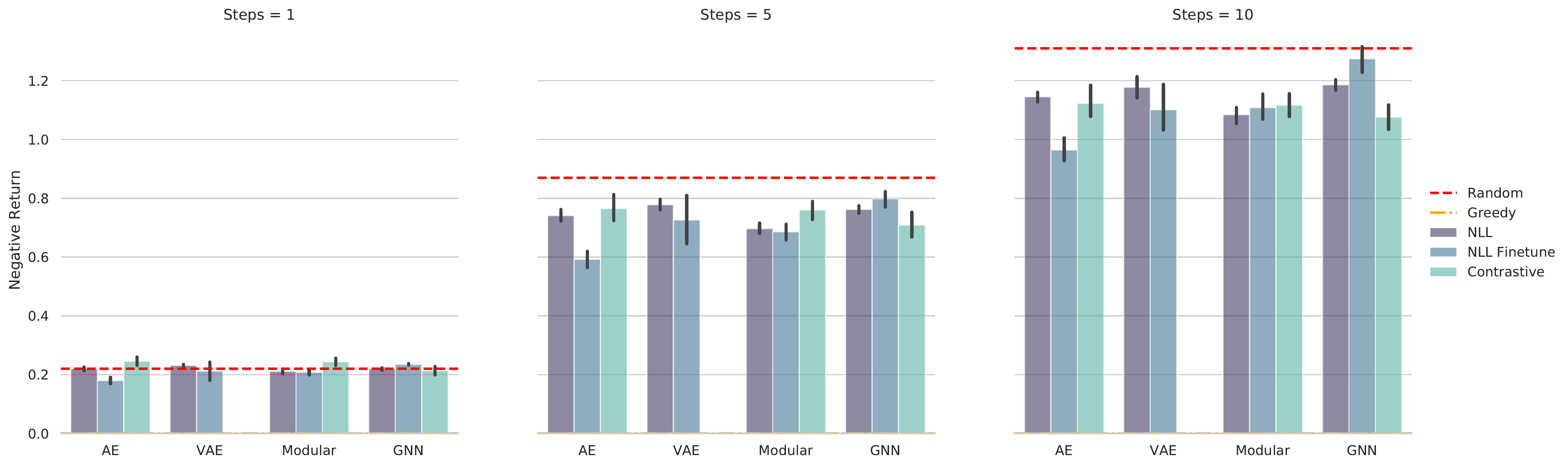}
    \includegraphics[width=15cm]{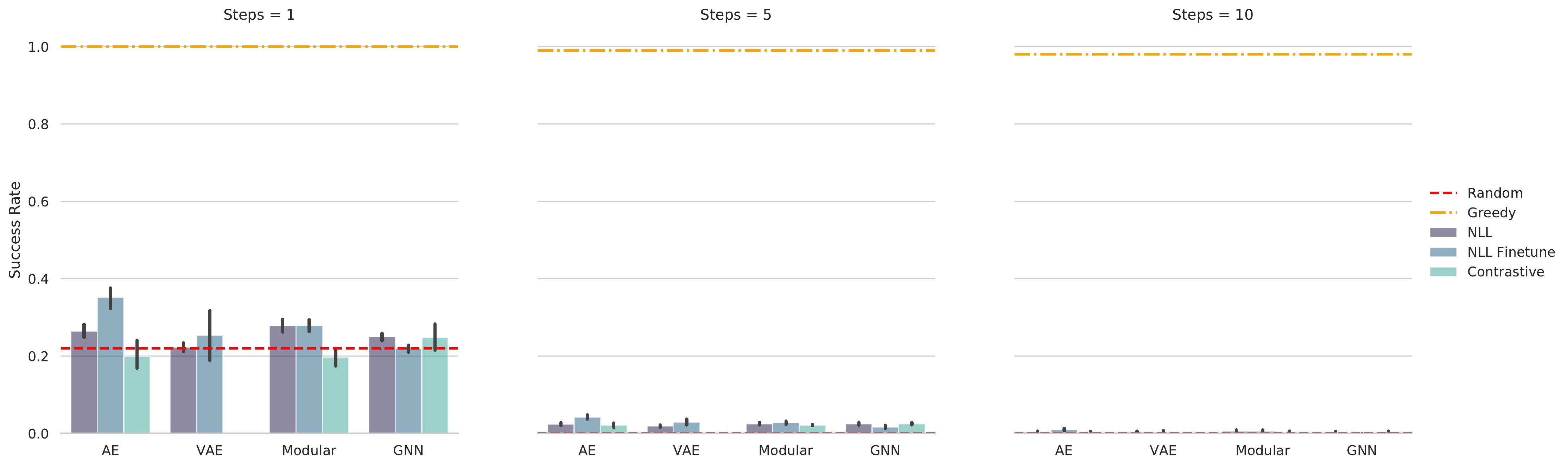}
\caption{Negative Return \textit{(lower is better)} and Success Rate \textit{(higher is better)} for different models and training losses for 1, 5 and 10 step prediction for the Unobserved Physics environment setting with 5 objects.}
\label{fig:physics_unobserved_5obj_rl}
\end{figure}

\begin{figure}
    \centering
    \includegraphics[width=15cm]{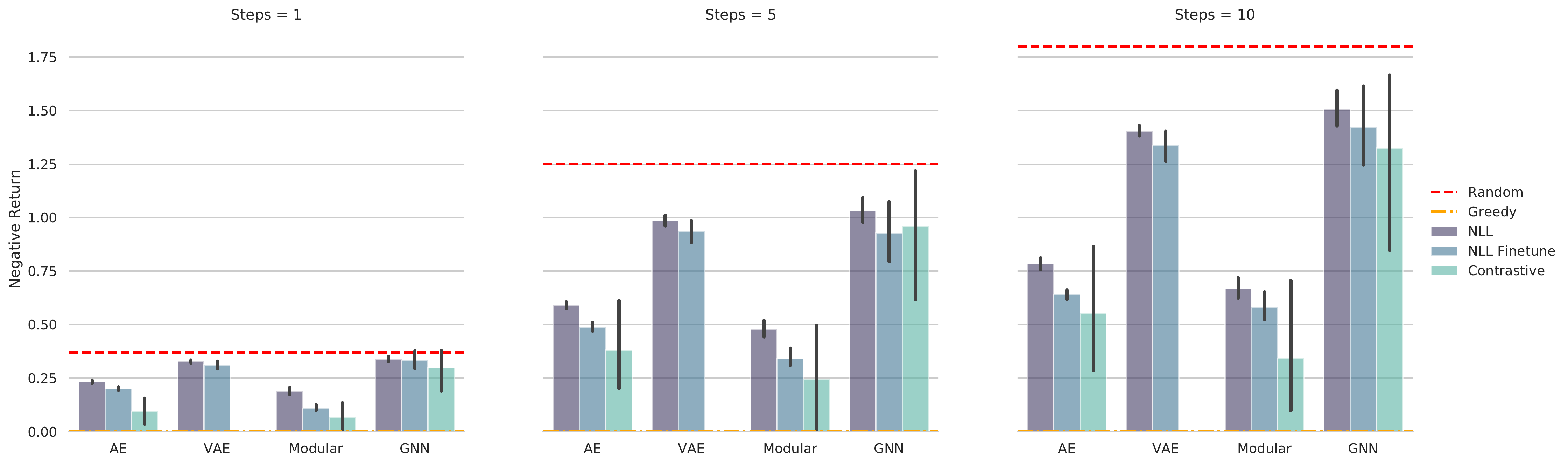}
    \includegraphics[width=15cm]{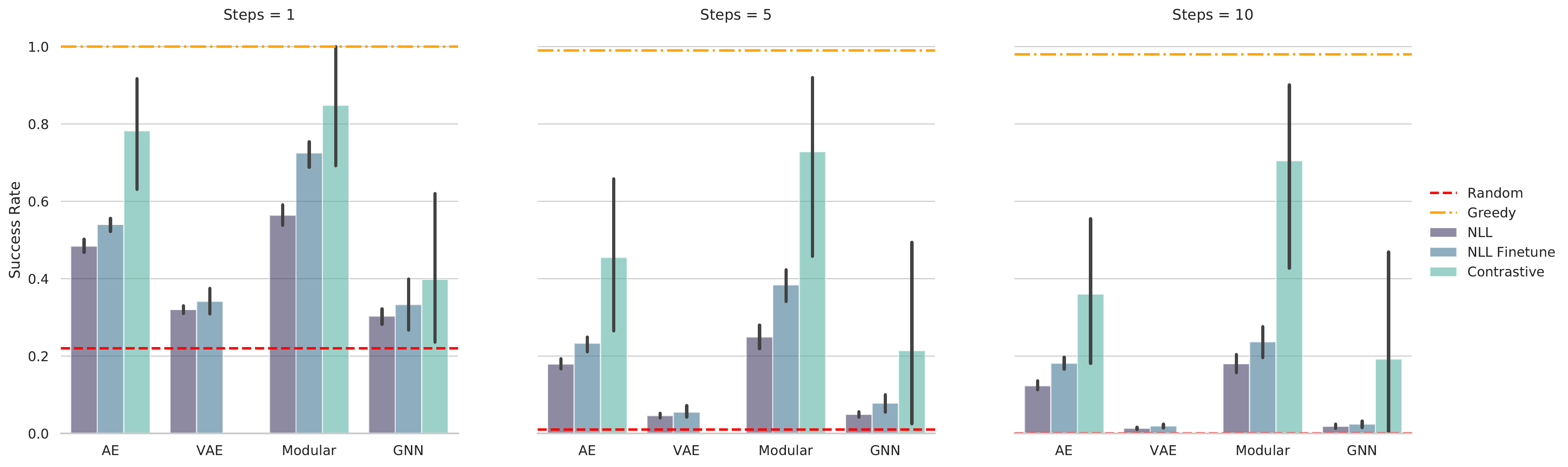}
\caption{Negative Return \textit{(lower is better)} and Success Rate \textit{(higher is better)} for different models and training losses for 1, 5 and 10 step prediction for the FixedUnobserved Physics environment setting with 3 objects.}
\label{fig:physics_fixedunobserved_3obj_rl}
\end{figure}

\begin{figure}
    \centering
    \includegraphics[width=15cm]{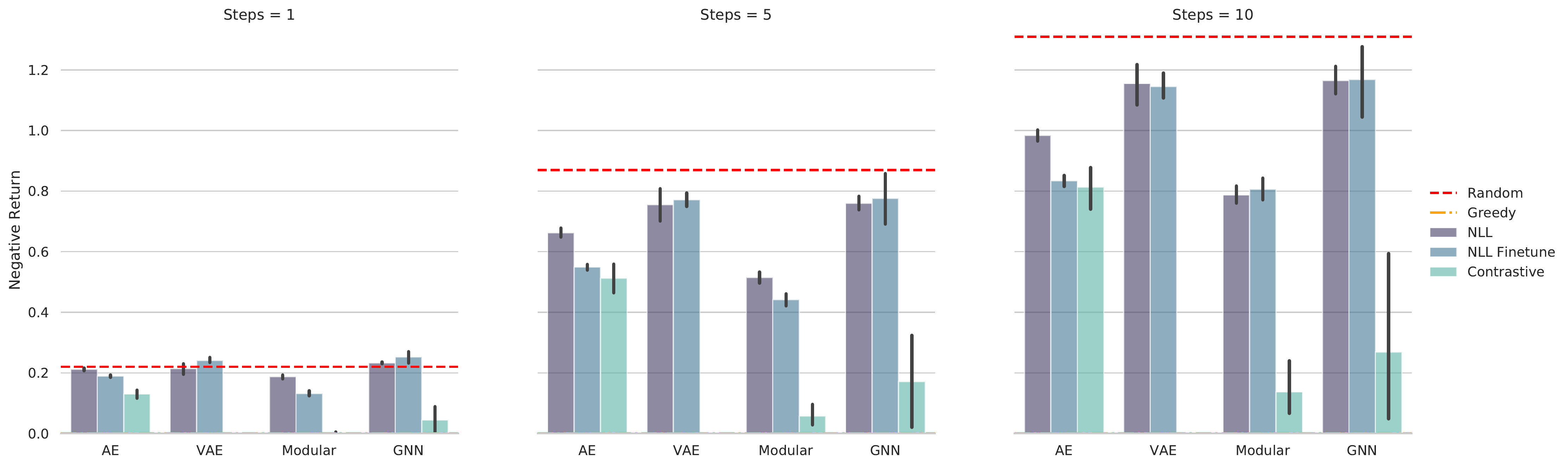}
    \includegraphics[width=15cm]{Figures/Physics_env_rl/FixedUnobserved/5/Success_Rate.pdf}
\caption{Negative Return \textit{(lower is better)} and Success Rate \textit{(higher is better)} for different models and training losses for 1, 5 and 10 step prediction for the FixedUnobserved Physics environment setting with 5 objects.}
\label{fig:physics_fixedunobserved_5obj_rl}
\end{figure}

\begin{itemize}
    \item \textit{Graph Neural Networks (GNNs)} generally don't perform well compared to \textit{Modular models} and \textit{Autoencoders (AEs)} on a wide variety of metrics (ranking metrics, reconstruction loss, downstream RL task) in the setting of \textit{likelihood based loss} (refer to \Cref{fig:physics_observed_3obj} - \Cref{fig:physics_fixedunobserved_5obj_rl} and \Cref{tab:physics_observed_3obj} - \ref{tab:physics_fixedunobserved_5obj_rl})
    \item Models trained with \textit{contrastive loss} are generally better at predictions made over longer time scales in terms of ranking metrics (refer to \Cref{fig:physics_observed_3obj} - \ref{fig:physics_fixedunobserved_5obj} and \Cref{tab:physics_observed_3obj} - \ref{tab:physics_fixedunobserved_5obj})
    \item Models trained with \textit{contrastive loss} are also generally better at downstream RL tasks as compared to those trained with \textit{likelihood based loss}. In particular there are some settings where the former were able to do almost perfect planning while the latter weren't able to do good planning in any setting (refer to \Cref{fig:physics_unobserved_3obj_rl,fig:physics_fixedunobserved_3obj_rl,fig:physics_fixedunobserved_5obj_rl} and \Cref{tab:physics_observed_3obj_rl,tab:physics_fixedunobserved_3obj_rl,tab:physics_fixedunobserved_5obj_rl})
    \item Modular models and Graph Neural Networks scale better than the monolithic counterparts when the number of objects in the causal graph increases. Further, while the ranking metrics still remain good, we see that the planning metrics suffer by a large margin (refer to \Cref{fig:physics_observed_3obj} - \ref{fig:physics_fixedunobserved_5obj_rl} and \Cref{tab:physics_observed_3obj} - \ref{tab:physics_fixedunobserved_5obj_rl})
    \item While Autoencoder models perform decently based on ranking metrics, they generally don't perform as well on downstream RL tasks when compared to Graph Neural Networks and Modular models (refer to \Cref{fig:physics_observed_3obj_rl} - \ref{fig:physics_fixedunobserved_5obj_rl} and \Cref{tab:physics_observed_3obj_rl} - \ref{tab:physics_fixedunobserved_5obj_rl})
    \item While ranking metrics on the unobserved environment are still decent (refer to \Cref{fig:physics_unobserved_3obj,fig:physics_unobserved_5obj} and \Cref{tab:physics_unobserved_3obj,tab:physics_unobserved_5obj}), we see that in terms of downstream RL planning, none of the models do much better than a random policy (refer to \Cref{fig:physics_unobserved_3obj_rl,fig:physics_unobserved_5obj_rl} and \Cref{fig:physics_unobserved_3obj_rl,fig:physics_unobserved_5obj_rl})
    \item We see a case where models that have very good ranking metrics over long time horizons (AE with NLL Finetune; \Cref{fig:physics_fixedunobserved_3obj} and \Cref{fig:physics_fixedunobserved_3obj_rl}) perform much worse on downstream RL tasks than GNNs and Modular models which had lower ranking metrics (\Cref{tab:physics_fixedunobserved_3obj_rl} and \Cref{fig:physics_fixedunobserved_3obj_rl}).
\end{itemize}

\begin{figure}
    \centering
    \includegraphics[width=15cm]{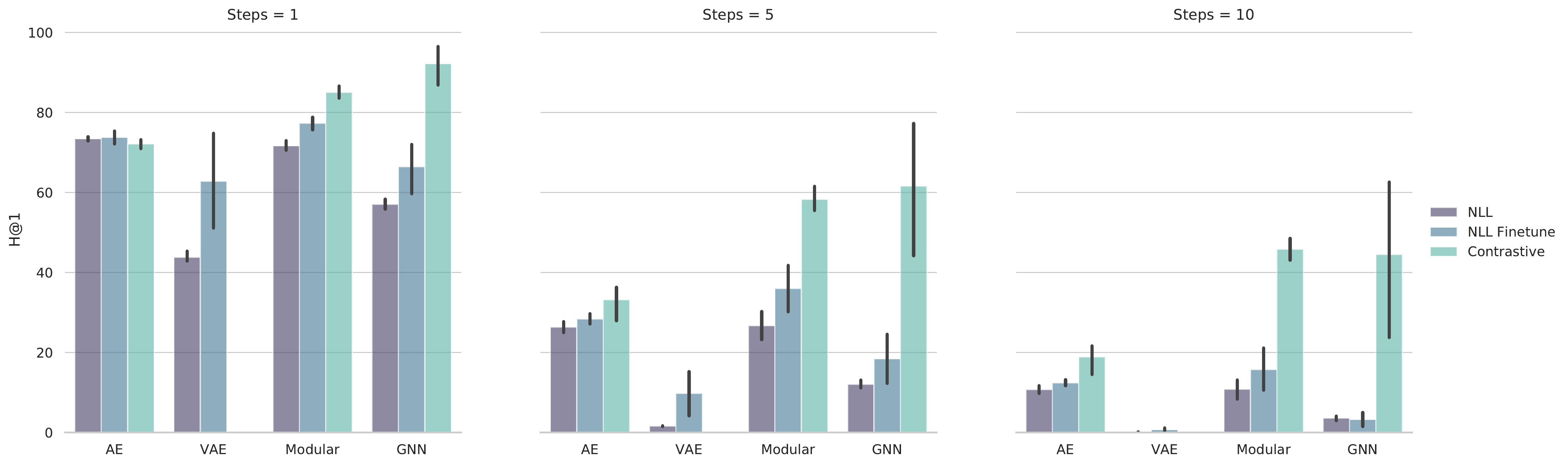}
    \includegraphics[width=15cm]{Figures/Physics_env/Observed/3/MRR.pdf}
    \includegraphics[width=15cm]{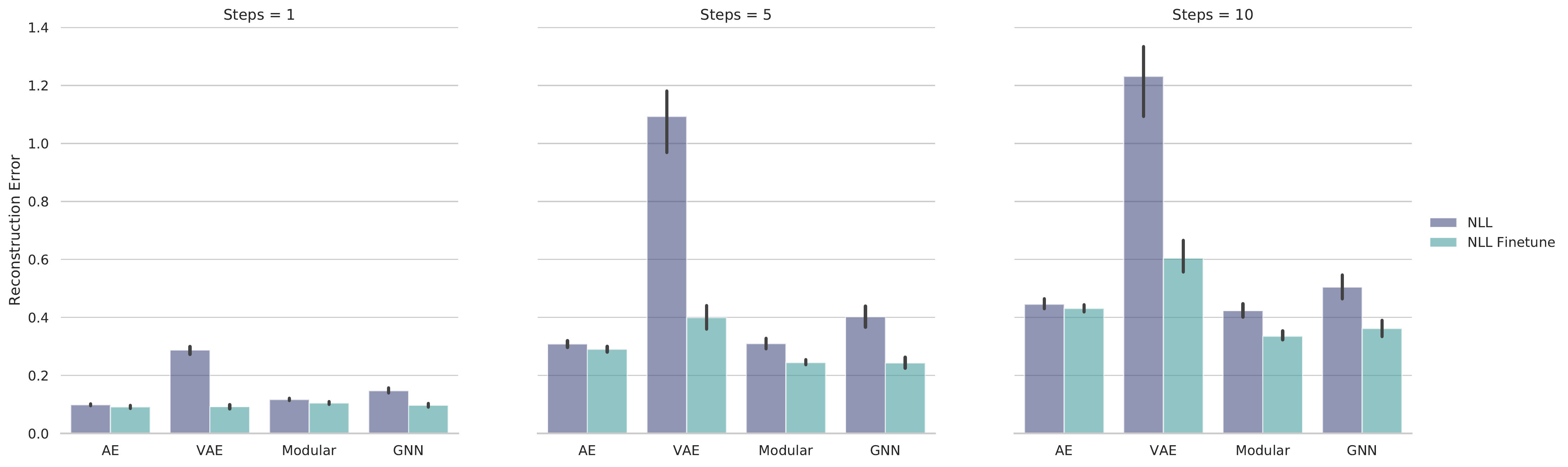}
    \caption{Hits at Rank 1 (H@1), Mean Reciprocal Rank (MRR) \textit{(higher is better)} and Reconstruction Error \textit{(lower is better)} for different models and training losses for 1, 5 and 10 step prediction for the Observed Physics environment Zero Shot setting with 3 objects.}
    \label{fig:physics_observed_3obj_0shot}
\end{figure}

\begin{figure}
    \centering
    \includegraphics[width=15cm]{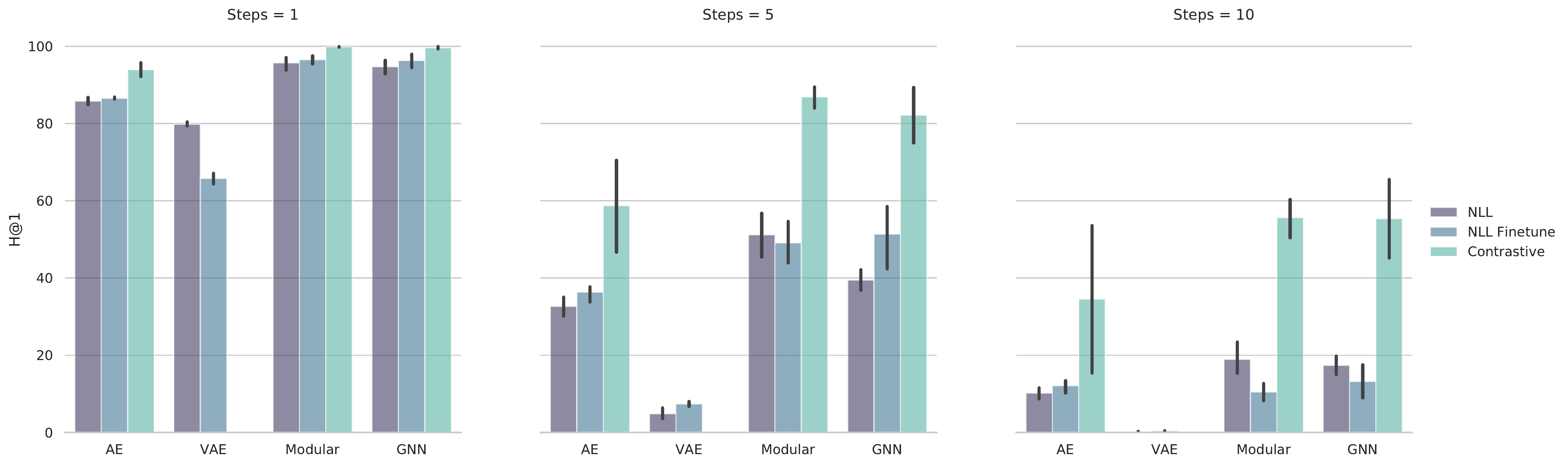}
    \includegraphics[width=15cm]{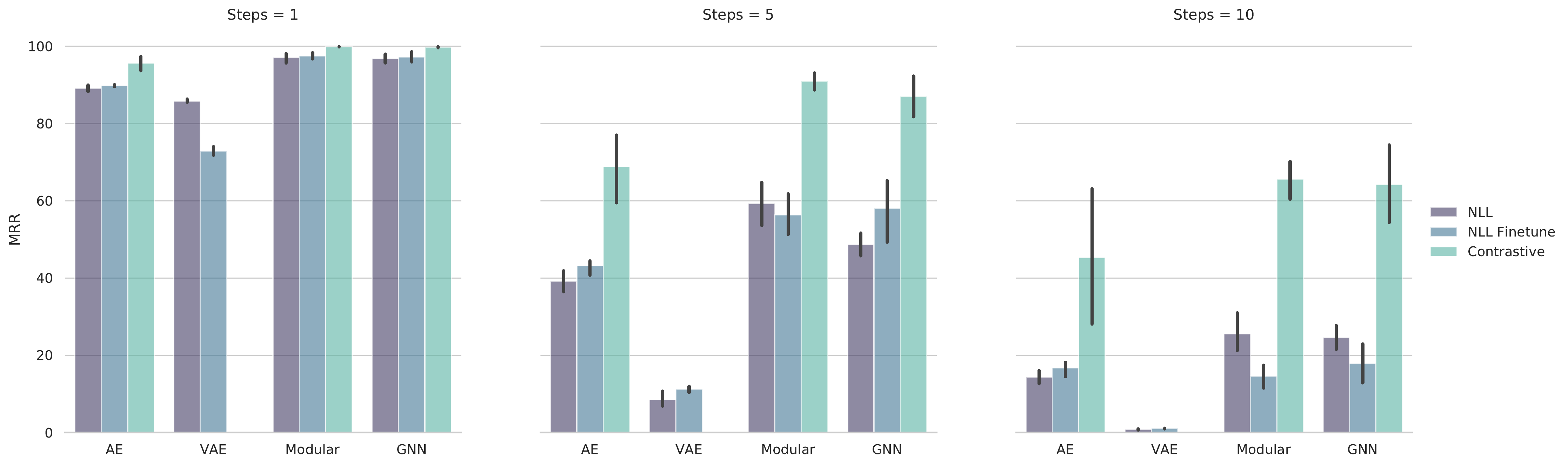}
    \includegraphics[width=15cm]{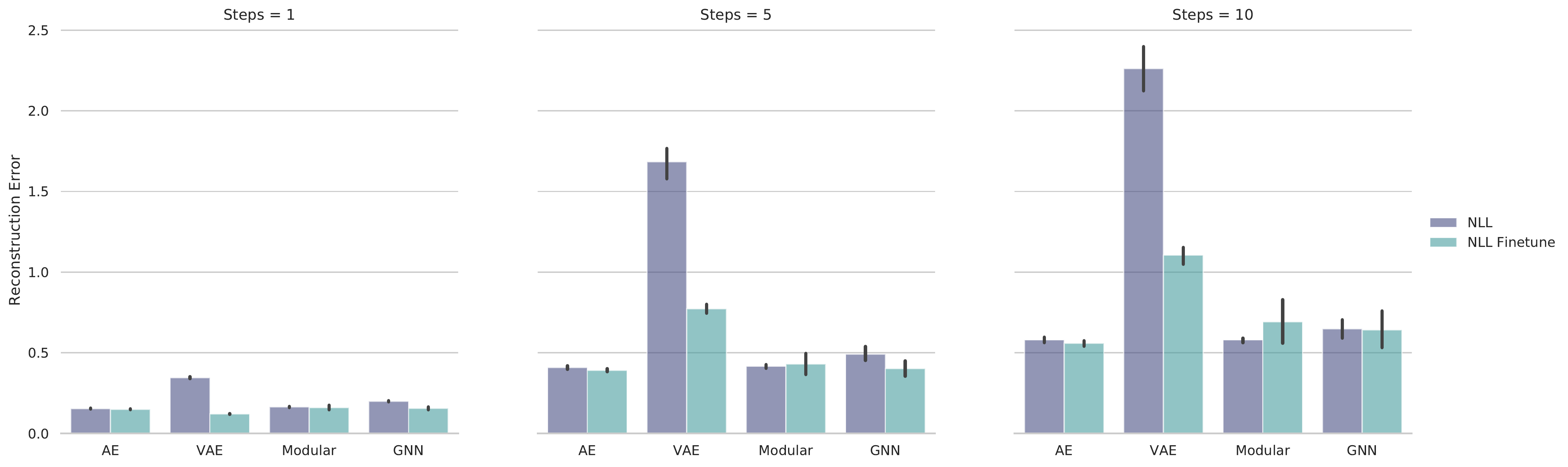}
    \caption{Hits at Rank 1 (H@1), Mean Reciprocal Rank (MRR) \textit{(higher is better)} and Reconstruction Error \textit{(lower is better)} for different models and training losses for 1, 5 and 10 step prediction for the Observed Physics environment Zero Shot setting with 5 objects.}
    \label{fig:physics_observed_5obj_0shot}
\end{figure}

\begin{figure}
    \centering
    \includegraphics[width=15cm]{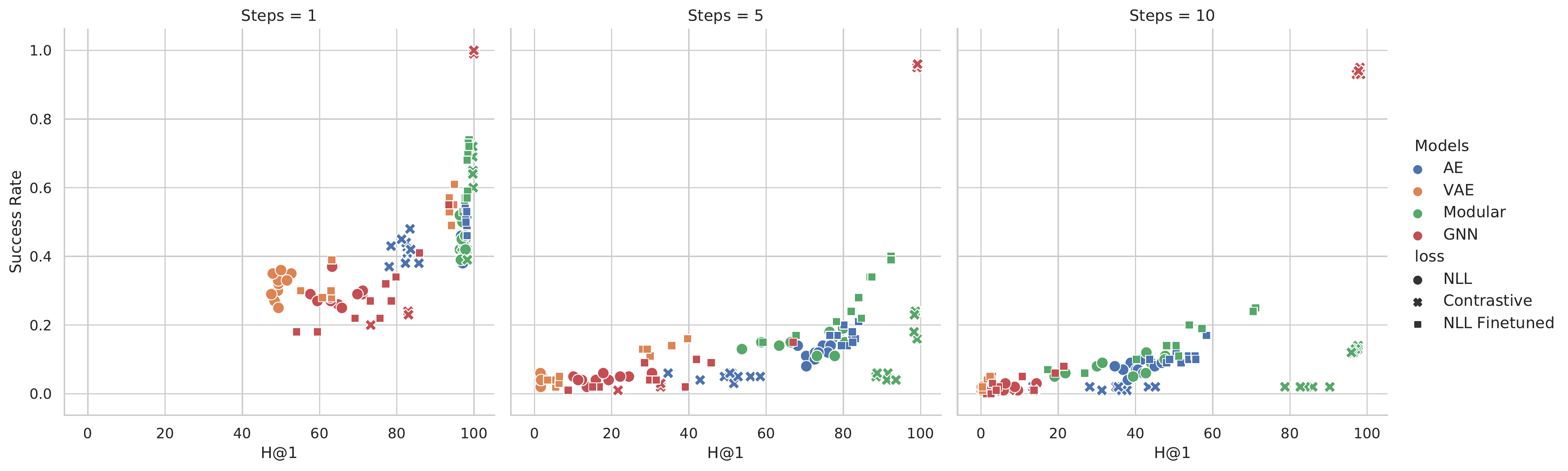}
    \includegraphics[width=15cm]{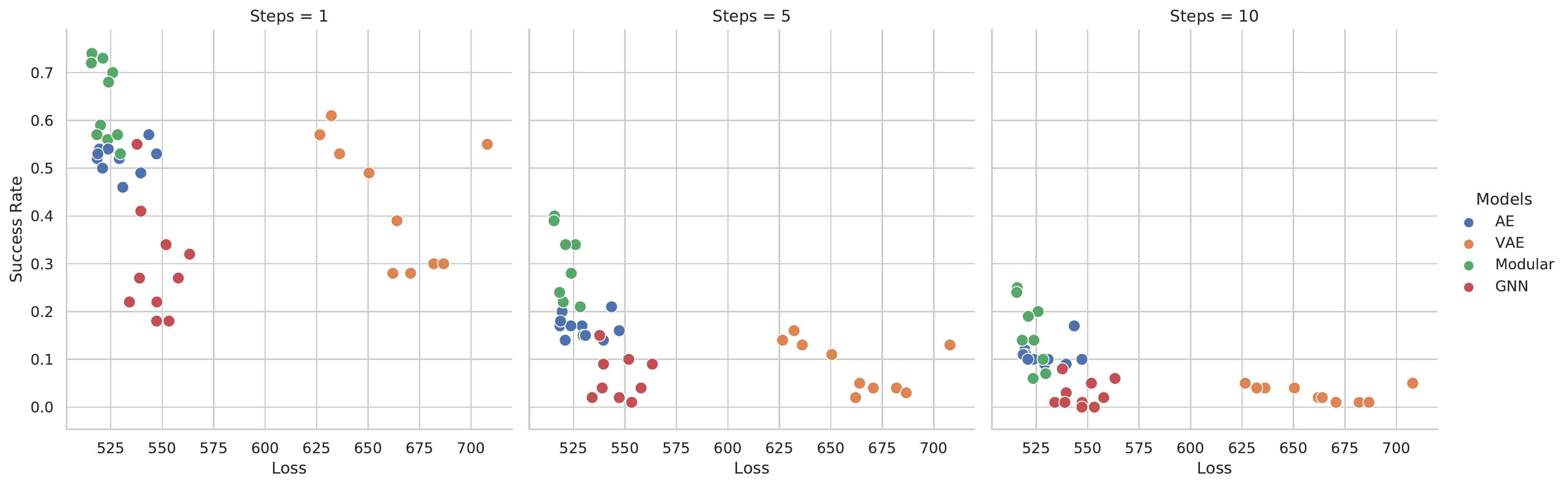}
    \includegraphics[width=15cm]{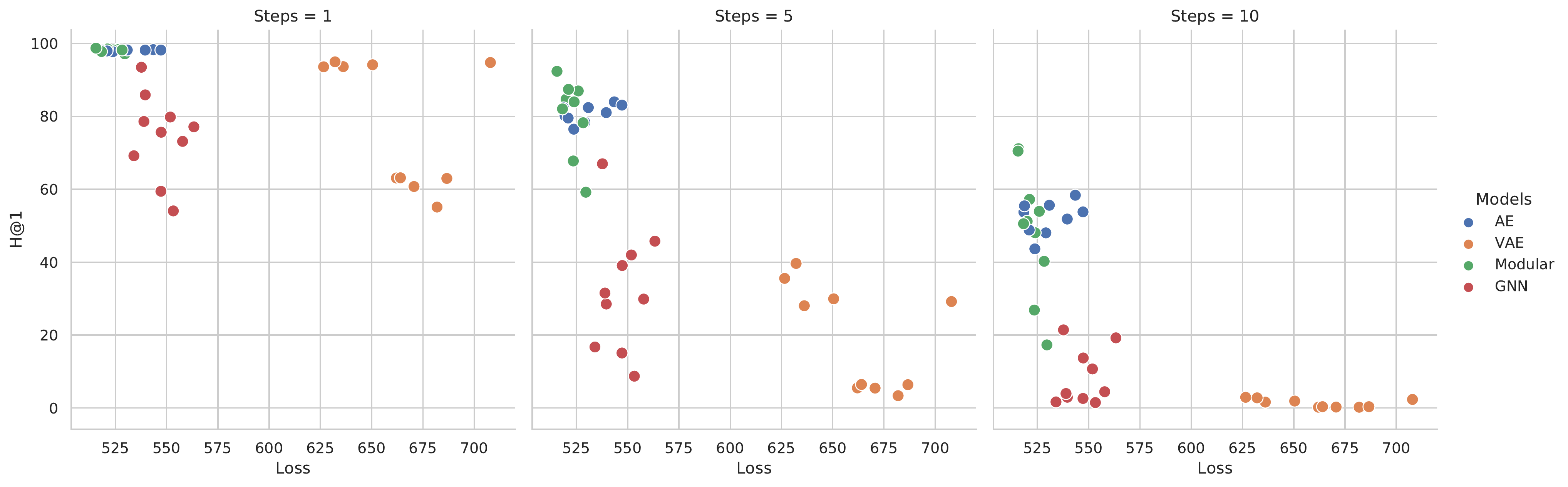}
    \caption{Plots for Observed Physics Environment with 3 objects. Note that (a) the ranking metric (H@1) does not always correspond to good RL performance. In particular, the ranking metric is good across multiple steps but RL performance generally degrades. (b) and (c) Ranking metric and success rate seem to be a bit negatively correlated with test loss.}
    \label{fig:physics_observed_scatter_3}
\end{figure}

\begin{figure}
    \centering
    \includegraphics[width=15cm]{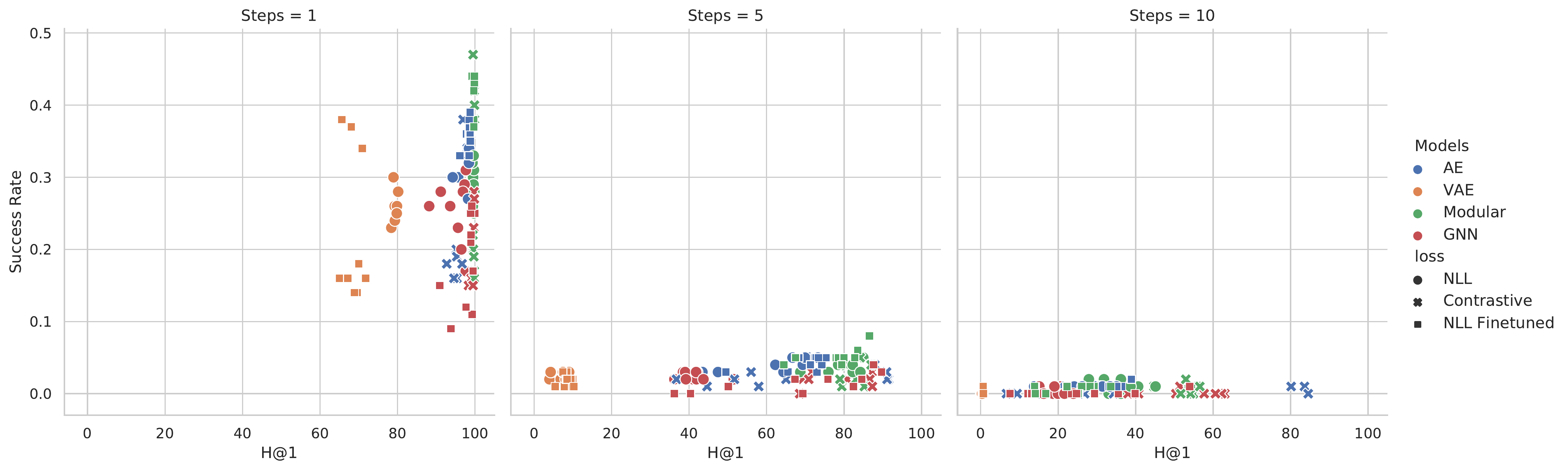}
    \includegraphics[width=15cm]{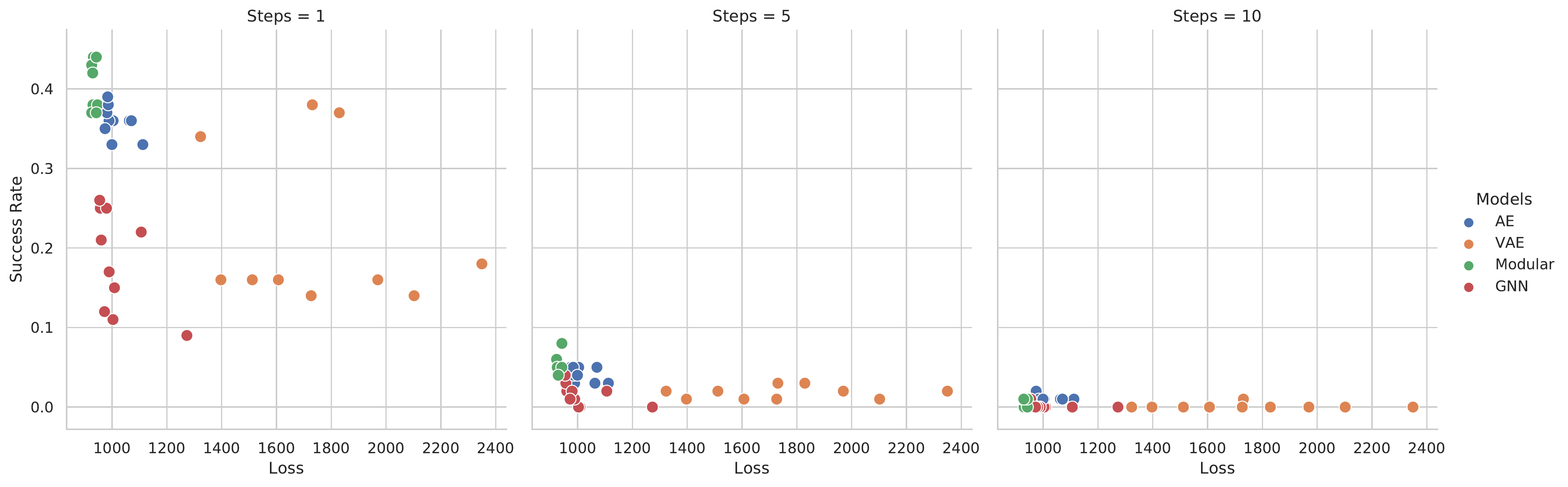}
    \includegraphics[width=15cm]{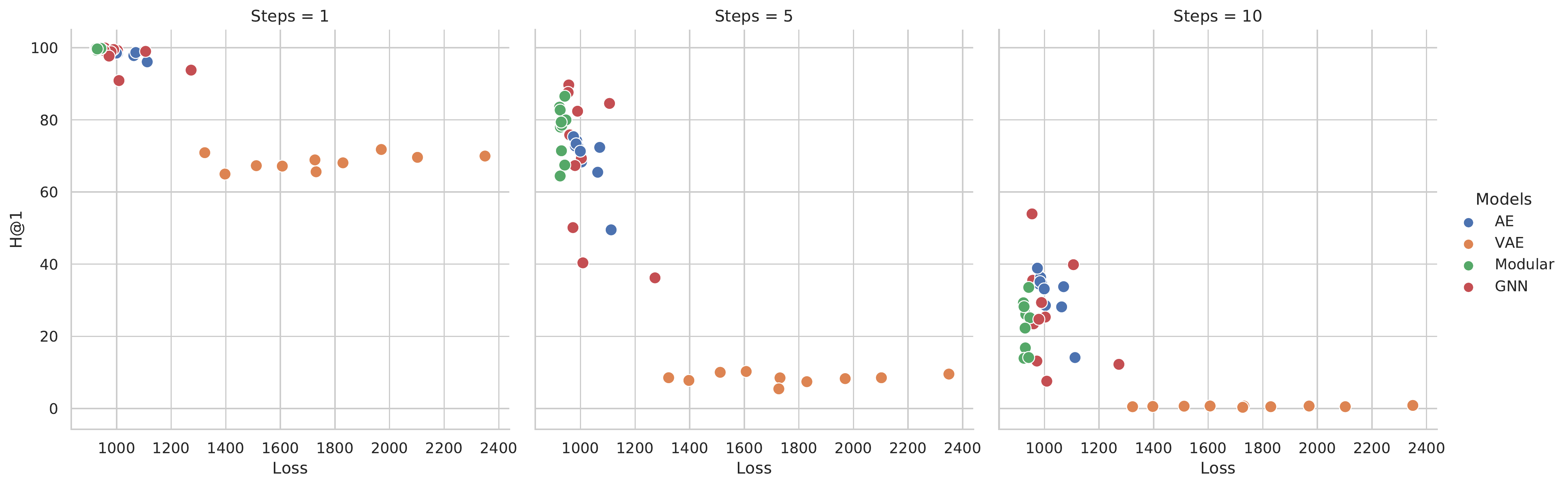}
    \caption{Plots for Observed Physics Environment with 5 objects. Note that (a) the ranking metric (H@1) does not always correspond to good RL performance. In particular, the ranking metric is good across multiple steps but RL performance generally degrades. (b) and (c) Ranking metric and success rate seem to be a bit negatively correlated with test loss.}
    \label{fig:physics_observed_scatter_5}
\end{figure}

\begin{table}[hbt!]
\scriptsize	
\centering 
\renewcommand{\arraystretch}{1.2}
\setlength{\tabcolsep}{2pt}
{\begin{tabular}{c|c|ccc|ccc|ccc}
\toprule
& & \multicolumn{3}{c}{\underline{1 Step}} & \multicolumn{3}{c}{\underline{5 Steps}} & \multicolumn{3}{c}{\underline{10 Steps}} \\
& Model & H@1 & MRR & Rec. & H@1 & MRR & Rec. & H@1 & MRR & Rec. \\
\hline
\multirowcell{4}{\textbf{NLL}}
 & AE & \highlight{\g{97.23}{0.37}} & \highlight{\g{98.23}{0.28}} & \highlight{\g{0.04}{0.0}} & \highlight{\g{72.78}{2.5}} & \highlight{\g{77.74}{2.14}} & \highlight{\g{0.1}{0.01}} & \highlight{\g{40.46}{3.48}} & \highlight{\g{47.4}{3.37}} & \highlight{\g{0.22}{0.01}}  \\
 & GNN & \g{64.86}{4.43} & \g{73.39}{4.08} & \g{0.11}{0.01} & \g{17.73}{6.15} & \g{25.44}{7.73} & \g{0.33}{0.05} & \g{6.4}{3.51} & \g{11.05}{5.17} & \g{0.44}{0.06}  \\
 & Modular & \g{97.13}{0.55} & \g{98.22}{0.42} & \highlight{\g{0.04}{0.0}} & \g{70.7}{9.01} & \g{76.46}{7.95} & \g{0.13}{0.02} & \g{36.66}{9.88} & \g{44.25}{10.14} & \g{0.26}{0.03}  \\
 & VAE & \g{49.52}{1.51} & \g{58.98}{1.79} & \g{0.25}{0.02} & \g{1.7}{0.13} & \g{3.4}{0.16} & \g{1.0}{0.11} & \g{0.16}{0.03} & \g{0.56}{0.06} & \g{1.18}{0.14}  \\
 \hline
 \multirowcell{4}{\textbf{NLL} \\ \textbf{Finetuned}}
 & AE & \g{98.08}{0.2} & \g{98.81}{0.15} & \highlight{\g{0.03}{0.0}} & \g{80.95}{2.2} & \g{84.54}{1.86} & \highlight{\g{0.07}{0.0}} & \highlight{\g{51.98}{4.12}} & \highlight{\g{57.96}{3.84}} & \highlight{\g{0.16}{0.01}}  \\
 & GNN & \g{74.64}{11.03} & \g{78.88}{10.19} & \g{0.04}{0.0} & \g{32.43}{16.24} & \g{39.39}{17.45} & \g{0.14}{0.05} & \g{8.23}{7.15} & \g{12.03}{9.29} & \g{0.28}{0.07}  \\
 & Modular & \highlight{\g{98.16}{0.49}} & \highlight{\g{99.0}{0.33}} & \highlight{\g{0.03}{0.0}} & \highlight{\g{81.49}{10.07}} & \highlight{\g{86.17}{8.66}} & \highlight{\g{0.07}{0.02}} & \g{48.7}{16.19} & \g{56.48}{16.41} & \g{0.17}{0.04}  \\
 & VAE & \g{77.61}{16.75} & \g{83.27}{13.68} & \g{0.04}{0.0} & \g{18.96}{13.9} & \g{25.5}{17.07} & \g{0.29}{0.08} & \g{1.3}{1.08} & \g{2.87}{1.96} & \g{0.51}{0.07}  \\
  \hline
 \multirowcell{3}{\textbf{Contrastive}}
  & AE & \g{82.11}{2.22} & \g{88.5}{1.61} &  - & \g{50.0}{6.43} & \g{65.2}{5.04} &  - & \g{34.36}{8.42} & \g{51.22}{8.17} &  -  \\
 & GNN & \g{93.86}{9.59} & \g{95.99}{6.42} &  - & \g{78.28}{32.39} & \g{82.29}{26.85} &  - & \g{72.06}{39.58} & \g{75.46}{35.65} &  -  \\
 & Modular & \highlight{\g{98.73}{1.04}} & \highlight{\g{99.31}{0.58}} &  - & \highlight{\g{94.7}{4.2}} & \highlight{\g{97.02}{2.38}} &  - & \highlight{\g{90.6}{6.87}} & \highlight{\g{94.45}{4.08}} &  -  \\
\hline
\end{tabular}
}
\caption{Hits at Rank 1 (H@1), Mean Reciprocal Rank (MRR) \textit{(higher is better)} and Reconstruction Error \textit{(lower is better)} for different models and training losses for 1, 5 and 10 step prediction for the Observed Physics environment setting with 3 objects.}
\label{tab:physics_observed_3obj}
\end{table}

\begin{table}[hbt!]
\scriptsize	
\centering 
\renewcommand{\arraystretch}{1.2}
\setlength{\tabcolsep}{2pt}
{\begin{tabular}{c|c|ccc|ccc|ccc}
\toprule
& & \multicolumn{3}{c}{\underline{1 Step}} & \multicolumn{3}{c}{\underline{5 Steps}} & \multicolumn{3}{c}{\underline{10 Steps}} \\
& Model & H@1 & MRR & Rec. & H@1 & MRR & Rec. & H@1 & MRR & Rec. \\
\hline
\multirowcell{4}{\textbf{NLL}}
 & AE & \g{97.77}{1.45} & \g{98.38}{1.05} & \highlight{\g{0.08}{0.01}} & \g{63.88}{9.77} & \g{69.55}{9.0} & \highlight{\g{0.25}{0.03}} & \g{27.18}{7.09} & \g{33.6}{7.71} & \highlight{\g{0.45}{0.03}}  \\
 & GNN & \g{95.13}{3.02} & \g{96.95}{2.24} & \g{0.19}{0.02} & \g{41.49}{3.95} & \g{50.63}{3.93} & \g{0.47}{0.05} & \g{19.28}{2.57} & \g{26.59}{3.06} & \g{0.63}{0.07}  \\
 & Modular & \highlight{\g{99.57}{0.16}} & \highlight{\g{99.73}{0.12}} & \g{0.09}{0.0} & \highlight{\g{79.14}{4.89}} & \highlight{\g{84.06}{4.09}} & \g{0.28}{0.01} & \highlight{\g{35.68}{6.99}} & \highlight{\g{43.82}{7.55}} & \g{0.48}{0.02}  \\
 & VAE & \g{79.35}{0.48} & \g{84.38}{0.4} & \g{0.34}{0.01} & \g{6.18}{1.76} & \g{10.68}{2.25} & \g{1.62}{0.1} & \g{0.28}{0.09} & \g{0.97}{0.22} & \g{2.21}{0.13}  \\
 \hline
 \multirowcell{4}{\textbf{NLL} \\ \textbf{Finetuned}}
 & AE & \g{98.29}{0.77} & \g{98.78}{0.53} & \g{0.07}{0.01} & \g{69.58}{7.23} & \g{74.59}{6.45} & \highlight{\g{0.2}{0.02}} & \highlight{\g{31.75}{6.64}} & \highlight{\g{38.22}{6.97}} & \highlight{\g{0.39}{0.02}}  \\
 & GNN & \g{97.71}{2.81} & \g{98.43}{2.13} & \g{0.07}{0.0} & \g{68.36}{18.69} & \g{73.78}{17.13} & \highlight{\g{0.2}{0.05}} & \g{26.52}{13.33} & \g{32.94}{14.63} & \g{0.46}{0.13}  \\
 & Modular & \highlight{\g{99.65}{0.2}} & \highlight{\g{99.77}{0.14}} & \highlight{\g{0.06}{0.0}} & \highlight{\g{77.21}{6.81}} & \highlight{\g{82.08}{5.83}} & \g{0.21}{0.04} & \g{23.15}{6.27} & \g{29.24}{7.12} & \g{0.53}{0.12}  \\
 & VAE & \g{68.44}{2.1} & \g{74.52}{1.6} & \g{0.09}{0.0} & \g{8.42}{1.32} & \g{12.42}{1.8} & \g{0.75}{0.03} & \g{0.58}{0.14} & \g{1.34}{0.28} & \g{1.07}{0.05}  \\
  \hline
 \multirowcell{3}{\textbf{Contrastive}}
 & AE & \g{96.12}{1.73} & \g{97.71}{1.12} &  - & \g{67.36}{20.12} & \g{76.98}{15.6} &  - & \g{44.65}{32.39} & \g{55.38}{29.98} &  -  \\
 & GNN & \g{99.28}{0.53} & \g{99.6}{0.31} &  - & \g{78.85}{7.5} & \g{84.81}{6.21} &  - & \g{50.1}{9.94} & \g{60.25}{10.11} &  -  \\
 & Modular & \highlight{\g{99.71}{0.13}} & \highlight{\g{99.84}{0.08}} &  - & \highlight{\g{84.3}{2.84}} & \highlight{\g{89.35}{2.26}} &  - & \highlight{\g{52.36}{4.02}} & \highlight{\g{63.28}{4.28}} &  -  \\
\hline
\end{tabular}
}
    \caption{Hits at Rank 1 (H@1), Mean Reciprocal Rank (MRR) \textit{(higher is better)} and Reconstruction Error \textit{(lower is better)} for different models and training losses for 1, 5 and 10 step prediction for the Observed Physics environment setting with 5 objects.}
\label{tab:physics_observed_5obj}
\end{table}

\begin{table}[hbt!]
\scriptsize	
\centering 
\renewcommand{\arraystretch}{1.2}
\setlength{\tabcolsep}{2pt}
{\begin{tabular}{c|c|ccc|ccc|ccc}
\toprule
& & \multicolumn{3}{c}{\underline{1 Step}} & \multicolumn{3}{c}{\underline{5 Steps}} & \multicolumn{3}{c}{\underline{10 Steps}} \\
& Model & H@1 & MRR & Rec. & H@1 & MRR & Rec. & H@1 & MRR & Rec. \\
\hline
\multirowcell{4}{\textbf{NLL}}
 & AE & \g{65.69}{1.93} & \g{73.4}{1.66} & \highlight{\g{0.12}{0.0}} & \g{17.98}{0.95} & \g{25.84}{1.15} & \highlight{\g{0.3}{0.01}} & \g{6.56}{0.6} & \g{11.64}{0.98} & \highlight{\g{0.39}{0.02}}  \\
 & GNN & \g{62.27}{3.7} & \g{70.16}{3.5} & \g{0.15}{0.01} & \g{19.32}{1.64} & \g{26.2}{2.14} & \g{0.34}{0.02} & \g{8.87}{1.35} & \g{14.09}{2.03} & \g{0.42}{0.02}  \\
 & Modular & \highlight{\g{75.23}{2.69}} & \highlight{\g{82.73}{2.01}} & \highlight{\g{0.12}{0.0}} & \highlight{\g{24.93}{2.64}} & \highlight{\g{33.96}{3.08}} & \g{0.31}{0.01} & \highlight{\g{10.39}{1.67}} & \highlight{\g{16.71}{2.31}} & \highlight{\g{0.39}{0.01}}  \\
 & VAE & \g{52.83}{1.98} & \g{61.68}{1.85} & \g{0.28}{0.01} & \g{1.96}{0.16} & \g{3.92}{0.26} & \g{0.88}{0.05} & \g{0.19}{0.04} & \g{0.62}{0.03} & \g{1.0}{0.07}  \\
 \hline
 \multirowcell{4}{\textbf{NLL} \\ \textbf{Finetuned}}
 & AE & \highlight{\g{95.35}{1.13}} & \highlight{\g{97.02}{0.75}} & \highlight{\g{0.06}{0.0}} & \highlight{\g{40.92}{7.81}} & \highlight{\g{49.77}{7.94}} & \g{0.21}{0.02} & \highlight{\g{9.41}{4.36}} & \highlight{\g{13.92}{5.64}} & \g{0.35}{0.03}  \\
 & GNN & \g{74.19}{5.88} & \g{80.08}{5.04} & \g{0.07}{0.0} & \g{20.13}{8.28} & \g{26.32}{9.43} & \highlight{\g{0.16}{0.01}} & \g{2.3}{2.97} & \g{3.94}{4.06} & \highlight{\g{0.25}{0.02}}  \\
 & Modular & \g{94.92}{1.84} & \g{96.79}{1.24} & \g{0.07}{0.0} & \g{27.62}{6.53} & \g{34.7}{7.51} & \g{0.21}{0.02} & \g{2.52}{1.21} & \g{4.16}{1.74} & \g{0.32}{0.03}  \\
 & VAE & \g{49.65}{4.14} & \g{59.58}{3.92} & \g{0.07}{0.0} & \g{7.82}{1.04} & \g{12.3}{1.48} & \g{0.25}{0.03} & \g{0.83}{0.16} & \g{2.05}{0.29} & \g{0.36}{0.04}  \\
  \hline
 \multirowcell{3}{\textbf{Contrastive}}
 & AE & \g{89.77}{3.3} & \g{94.0}{2.11} &  - & \g{37.57}{9.15} & \g{53.53}{8.72} &  - & \g{13.87}{7.64} & \g{26.54}{10.18} &  -  \\
 & GNN & \g{89.58}{5.13} & \g{93.4}{3.42} &  - & \g{40.33}{10.2} & \g{50.14}{10.19} &  - & \g{17.74}{6.99} & \g{25.67}{8.26} &  -  \\
 & Modular & \highlight{\g{96.55}{3.09}} & \highlight{\g{97.96}{1.96}} &  - & \highlight{\g{62.15}{12.59}} & \highlight{\g{71.49}{11.61}} &  - & \highlight{\g{31.02}{10.94}} & \highlight{\g{42.39}{12.6}} &  -  \\
\hline
\end{tabular}
}
    \caption{Hits at Rank 1 (H@1), Mean Reciprocal Rank (MRR) \textit{(higher is better)} and Reconstruction Error \textit{(lower is better)} for different models and training losses for 1, 5 and 10 step prediction for the Unobserved Physics environment setting with 3 objects.}
\label{tab:physics_unobserved_3obj}
\end{table}

\begin{table}[hbt!]
\scriptsize	
\centering 
\renewcommand{\arraystretch}{1.2}
\setlength{\tabcolsep}{2pt}
{\begin{tabular}{c|c|ccc|ccc|ccc}
\toprule
& & \multicolumn{3}{c}{\underline{1 Step}} & \multicolumn{3}{c}{\underline{5 Steps}} & \multicolumn{3}{c}{\underline{10 Steps}} \\
& Model & H@1 & MRR & Rec. & H@1 & MRR & Rec. & H@1 & MRR & Rec. \\
\hline
\multirowcell{4}{\textbf{NLL}}
 & AE & \g{89.49}{0.68} & \g{92.15}{0.62} & \highlight{\g{0.15}{0.0}} & \g{37.78}{1.7} & \g{45.92}{1.78} & \highlight{\g{0.35}{0.01}} & \g{15.04}{1.74} & \g{21.52}{2.21} & \highlight{\g{0.46}{0.01}}  \\
 & GNN & \g{95.76}{2.07} & \g{97.3}{1.53} & \g{0.17}{0.01} & \g{49.46}{1.98} & \g{57.92}{2.14} & \g{0.42}{0.04} & \g{28.5}{2.54} & \g{36.75}{2.99} & \g{0.54}{0.05}  \\
 & Modular & \highlight{\g{98.19}{1.26}} & \highlight{\g{98.93}{0.81}} & \highlight{\g{0.15}{0.01}} & \highlight{\g{57.51}{5.46}} & \highlight{\g{66.3}{5.16}} & \g{0.37}{0.03} & \highlight{\g{31.67}{4.13}} & \highlight{\g{40.84}{4.55}} & \g{0.49}{0.04}  \\
 & VAE & \g{77.21}{3.91} & \g{81.44}{3.54} & \g{0.33}{0.01} & \g{26.01}{2.63} & \g{32.41}{2.79} & \g{0.89}{0.04} & \g{9.18}{1.03} & \g{13.74}{1.25} & \g{1.18}{0.07}  \\
 \hline
 \multirowcell{4}{\textbf{NLL} \\ \textbf{Finetuned}}
 & AE & \g{95.79}{0.58} & \g{97.27}{0.43} & \g{0.11}{0.0} & \g{27.77}{1.72} & \g{35.19}{1.88} & \g{0.22}{0.01} & \g{3.73}{0.45} & \g{5.92}{0.57} & \g{0.32}{0.02}  \\
 & GNN & \g{99.04}{0.72} & \g{99.43}{0.44} & \highlight{\g{0.1}{0.0}} & \highlight{\g{58.45}{7.06}} & \highlight{\g{65.86}{6.56}} & \highlight{\g{0.2}{0.01}} & \highlight{\g{10.34}{3.74}} & \highlight{\g{15.38}{4.81}} & \highlight{\g{0.28}{0.01}}  \\
 & Modular & \highlight{\g{99.87}{0.05}} & \highlight{\g{99.93}{0.03}} & \highlight{\g{0.1}{0.0}} & \g{42.15}{9.03} & \g{49.12}{9.4} & \g{0.22}{0.01} & \g{4.35}{2.47} & \g{6.32}{3.35} & \g{0.36}{0.04}  \\
 & VAE & \g{65.67}{5.74} & \g{72.42}{5.11} & \g{0.11}{0.0} & \g{15.62}{3.52} & \g{20.41}{4.25} & \g{0.3}{0.02} & \g{3.55}{1.5} & \g{5.57}{2.15} & \g{0.42}{0.03}  \\
  \hline
 \multirowcell{3}{\textbf{Contrastive}}
 & AE & \g{97.23}{0.93} & \g{98.38}{0.54} &  - & \g{56.62}{5.66} & \g{68.68}{4.46} &  - & \g{22.86}{5.52} & \g{35.88}{6.53} &  -  \\
 & GNN & \g{99.67}{0.21} & \g{99.81}{0.12} &  - & \g{82.52}{6.75} & \g{86.9}{5.33} &  - & \highlight{\g{55.12}{11.8}} & \highlight{\g{63.04}{10.74}} &  -  \\
 & Modular & \highlight{\g{99.8}{0.14}} & \highlight{\g{99.89}{0.08}} &  - & \highlight{\g{82.98}{3.25}} & \highlight{\g{87.44}{2.72}} &  - & \g{50.92}{4.73} & \g{59.51}{4.65} &  -  \\
\hline
\end{tabular}
}
    \caption{Hits at Rank 1 (H@1), Mean Reciprocal Rank (MRR) \textit{(higher is better)} and Reconstruction Error \textit{(lower is better)} for different models and training losses for 1, 5 and 10 step prediction for the Unobserved Physics environment setting with 5 objects.}
\label{tab:physics_unobserved_5obj}
\end{table}

\begin{table}[hbt!]
\scriptsize	
\centering 
\renewcommand{\arraystretch}{1.2}
\setlength{\tabcolsep}{2pt}
{\begin{tabular}{c|c|ccc|ccc|ccc}
\toprule
& & \multicolumn{3}{c}{\underline{1 Step}} & \multicolumn{3}{c}{\underline{5 Steps}} & \multicolumn{3}{c}{\underline{10 Steps}} \\
& Model & H@1 & MRR & Rec. & H@1 & MRR & Rec. & H@1 & MRR & Rec. \\
\hline
\multirowcell{4}{\textbf{NLL}}
 & AE & \highlight{\g{99.0}{0.1}} & \highlight{\g{99.44}{0.06}} & \highlight{\g{0.04}{0.0}} & \highlight{\g{95.0}{0.41}} & \highlight{\g{96.81}{0.31}} & \highlight{\g{0.06}{0.0}} & \highlight{\g{84.54}{1.5}} & \highlight{\g{89.24}{1.2}} & \highlight{\g{0.1}{0.01}}  \\
 & GNN & \g{70.68}{4.95} & \g{79.43}{4.13} & \g{0.11}{0.02} & \g{23.82}{7.74} & \g{33.28}{9.14} & \g{0.27}{0.05} & \g{10.11}{5.08} & \g{16.65}{6.91} & \g{0.36}{0.06}  \\
 & Modular & \g{98.03}{0.22} & \g{98.84}{0.17} & \g{0.05}{0.0} & \g{88.12}{2.65} & \g{91.8}{2.2} & \g{0.08}{0.01} & \g{68.6}{9.01} & \g{76.12}{7.96} & \g{0.12}{0.02}  \\
 & VAE & \g{53.12}{2.76} & \g{63.42}{2.58} & \g{0.21}{0.01} & \g{2.2}{0.24} & \g{4.53}{0.4} & \g{0.61}{0.05} & \g{0.2}{0.04} & \g{0.82}{0.08} & \g{0.76}{0.07}  \\
 \hline
 \multirowcell{4}{\textbf{NLL} \\ \textbf{Finetuned}}
 & AE & \highlight{\g{99.24}{0.08}} & \highlight{\g{99.59}{0.05}} & \highlight{\g{0.04}{0.0}} & \highlight{\g{96.73}{0.37}} & \highlight{\g{98.02}{0.22}} & \highlight{\g{0.05}{0.0}} & \highlight{\g{90.56}{1.0}} & \highlight{\g{93.72}{0.69}} & \highlight{\g{0.07}{0.0}}  \\
 & GNN & \g{75.16}{12.45} & \g{79.97}{11.99} & \g{0.05}{0.0} & \g{34.78}{14.54} & \g{42.8}{16.43} & \g{0.11}{0.04} & \g{12.76}{7.38} & \g{17.88}{9.47} & \g{0.21}{0.07}  \\
 & Modular & \g{98.76}{0.15} & \g{99.35}{0.09} & \highlight{\g{0.04}{0.0}} & \g{91.3}{2.18} & \g{94.54}{1.63} & \g{0.06}{0.0} & \g{66.7}{7.96} & \g{75.15}{7.17} & \g{0.1}{0.01}  \\
 & VAE & \g{68.53}{13.05} & \g{76.55}{10.71} & \g{0.05}{0.0} & \g{21.38}{10.49} & \g{29.76}{12.17} & \g{0.18}{0.03} & \g{1.72}{1.0} & \g{3.85}{1.66} & \g{0.29}{0.04}  \\
  \hline
 \multirowcell{3}{\textbf{Contrastive}}
 & AE & \g{77.67}{10.51} & \g{86.21}{7.08} &  - & \g{53.49}{23.05} & \g{68.11}{17.57} &  - & \g{43.65}{26.31} & \g{59.13}{21.53} &  -  \\
 & GNN & \g{84.94}{8.08} & \g{90.1}{5.62} &  - & \g{42.88}{28.35} & \g{51.75}{24.48} &  - & \g{28.06}{35.18} & \g{34.19}{32.68} &  -  \\
 & Modular & \highlight{\g{88.42}{6.43}} & \highlight{\g{93.32}{4.48}} &  - & \highlight{\g{71.54}{17.3}} & \highlight{\g{83.07}{12.72}} &  - & \highlight{\g{66.07}{20.34}} & \highlight{\g{79.36}{15.62}} &  -  \\
\hline
\end{tabular}
}
    \caption{Hits at Rank 1 (H@1), Mean Reciprocal Rank (MRR) \textit{(higher is better)} and Reconstruction Error \textit{(lower is better)} for different models and training losses for 1, 5 and 10 step prediction for the FixedUnobserved Physics environment setting with 3 objects.}
\label{tab:physics_fixedunobserved_3obj}
\end{table}

\begin{table}[hbt!]
\scriptsize	
\centering 
\renewcommand{\arraystretch}{1.2}
\setlength{\tabcolsep}{2pt}
{\begin{tabular}{c|c|ccc|ccc|ccc}
\toprule
& & \multicolumn{3}{c}{\underline{1 Step}} & \multicolumn{3}{c}{\underline{5 Steps}} & \multicolumn{3}{c}{\underline{10 Steps}} \\
& Model & H@1 & MRR & Rec. & H@1 & MRR & Rec. & H@1 & MRR & Rec. \\
\hline
\multirowcell{4}{\textbf{NLL}}
 & AE & \g{98.51}{0.14} & \g{98.85}{0.09} & \highlight{\g{0.07}{0.0}} & \g{84.3}{1.58} & \g{87.34}{1.18} & \g{0.16}{0.0} & \g{58.14}{2.86} & \g{64.57}{2.51} & \g{0.26}{0.01}  \\
 & GNN & \g{95.61}{2.72} & \g{97.13}{2.01} & \g{0.13}{0.02} & \g{53.89}{11.58} & \g{62.05}{10.9} & \g{0.32}{0.05} & \g{28.64}{11.56} & \g{36.96}{12.45} & \g{0.44}{0.07}  \\
 & Modular & \highlight{\g{99.48}{0.25}} & \highlight{\g{99.63}{0.21}} & \highlight{\g{0.07}{0.0}} & \highlight{\g{94.13}{1.31}} & \highlight{\g{95.54}{1.24}} & \highlight{\g{0.15}{0.01}} & \highlight{\g{78.18}{3.17}} & \highlight{\g{82.76}{2.95}} & \highlight{\g{0.23}{0.02}}  \\
 & VAE & \g{74.2}{0.99} & \g{78.72}{0.98} & \g{0.23}{0.01} & \g{22.18}{0.93} & \g{27.69}{0.91} & \g{0.63}{0.04} & \g{5.68}{0.88} & \g{9.0}{1.0} & \g{0.83}{0.06}  \\
 \hline
 \multirowcell{4}{\textbf{NLL} \\ \textbf{Finetuned}}
 & AE & \g{99.18}{0.16} & \g{99.37}{0.11} & \g{0.07}{0.0} & \g{91.48}{1.86} & \g{93.19}{1.41} & \g{0.12}{0.01} & \highlight{\g{73.28}{3.92}} & \highlight{\g{77.99}{3.36}} & \highlight{\g{0.2}{0.02}}  \\
 & GNN & \g{95.86}{3.39} & \g{97.36}{2.31} & \highlight{\g{0.06}{0.0}} & \g{65.56}{16.0} & \g{71.71}{14.42} & \g{0.13}{0.03} & \g{35.26}{20.76} & \g{41.63}{20.99} & \g{0.26}{0.07}  \\
 & Modular & \highlight{\g{99.85}{0.12}} & \highlight{\g{99.91}{0.08}} & \highlight{\g{0.06}{0.0}} & \highlight{\g{95.04}{1.85}} & \highlight{\g{96.59}{1.39}} & \highlight{\g{0.11}{0.01}} & \g{61.72}{9.28} & \g{68.6}{8.84} & \g{0.23}{0.02}  \\
 & VAE & \g{54.28}{4.29} & \g{62.55}{3.59} & \g{0.07}{0.0} & \g{10.07}{3.42} & \g{14.12}{4.4} & \g{0.28}{0.01} & \g{1.91}{1.13} & \g{3.3}{1.72} & \g{0.4}{0.02}  \\
  \hline
 \multirowcell{3}{\textbf{Contrastive}}
 & AE & \g{92.83}{12.62} & \g{94.9}{10.1} &  - & \g{79.39}{25.3} & \g{85.46}{21.36} &  - & \g{72.04}{26.37} & \g{80.28}{22.74} &  -  \\
 & GNN & \highlight{\g{99.93}{0.11}} & \highlight{\g{99.97}{0.06}} &  - & \g{96.21}{6.69} & \g{97.41}{4.64} &  - & \g{88.83}{18.69} & \g{91.34}{14.86} &  -  \\
 & Modular & \g{99.86}{0.07} & \g{99.93}{0.04} &  - & \highlight{\g{98.36}{0.49}} & \highlight{\g{98.94}{0.42}} &  - & \highlight{\g{93.44}{2.91}} & \highlight{\g{95.63}{1.94}} &  -  \\
\hline
\end{tabular}
}
    \caption{Hits at Rank 1 (H@1), Mean Reciprocal Rank (MRR) \textit{(higher is better)} and Reconstruction Error \textit{(lower is better)} for different models and training losses for 1, 5 and 10 step prediction for the FixedUnobserved Physics environment setting with 5 objects.}
\label{tab:physics_fixedunobserved_5obj}
\end{table}

\begin{table}[hbt!]
\scriptsize	
\centering 
\renewcommand{\arraystretch}{1.2}
\setlength{\tabcolsep}{5pt}
{\begin{tabular}{c|c|cc|cc|cc}
\toprule
& & \multicolumn{2}{c}{\underline{1 Step}} & \multicolumn{2}{c}{\underline{5 Steps}} & \multicolumn{2}{c}{\underline{10 Steps}} \\
& Model & Reward & Success & Reward & Success & Reward & Success \\
\hline
\multirowcell{2}{\textbf{Baselines}} & Random Baseline & -0.37 & 0.22 & -1.26 & 0.01 & -1.78 & 0.00 \\
& Greedy Baseline & 0.00 & 1.00 & -0.00 & 0.99 & -0.01 & 0.98 \\
\hline
\multirowcell{4}{\textbf{NLL}}
 & AE & \g{-0.26}{0.01} & \g{0.44}{0.03} & \g{-0.73}{0.04} & \g{0.12}{0.02} & \g{-1.02}{0.06} & \highlight{\g{0.08}{0.02}} \\
 & GNN & \g{-0.34}{0.02} & \g{0.29}{0.03} & \g{-1.04}{0.07} & \g{0.04}{0.01} & \g{-1.49}{0.1} & \g{0.02}{0.01} \\
 & Modular & \highlight{\g{-0.25}{0.02}} & \highlight{\g{0.46}{0.04}} & \highlight{\g{-0.67}{0.06}} & \highlight{\g{0.15}{0.02}} & \highlight{\g{-0.97}{0.09}} & \highlight{\g{0.08}{0.02}} \\
 & VAE & \g{-0.33}{0.02} & \g{0.32}{0.03} & \g{-1.0}{0.03} & \g{0.04}{0.01} & \g{-1.37}{0.04} & \g{0.01}{0.0} \\
\hline
\multirowcell{4}{\textbf{NLL} \\ \textbf{Finetuned}}
 & AE & \g{-0.22}{0.02} & \g{0.52}{0.03} & \g{-0.62}{0.04} & \g{0.17}{0.02} & \g{-0.9}{0.05} & \g{0.11}{0.02} \\
 & GNN & \g{-0.36}{0.06} & \g{0.3}{0.11} & \g{-1.06}{0.24} & \g{0.06}{0.04} & \g{-1.57}{0.33} & \g{0.03}{0.03} \\
 & Modular & \highlight{\g{-0.16}{0.04}} & \highlight{\g{0.64}{0.08}} & \highlight{\g{-0.48}{0.11}} & \highlight{\g{0.27}{0.09}} & \highlight{\g{-0.79}{0.17}} & \highlight{\g{0.15}{0.06}} \\
 & VAE & \g{-0.26}{0.07} & \g{0.43}{0.13} & \g{-0.85}{0.2} & \g{0.08}{0.05} & \g{-1.28}{0.21} & \g{0.03}{0.02} \\
\hline
\multirowcell{3}{\textbf{Contrastive}}
 & AE & \g{-0.27}{0.02} & \g{0.42}{0.03} & \g{-0.97}{0.04} & \g{0.05}{0.01} & \g{-1.44}{0.05} & \g{0.02}{0.0} \\
 & GNN & \highlight{\g{-0.11}{0.17}} & \highlight{\g{0.77}{0.36}} & \highlight{\g{-0.36}{0.52}} & \highlight{\g{0.68}{0.43}} & \highlight{\g{-0.5}{0.72}} & \highlight{\g{0.66}{0.43}} \\
 & Modular & \g{-0.2}{0.07} & \g{0.54}{0.13} & \g{-0.76}{0.23} & \g{0.13}{0.08} & \g{-1.06}{0.28} & \g{0.07}{0.05} \\
\hline
\end{tabular}
}
\caption{Negative Return \textit{(lower is better)} and Success Rate \textit{(higher is better)} for different models and training losses for 1, 5 and 10 step prediction for the Observed Physics environment setting with 3 objects.}
\label{tab:physics_observed_3obj_rl}
\end{table}

\begin{table}[hbt!]
\scriptsize	
\centering 
\renewcommand{\arraystretch}{1.2}
\setlength{\tabcolsep}{5pt}
{\begin{tabular}{c|c|cc|cc|cc}
\toprule
& & \multicolumn{2}{c}{\underline{1 Step}} & \multicolumn{2}{c}{\underline{5 Steps}} & \multicolumn{2}{c}{\underline{10 Steps}} \\
& Model & Reward & Success & Reward & Success & Reward & Success \\
\hline
\multirowcell{2}{\textbf{Baselines}} & Random Baseline & -0.22 & 0.22 & -0.87 & 0.00 & -1.31 & 0.00 \\
& Greedy Baseline & 0.00 & 1.00 & -0.00 & 0.99 & -0.00 & 0.98 \\
\hline
\multirowcell{4}{\textbf{NLL}}
 & AE & \highlight{\g{-0.2}{0.01}} & \highlight{\g{0.32}{0.03}} & \g{-0.66}{0.04} & \highlight{\g{0.04}{0.01}} & \g{-1.04}{0.04} & \highlight{\g{0.01}{0.0}} \\
 & GNN & \g{-0.22}{0.02} & \g{0.25}{0.04} & \g{-0.74}{0.04} & \g{0.02}{0.0} & \g{-1.16}{0.06} & \g{0.0}{0.0} \\
 & Modular & \g{-0.21}{0.01} & \g{0.29}{0.03} & \highlight{\g{-0.65}{0.02}} & \highlight{\g{0.04}{0.01}} & \highlight{\g{-1.02}{0.03}} & \highlight{\g{0.01}{0.01}} \\
 & VAE & \g{-0.22}{0.01} & \g{0.26}{0.02} & \g{-0.73}{0.02} & \g{0.02}{0.0} & \g{-1.08}{0.02} & \g{0.0}{0.0} \\
\hline
\multirowcell{4}{\textbf{NLL} \\ \textbf{Finetuned}}
 & AE & \g{-0.18}{0.01} & \g{0.36}{0.02} & \g{-0.62}{0.02} & \g{0.04}{0.01} & \highlight{\g{-1.0}{0.02}} & \highlight{\g{0.01}{0.0}} \\
 & GNN & \g{-0.26}{0.03} & \g{0.18}{0.06} & \g{-0.84}{0.12} & \g{0.02}{0.01} & \g{-1.27}{0.18} & \g{0.0}{0.0} \\
 & Modular & \highlight{\g{-0.17}{0.01}} & \highlight{\g{0.41}{0.03}} & \highlight{\g{-0.6}{0.03}} & \highlight{\g{0.05}{0.01}} & \g{-1.02}{0.04} & \highlight{\g{0.01}{0.0}} \\
 & VAE & \g{-0.23}{0.04} & \g{0.22}{0.1} & \g{-0.79}{0.08} & \g{0.02}{0.01} & \g{-1.17}{0.1} & \g{0.0}{0.0} \\
\hline
\multirowcell{3}{\textbf{Contrastive}}
 & AE & \g{-0.22}{0.03} & \g{0.24}{0.08} & \g{-0.74}{0.04} & \highlight{\g{0.02}{0.01}} & \g{-1.09}{0.07} & \g{0.0}{0.0} \\
 & GNN & \g{-0.23}{0.03} & \g{0.22}{0.07} & \g{-0.74}{0.06} & \highlight{\g{0.02}{0.01}} & \g{-1.08}{0.05} & \g{0.0}{0.0} \\
 & Modular & \highlight{\g{-0.21}{0.05}} & \highlight{\g{0.28}{0.12}} & \highlight{\g{-0.68}{0.09}} & \highlight{\g{0.02}{0.02}} & \highlight{\g{-1.01}{0.09}} & \highlight{\g{0.01}{0.01}} \\
\hline
\end{tabular}
}
\caption{Negative Return \textit{(lower is better)} and Success Rate \textit{(higher is better)} for different models and training losses for 1, 5 and 10 step prediction for the Observed Physics environment setting with 5 objects.}
\label{tab:physics_observed_5obj_rl}
\end{table}

\begin{table}[hbt!]
\scriptsize	
\centering 
\renewcommand{\arraystretch}{1.2}
\setlength{\tabcolsep}{5pt}
{\begin{tabular}{c|c|cc|cc|cc}
\toprule
& & \multicolumn{2}{c}{\underline{1 Step}} & \multicolumn{2}{c}{\underline{5 Steps}} & \multicolumn{2}{c}{\underline{10 Steps}} \\
& Model & Reward & Success & Reward & Success & Reward & Success \\
\hline
\multirowcell{2}{\textbf{Baselines}} & Random Baseline & -0.37 & 0.22 & -1.26 & 0.01 & -1.78 & 0.00 \\
& Greedy Baseline & 0.00 & 1.00 & -0.00 & 0.99 & -0.01 & 0.98 \\
\hline
\multirowcell{4}{\textbf{NLL}}
 & AE & \highlight{\g{-0.31}{0.01}} & \highlight{\g{0.35}{0.02}} & \g{-0.95}{0.02} & \highlight{\g{0.06}{0.01}} & \g{-1.39}{0.04} & \highlight{\g{0.02}{0.01}} \\
 & GNN & \g{-0.36}{0.01} & \g{0.27}{0.01} & \g{-1.13}{0.02} & \g{0.03}{0.0} & \g{-1.64}{0.03} & \g{0.01}{0.0} \\
 & Modular & \g{-0.32}{0.01} & \g{0.34}{0.01} & \highlight{\g{-0.94}{0.02}} & \highlight{\g{0.06}{0.01}} & \highlight{\g{-1.36}{0.04}} & \highlight{\g{0.02}{0.01}} \\
 & VAE & \g{-0.37}{0.01} & \g{0.26}{0.03} & \g{-1.06}{0.06} & \g{0.04}{0.01} & \g{-1.48}{0.05} & \g{0.01}{0.0} \\
\hline
\multirowcell{4}{\textbf{NLL} \\ \textbf{Finetuned}}
 & AE & \highlight{\g{-0.26}{0.02}} & \highlight{\g{0.44}{0.03}} & \highlight{\g{-0.83}{0.06}} & \highlight{\g{0.08}{0.02}} & \highlight{\g{-1.23}{0.07}} & \highlight{\g{0.03}{0.01}} \\
 & GNN & \g{-0.37}{0.02} & \g{0.26}{0.03} & \g{-1.13}{0.05} & \g{0.03}{0.01} & \g{-1.71}{0.1} & \g{0.01}{0.01} \\
 & Modular & \g{-0.27}{0.03} & \g{0.43}{0.05} & \g{-0.89}{0.07} & \g{0.07}{0.02} & \g{-1.32}{0.09} & \g{0.02}{0.01} \\
 & VAE & \g{-0.39}{0.03} & \g{0.22}{0.03} & \g{-1.18}{0.07} & \g{0.02}{0.01} & \g{-1.61}{0.09} & \g{0.0}{0.0} \\
\hline
\multirowcell{3}{\textbf{Contrastive}}
 & AE & \highlight{\g{-0.31}{0.02}} & \g{0.36}{0.04} & \highlight{\g{-0.96}{0.04}} & \highlight{\g{0.05}{0.01}} & \highlight{\g{-1.36}{0.05}} & \highlight{\g{0.01}{0.01}} \\
 & GNN & \g{-0.39}{0.02} & \g{0.2}{0.04} & \g{-1.22}{0.06} & \g{0.02}{0.01} & \g{-1.64}{0.04} & \g{0.0}{0.0} \\
 & Modular & \highlight{\g{-0.31}{0.03}} & \highlight{\g{0.37}{0.06}} & \g{-1.07}{0.07} & \g{0.04}{0.01} & \g{-1.54}{0.09} & \highlight{\g{0.01}{0.0}} \\
\hline
\end{tabular}
}
\caption{Negative Return \textit{(lower is better)} and Success Rate \textit{(higher is better)} for different models and training losses for 1, 5 and 10 step prediction for the Unobserved Physics environment setting with 3 objects.}
\label{tab:physics_unobserved_3obj_rl}
\end{table}

\begin{table}[hbt!]
\scriptsize	
\centering 
\renewcommand{\arraystretch}{1.2}
\setlength{\tabcolsep}{5pt}
{\begin{tabular}{c|c|cc|cc|cc}
\toprule
& & \multicolumn{2}{c}{\underline{1 Step}} & \multicolumn{2}{c}{\underline{5 Steps}} & \multicolumn{2}{c}{\underline{10 Steps}} \\
& Model & Reward & Success & Reward & Success & Reward & Success \\
\hline
\multirowcell{2}{\textbf{Baselines}} & Random Baseline & -0.22 & 0.22 & -0.87 & 0.00 & -1.31 & 0.00 \\
& Greedy Baseline & 0.00 & 1.00 & -0.00 & 0.99 & -0.00 & 0.98 \\
\hline
\multirowcell{4}{\textbf{NLL}}
 & AE & \g{-0.22}{0.01} & \g{0.26}{0.03} & \g{-0.74}{0.03} & \highlight{\g{0.02}{0.01}} & \g{-1.14}{0.03} & \g{0.0}{0.0} \\
 & GNN & \g{-0.22}{0.01} & \g{0.25}{0.02} & \g{-0.76}{0.02} & \highlight{\g{0.02}{0.01}} & \g{-1.19}{0.03} & \g{0.0}{0.0} \\
 & Modular & \highlight{\g{-0.21}{0.01}} & \highlight{\g{0.28}{0.03}} & \highlight{\g{-0.7}{0.03}} & \highlight{\g{0.02}{0.0}} & \highlight{\g{-1.08}{0.04}} & \highlight{\g{0.01}{0.0}} \\
 & VAE & \g{-0.23}{0.01} & \g{0.22}{0.02} & \g{-0.78}{0.03} & \highlight{\g{0.02}{0.01}} & \g{-1.18}{0.06} & \g{0.0}{0.0} \\
\hline
\multirowcell{4}{\textbf{NLL} \\ \textbf{Finetuned}}
 & AE & \highlight{\g{-0.18}{0.02}} & \highlight{\g{0.35}{0.04}} & \highlight{\g{-0.59}{0.05}} & \highlight{\g{0.04}{0.01}} & \highlight{\g{-0.96}{0.07}} & \highlight{\g{0.01}{0.0}} \\
 & GNN & \g{-0.23}{0.0} & \g{0.22}{0.02} & \g{-0.8}{0.04} & \g{0.02}{0.01} & \g{-1.28}{0.07} & \g{0.0}{0.0} \\
 & Modular & \g{-0.21}{0.01} & \g{0.28}{0.03} & \g{-0.69}{0.04} & \g{0.03}{0.01} & \g{-1.11}{0.07} & \highlight{\g{0.01}{0.0}} \\
 & VAE & \g{-0.21}{0.05} & \g{0.25}{0.11} & \g{-0.73}{0.13} & \g{0.03}{0.01} & \g{-1.1}{0.13} & \g{0.0}{0.0} \\
\hline
\multirowcell{3}{\textbf{Contrastive}}
 & AE & \g{-0.25}{0.02} & \g{0.2}{0.06} & \g{-0.76}{0.07} & \highlight{\g{0.02}{0.01}} & \g{-1.12}{0.09} & \highlight{\g{0.0}{0.0}} \\
 & GNN & \highlight{\g{-0.21}{0.02}} & \highlight{\g{0.25}{0.06}} & \highlight{\g{-0.71}{0.07}} & \highlight{\g{0.02}{0.0}} & \highlight{\g{-1.08}{0.07}} & \highlight{\g{0.0}{0.0}} \\
 & Modular & \g{-0.24}{0.02} & \g{0.2}{0.04} & \g{-0.76}{0.05} & \highlight{\g{0.02}{0.0}} & \g{-1.12}{0.06} & \highlight{\g{0.0}{0.0}} \\
 \hline
\end{tabular}
}
\caption{Negative Return \textit{(lower is better)} and Success Rate \textit{(higher is better)} for different models and training losses for 1, 5 and 10 step prediction for the Unobserved Physics environment setting with 5 objects.}
\label{tab:physics_unobserved_5obj_rl}
\end{table}

\begin{table}[hbt!]
\scriptsize	
\centering 
\renewcommand{\arraystretch}{1.2}
\setlength{\tabcolsep}{5pt}
{\begin{tabular}{c|c|cc|cc|cc}
\toprule
& & \multicolumn{2}{c}{\underline{1 Step}} & \multicolumn{2}{c}{\underline{5 Steps}} & \multicolumn{2}{c}{\underline{10 Steps}} \\
& Model & Reward & Success & Reward & Success & Reward & Success \\
\hline
\multirowcell{2}{\textbf{Baselines}} & Random Baseline & -0.37 & 0.22 & -1.26 & 0.01 & -1.78 & 0.00 \\
& Greedy Baseline & 0.00 & 1.00 & -0.00 & 0.99 & -0.01 & 0.98 \\
\hline
\multirowcell{4}{\textbf{NLL}}
 & AE & \g{-0.23}{0.01} & \g{0.48}{0.03} & \g{-0.59}{0.03} & \g{0.18}{0.02} & \g{-0.78}{0.05} & \g{0.12}{0.02} \\
 & GNN & \g{-0.34}{0.02} & \g{0.3}{0.03} & \g{-1.03}{0.09} & \g{0.05}{0.01} & \g{-1.51}{0.14} & \g{0.02}{0.01} \\
 & Modular & \highlight{\g{-0.19}{0.02}} & \highlight{\g{0.56}{0.04}} & \highlight{\g{-0.48}{0.06}} & \highlight{\g{0.25}{0.05}} & \highlight{\g{-0.67}{0.08}} & \highlight{\g{0.18}{0.04}} \\
 & VAE & \g{-0.33}{0.01} & \g{0.32}{0.02} & \g{-0.98}{0.04} & \g{0.05}{0.01} & \g{-1.4}{0.04} & \g{0.01}{0.0} \\
\hline
\multirowcell{4}{\textbf{NLL} \\ \textbf{Finetuned}}
 & AE & \g{-0.2}{0.01} & \g{0.54}{0.03} & \g{-0.49}{0.03} & \g{0.23}{0.03} & \g{-0.64}{0.04} & \g{0.18}{0.02} \\
 & GNN & \g{-0.33}{0.07} & \g{0.33}{0.11} & \g{-0.93}{0.22} & \g{0.08}{0.04} & \g{-1.42}{0.3} & \g{0.02}{0.01} \\
 & Modular & \highlight{\g{-0.11}{0.02}} & \highlight{\g{0.73}{0.05}} & \highlight{\g{-0.34}{0.06}} & \highlight{\g{0.38}{0.07}} & \highlight{\g{-0.58}{0.1}} & \highlight{\g{0.24}{0.06}} \\
 & VAE & \g{-0.31}{0.03} & \g{0.34}{0.05} & \g{-0.94}{0.09} & \g{0.06}{0.02} & \g{-1.34}{0.11} & \g{0.02}{0.01} \\
\hline
\multirowcell{3}{\textbf{Contrastive}}
 & AE & \g{-0.09}{0.1} & \g{0.78}{0.23} & \g{-0.38}{0.33} & \g{0.45}{0.31} & \g{-0.55}{0.46} & \g{0.36}{0.31} \\
 & GNN & \g{-0.3}{0.15} & \g{0.4}{0.3} & \g{-0.96}{0.48} & \g{0.21}{0.38} & \g{-1.32}{0.65} & \g{0.19}{0.37} \\
 & Modular & \highlight{\g{-0.07}{0.11}} & \highlight{\g{0.85}{0.24}} & \highlight{\g{-0.24}{0.38}} & \highlight{\g{0.73}{0.41}} & \highlight{\g{-0.34}{0.51}} & \highlight{\g{0.7}{0.43}} \\
\hline
\end{tabular}
}
\caption{Negative Return \textit{(lower is better)} and Success Rate \textit{(higher is better)} for different models and training losses for 1, 5 and 10 step prediction for the FixedUnobserved Physics environment setting with 3 objects.}
\label{tab:physics_fixedunobserved_3obj_rl}
\end{table}

\begin{table}[hbt!]
\scriptsize	
\centering 
\renewcommand{\arraystretch}{1.2}
\setlength{\tabcolsep}{5pt}
{\begin{tabular}{c|c|cc|cc|cc}
\toprule
& & \multicolumn{2}{c}{\underline{1 Step}} & \multicolumn{2}{c}{\underline{5 Steps}} & \multicolumn{2}{c}{\underline{10 Steps}} \\
& Model & Reward & Success & Reward & Success & Reward & Success \\
\hline
\multirowcell{2}{\textbf{Baselines}} & Random Baseline & -0.22 & 0.22 & -0.87 & 0.00 & -1.31 & 0.00 \\
& Greedy Baseline & 0.00 & 1.00 & -0.00 & 0.99 & -0.00 & 0.98 \\
\hline
\multirowcell{4}{\textbf{NLL}}
 & AE & \g{-0.21}{0.01} & \g{0.28}{0.01} & \g{-0.66}{0.02} & \g{0.04}{0.01} & \g{-0.98}{0.03} & \g{0.01}{0.0} \\
 & GNN & \g{-0.23}{0.0} & \g{0.22}{0.02} & \g{-0.76}{0.04} & \g{0.02}{0.0} & \g{-1.17}{0.07} & \g{0.0}{0.0} \\
 & Modular & \highlight{\g{-0.19}{0.01}} & \highlight{\g{0.36}{0.03}} & \highlight{\g{-0.51}{0.03}} & \highlight{\g{0.08}{0.01}} & \highlight{\g{-0.79}{0.05}} & \highlight{\g{0.03}{0.01}} \\
 & VAE & \g{-0.21}{0.03} & \g{0.27}{0.06} & \g{-0.75}{0.09} & \g{0.02}{0.01} & \g{-1.16}{0.1} & \g{0.0}{0.0} \\
\hline
\multirowcell{4}{\textbf{NLL} \\ \textbf{Finetuned}}
 & AE & \g{-0.19}{0.01} & \g{0.35}{0.01} & \g{-0.55}{0.02} & \g{0.06}{0.01} & \g{-0.83}{0.03} & \g{0.02}{0.0} \\
 & GNN & \g{-0.25}{0.03} & \g{0.2}{0.08} & \g{-0.78}{0.14} & \g{0.02}{0.03} & \g{-1.17}{0.19} & \g{0.01}{0.01} \\
 & Modular & \highlight{\g{-0.13}{0.01}} & \highlight{\g{0.52}{0.05}} & \highlight{\g{-0.44}{0.03}} & \highlight{\g{0.11}{0.02}} & \highlight{\g{-0.81}{0.06}} & \highlight{\g{0.03}{0.01}} \\
 & VAE & \g{-0.24}{0.01} & \g{0.19}{0.02} & \g{-0.77}{0.04} & \g{0.02}{0.0} & \g{-1.14}{0.07} & \g{0.0}{0.0} \\
\hline
\multirowcell{3}{\textbf{Contrastive}}
 & AE & \g{-0.13}{0.02} & \g{0.5}{0.07} & \g{-0.51}{0.08} & \g{0.09}{0.04} & \g{-0.81}{0.11} & \g{0.03}{0.01} \\
 & GNN & \g{-0.04}{0.09} & \g{0.84}{0.3} & \g{-0.17}{0.3} & \g{0.74}{0.36} & \g{-0.27}{0.44} & \highlight{\g{0.68}{0.34}} \\
 & Modular & \highlight{\g{-0.0}{0.0}} & \highlight{\g{0.99}{0.02}} & \highlight{\g{-0.06}{0.06}} & \highlight{\g{0.78}{0.2}} & \highlight{\g{-0.14}{0.14}} & \g{0.63}{0.27} \\
 \hline
\end{tabular}
}
\caption{Negative Return \textit{(lower is better)} and Success Rate \textit{(higher is better)} for different models and training losses for 1, 5 and 10 step prediction for the FixedUnobserved Physics environment setting with 5 objects.}
\label{tab:physics_fixedunobserved_5obj_rl}
\end{table}

\begin{table}[hbt!]
\scriptsize	
\centering 
\renewcommand{\arraystretch}{1.2}
\setlength{\tabcolsep}{2pt}
{\begin{tabular}{c|c|ccc|ccc|ccc}
\toprule
& & \multicolumn{3}{c}{\underline{1 Step}} & \multicolumn{3}{c}{\underline{5 Steps}} & \multicolumn{3}{c}{\underline{10 Steps}} \\
& Model & H@1 & MRR & Rec. & H@1 & MRR & Rec. & H@1 & MRR & Rec. \\
\hline
\multirowcell{4}{\textbf{NLL}}
 & AE & \highlight{\g{73.41}{0.63}} & \highlight{\g{78.83}{0.54}} & \highlight{\g{0.1}{0.0}} & \g{26.32}{1.55} & \g{31.97}{1.53} & \highlight{\g{0.31}{0.01}} & \g{10.73}{1.19} & \g{14.47}{1.32} & \g{0.45}{0.02}  \\
 & GNN & \g{57.06}{1.49} & \g{65.9}{1.26} & \g{0.15}{0.01} & \g{12.08}{1.15} & \g{18.33}{1.49} & \g{0.4}{0.04} & \g{3.6}{0.67} & \g{6.97}{1.11} & \g{0.5}{0.05}  \\
 & Modular & \g{71.67}{1.47} & \g{77.8}{1.21} & \g{0.12}{0.0} & \highlight{\g{26.7}{4.2}} & \highlight{\g{33.58}{4.68}} & \highlight{\g{0.31}{0.02}} & \highlight{\g{10.84}{2.89}} & \highlight{\g{15.38}{3.71}} & \highlight{\g{0.42}{0.03}}  \\
 & VAE & \g{43.78}{1.57} & \g{55.48}{1.93} & \g{0.29}{0.02} & \g{1.59}{0.1} & \g{3.18}{0.1} & \g{1.09}{0.12} & \g{0.12}{0.03} & \g{0.46}{0.04} & \g{1.23}{0.14}  \\
 \hline
 \multirowcell{4}{\textbf{NLL} \\ \textbf{Finetuned}}
 & AE & \g{73.78}{1.86} & \g{79.35}{1.73} & \highlight{\g{0.09}{0.01}} & \g{28.37}{1.55} & \g{33.98}{1.65} & \g{0.29}{0.01} & \g{12.4}{0.88} & \g{16.09}{1.04} & \g{0.43}{0.01}  \\
 & GNN & \g{66.42}{7.06} & \g{72.43}{6.67} & \g{0.1}{0.01} & \g{18.42}{7.28} & \g{24.18}{8.74} & \highlight{\g{0.24}{0.02}} & \g{3.24}{1.99} & \g{5.26}{2.78} & \g{0.36}{0.03}  \\
 & Modular & \highlight{\g{77.33}{1.83}} & \highlight{\g{82.91}{1.67}} & \g{0.11}{0.01} & \highlight{\g{35.97}{6.75}} & \highlight{\g{43.71}{7.17}} & \highlight{\g{0.24}{0.01}} & \highlight{\g{15.73}{6.19}} & \highlight{\g{21.38}{7.48}} & \highlight{\g{0.34}{0.02}}  \\
 & VAE & \g{62.8}{14.23} & \g{71.95}{12.27} & \highlight{\g{0.09}{0.01}} & \g{9.77}{6.48} & \g{14.44}{8.41} & \g{0.4}{0.05} & \g{0.69}{0.49} & \g{1.63}{1.02} & \g{0.6}{0.06}  \\
  \hline
 \multirowcell{3}{\textbf{Contrastive}}
 & AE & \g{72.16}{1.31} & \g{78.78}{1.07} &  - & \g{33.23}{5.11} & \g{45.72}{4.26} &  - & \g{18.92}{4.56} & \g{31.02}{5.04} &  -  \\
 & GNN & \highlight{\g{92.19}{5.86}} & \highlight{\g{94.89}{4.05}} &  - & \highlight{\g{61.6}{19.42}} & \highlight{\g{69.77}{17.93}} &  - & \g{44.51}{21.94} & \g{53.42}{22.27} &  -  \\
 & Modular & \g{85.03}{1.73} & \g{88.08}{1.88} &  - & \g{58.26}{3.25} & \g{65.85}{3.68} &  - & \highlight{\g{45.83}{3.15}} & \highlight{\g{54.69}{3.21}} &  -  \\
\hline
\end{tabular}
}
\caption{Hits at Rank 1 (H@1), Mean Reciprocal Rank (MRR) \textit{(higher is better)} and Reconstruction Error \textit{(lower is better)} for different models and training losses for 1, 5 and 10 step prediction for the Observed Physics environment Zero Shot setting with 3 objects.}
\label{tab:physics_observed_3obj_0shot}
\end{table}

\begin{table}[hbt!]
\scriptsize	
\centering 
\renewcommand{\arraystretch}{1.2}
\setlength{\tabcolsep}{2pt}
{\begin{tabular}{c|c|ccc|ccc|ccc}
\toprule
& & \multicolumn{3}{c}{\underline{1 Step}} & \multicolumn{3}{c}{\underline{5 Steps}} & \multicolumn{3}{c}{\underline{10 Steps}} \\
& Model & H@1 & MRR & Rec. & H@1 & MRR & Rec. & H@1 & MRR & Rec. \\
\hline
\multirowcell{4}{\textbf{NLL}}
 & AE & \g{85.81}{1.18} & \g{89.11}{1.05} & \highlight{\g{0.15}{0.0}} & \g{32.64}{2.82} & \g{39.22}{2.92} & \highlight{\g{0.41}{0.01}} & \g{10.2}{1.74} & \g{14.34}{2.0} & \highlight{\g{0.58}{0.02}}  \\
 & GNN & \g{94.67}{2.05} & \g{96.86}{1.35} & \g{0.2}{0.0} & \g{39.49}{3.03} & \g{48.71}{3.35} & \g{0.49}{0.05} & \g{17.39}{2.85} & \g{24.61}{3.59} & \g{0.65}{0.06}  \\
 & Modular & \highlight{\g{95.68}{1.94}} & \highlight{\g{97.14}{1.5}} & \g{0.16}{0.0} & \highlight{\g{51.19}{6.06}} & \highlight{\g{59.25}{6.13}} & \g{0.42}{0.01} & \highlight{\g{18.94}{4.4}} & \highlight{\g{25.58}{5.39}} & \highlight{\g{0.58}{0.02}}  \\
 & VAE & \g{79.8}{0.66} & \g{85.83}{0.54} & \g{0.35}{0.01} & \g{4.83}{1.62} & \g{8.52}{2.25} & \g{1.68}{0.1} & \g{0.23}{0.07} & \g{0.76}{0.18} & \g{2.26}{0.15}  \\
 \hline
 \multirowcell{4}{\textbf{NLL} \\ \textbf{Finetuned}}
 & AE & \g{86.52}{0.32} & \g{89.83}{0.29} & \g{0.15}{0.0} & \g{36.33}{2.52} & \g{43.14}{2.41} & \highlight{\g{0.39}{0.01}} & \g{12.12}{1.92} & \g{16.72}{2.29} & \highlight{\g{0.56}{0.02}}  \\
 & GNN & \g{96.29}{1.99} & \g{97.27}{1.57} & \g{0.15}{0.01} & \highlight{\g{51.4}{9.48}} & \highlight{\g{58.06}{9.27}} & \g{0.4}{0.06} & \highlight{\g{13.22}{5.04}} & \highlight{\g{17.9}{6.0}} & \g{0.64}{0.14}  \\
 & Modular & \highlight{\g{96.5}{1.23}} & \highlight{\g{97.55}{0.94}} & \g{0.16}{0.02} & \g{49.09}{6.16} & \g{56.4}{6.05} & \g{0.43}{0.08} & \g{10.47}{2.52} & \g{14.57}{3.2} & \g{0.69}{0.16}  \\
 & VAE & \g{65.76}{1.61} & \g{72.93}{1.24} & \highlight{\g{0.12}{0.0}} & \g{7.39}{0.77} & \g{11.18}{0.95} & \g{0.77}{0.03} & \g{0.43}{0.06} & \g{1.02}{0.1} & \g{1.11}{0.06}  \\
  \hline
 \multirowcell{3}{\textbf{Contrastive}}
 & AE & \g{93.92}{2.23} & \g{95.64}{2.18} &  - & \g{58.72}{13.26} & \g{68.87}{10.01} &  - & \g{34.58}{21.13} & \g{45.31}{20.27} &  -  \\
 & GNN & \g{99.63}{0.37} & \g{99.8}{0.21} &  - & \g{82.16}{8.14} & \g{87.05}{6.6} &  - & \g{55.34}{12.14} & \g{64.19}{11.66} &  -  \\
 & Modular & \highlight{\g{99.84}{0.11}} & \highlight{\g{99.91}{0.06}} &  - & \highlight{\g{86.88}{3.19}} & \highlight{\g{91.02}{2.51}} &  - & \highlight{\g{55.64}{5.68}} & \highlight{\g{65.58}{5.57}} &  -  \\
\hline
\end{tabular}
}
\caption{Hits at Rank 1 (H@1), Mean Reciprocal Rank (MRR) \textit{(higher is better)} and Reconstruction Error \textit{(lower is better)} for different models and training losses for 1, 5 and 10 step prediction for the Observed Physics environment Zero Shot setting with 5 objects.}
\label{tab:physics_observed_5obj_0shot}
\end{table}

\section{Chemistry Environment}\label{appendix:chemistry}
\subsection{Detailed Setup}
\label{appendix:chemistry_details}
The chemistry environment consists of objects of different shapes and colors. Each object forms a node of a directed acyclic graph. The shapes and positions of the objects are fixed across episodes while the color of each object is sampled from a conditional probability table and depends on the colors of its ancestors. 

Considering a set of $M$ objects: ($X_i = \{s_i, c_i, p_i\} \; \quad \forall i \in \{1,\ldots,M\}$). Here, $s_i$, $c_i$ and $p_i$ denote the shape, color and position of the object respectively. As mentioned previously, the shapes and the positions are fixed across episodes but different for each object. The color of an object is a categorical variable that can take one of the $K$ possible values. To model the CPT we use an MLP for each object, the input to an object's MLP is the current state of each of its parent nodes and the outputs is a probability distribution over $k$ colors out of which one color is sampled for that object. We can control the skewness of the distribution of each object by controlling the initialization of the MLP parameters. It is more hard for a model to learn the correct probability distribution when the distribution is less skewed.

In the chemistry environment, an intervention corresponds to changing the color of an object to a particular color from fixed set of $K$ colors. When an intervention is performed on an object, a new color is sampled for each of its descendants using their respective MLPs as mentioned above. Each object changes its color to the newly sampled color at the same instant.

Note that all our experiments for this environment were run for a setting of 5 objects and 5 colors unless specified otherwise.

\subsection{Ranking Loss and Causal Structure}
\label{appendix:chemistry_static_dynamic}
Initially, our vanilla chemistry environment had objects being initialized at random positions per episode while maintaining a fixed causal graph underneath. We call this setting the \textbf{dynamic} setting. We noticed that in this case, the ranking metrics were very good but performance on downstream RL task as well as qualitative reconstruction was very poor. On further investigation, we reached the conclusion that under this setting, a model could do very well under the ranking metrics without learning the causal structure at all.

If the encoder learns to encode the positions and shapes of different objects, then it already does a great job at ranking. This is because ranking is done with respect to a large buffer of encoded states and since objects are randomly initialized per episode, there is very little probability that two encoded states share the exact same object shapes and positions. Thus, as long as the encoder and the transition function exploit the fact that two encoded states should be close by \textit{iff} they have the same objects in the same positions, then it would do very well on the ranking metrics. Note that in the above argument, the model had a way of ranking well without even learning anything about the edges in the graph, i.e. the structure of interactions between the objects.

To alleviate this problem, we decided to keep the positions of the objects fixed across episodes too. We call this setting the \textbf{static} setting. This means that models will not be able to perform well on ranking metrics by just encoding the positions or shapes of the objects (since they are now shared across episodes). The only way to do well on ranking metrics then is to learn the underlying causal structure. We immediately saw a plummet in ranking metrics that confirmed our suspicions that the models were not able to learn the underlying causal structure.

For a demonstration of the mentioned problem refer to \Cref{fig:static_dynamic}. In the figure we can see that for the dynamic setting, models achieve a much higher score on the ranking metrics (H@1 and MRR) as compared to the static setting while doing much worse on the downstream RL task as compared the static setting. This further reinforces the importance of using downstream RL tasks for evaluation.

This also shows that inferring the causal graph even in the case of small graphs is a complex problem that current models are not able to solve well. We believe that the existence of this suite of environments provides a platform for extensive study of causality in world models.

\subsection{Experimental Results}
\label{appendix:chemistry_insight}
\begin{figure}
    \centering
    \includegraphics[width=15cm]{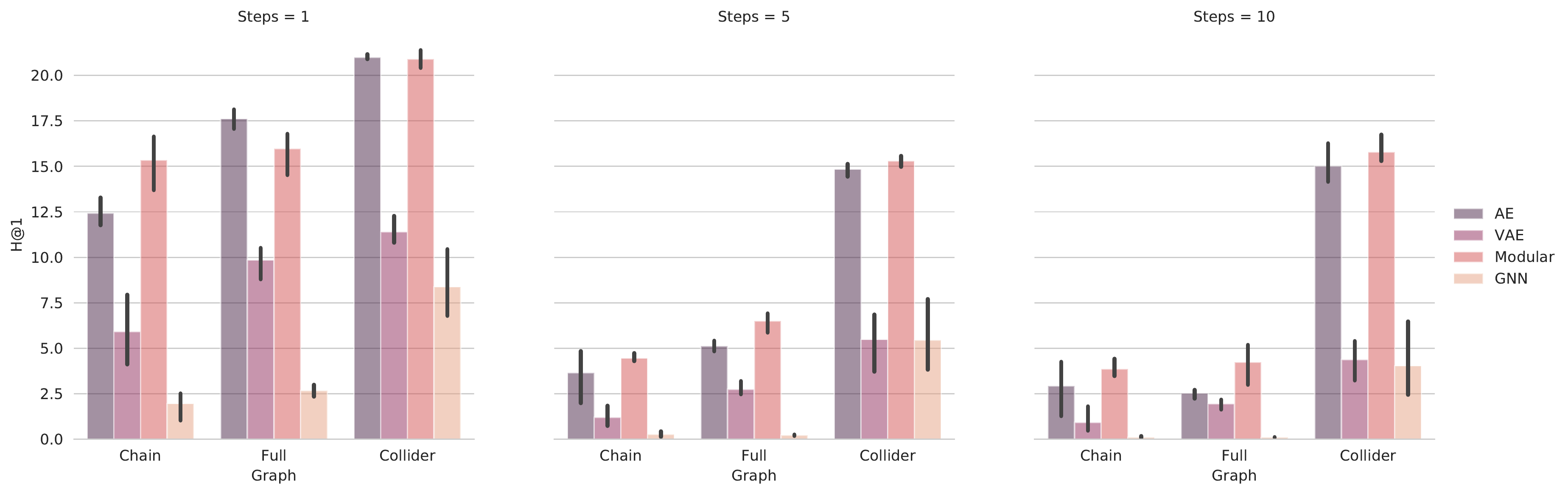}
    \includegraphics[width=15cm]{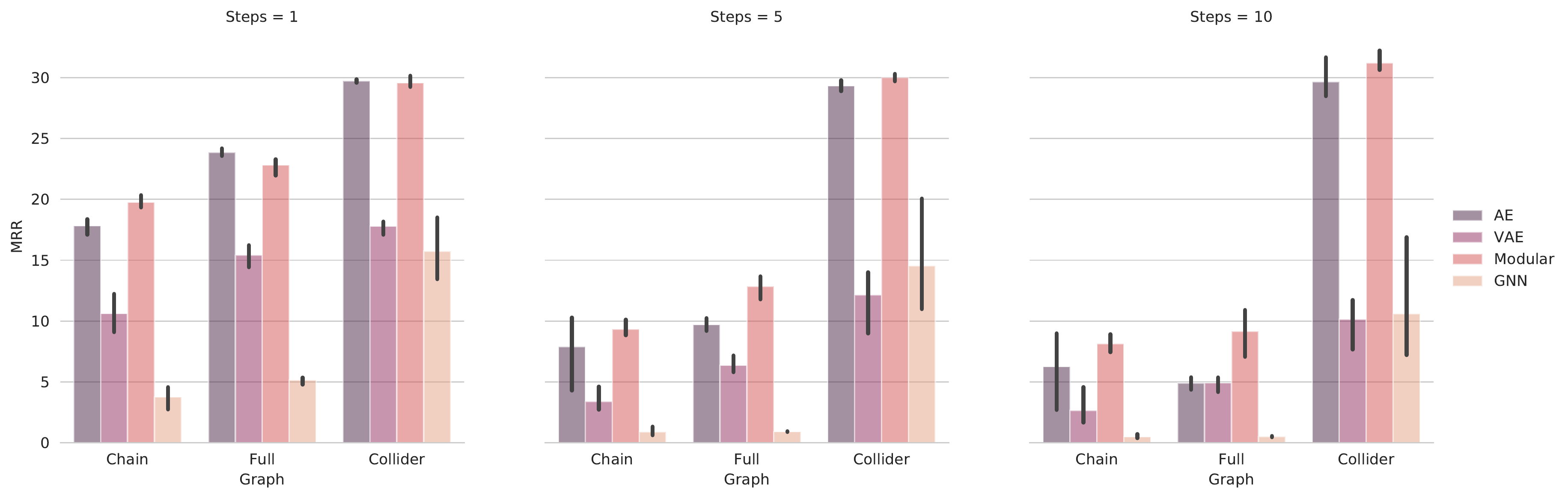}
    \includegraphics[width=15cm]{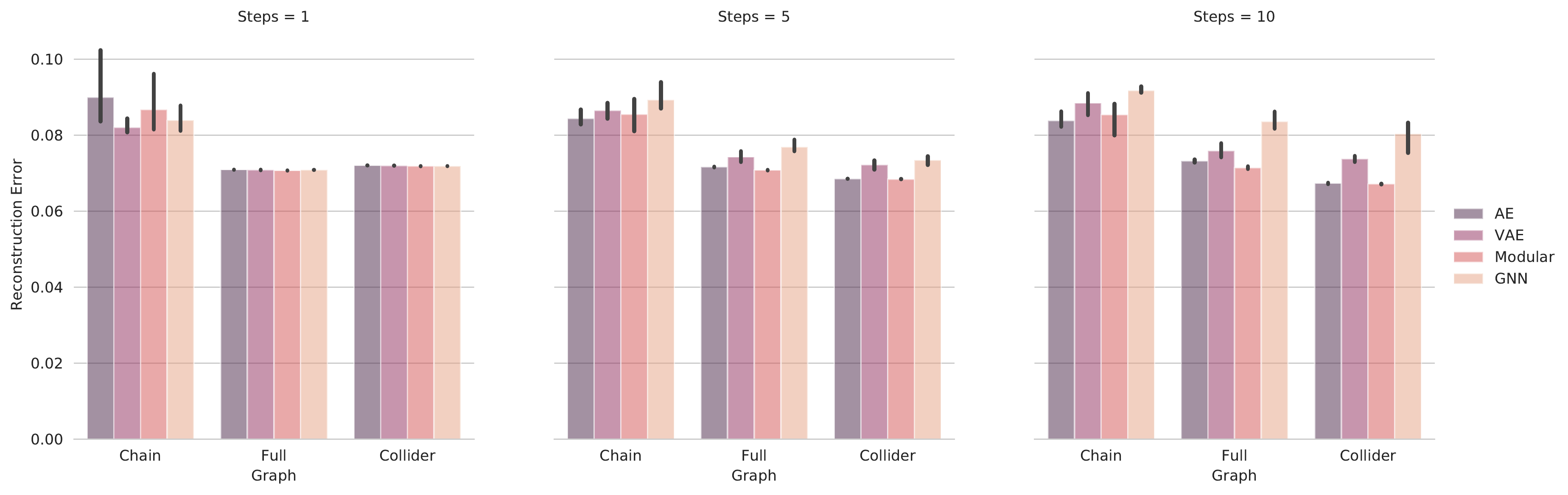}
    \caption{Hits at Rank 1 (H@1), Mean Reciprocal Rank (MRR) \textit{(higher is better)} and Reconstruction Error \textit{(lower is better)} for different models trained using NLL Loss for 1, 5 and 10 step prediction for the vanilla chemistry environment with 5 objects and 5 colors.}
    \label{fig:chemistry_nll_finetune}
\end{figure}

\begin{figure}
    \centering
    \includegraphics[width  = 15cm]{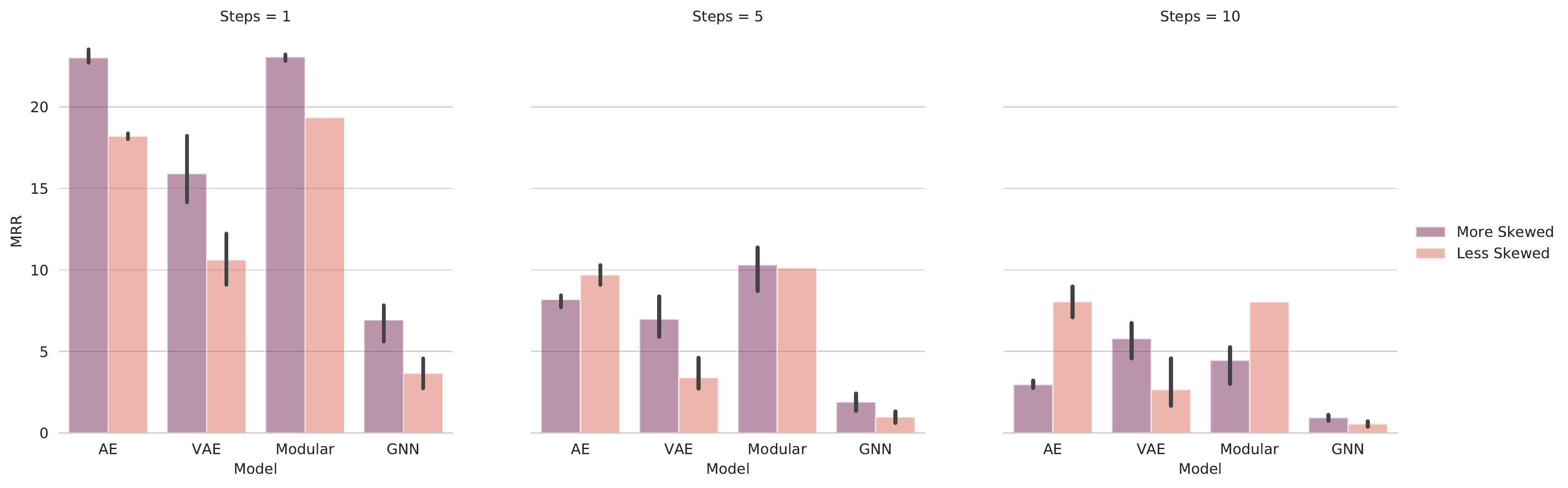}
    
    \caption{H@1 performance of models for data generated at different levels of skewness(stochasticity) for the chain graph. As we see almost all models perform better on more skewed data as the data uncertainty is less on more skewed data as compared to less skewed data.}
    \label{fig:skew_demo}
\end{figure}

\begin{figure}
\vspace{-3\baselineskip}
    \centering 
    \includegraphics[width = 15cm]{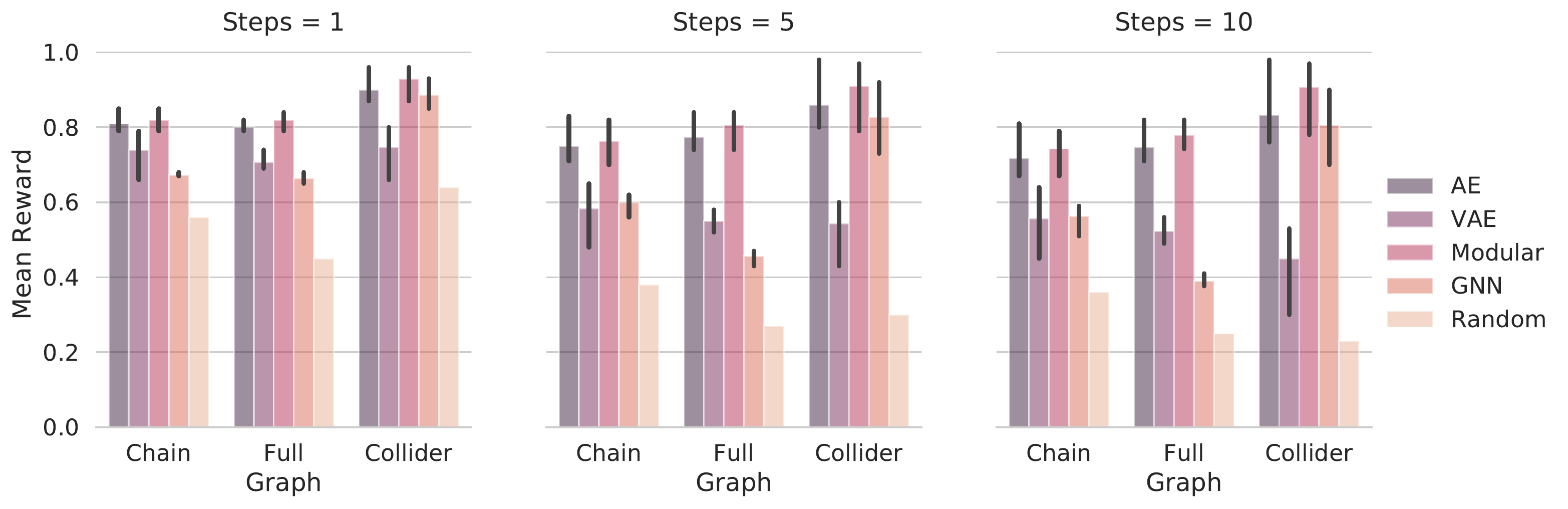}
    \includegraphics[width = 15cm]{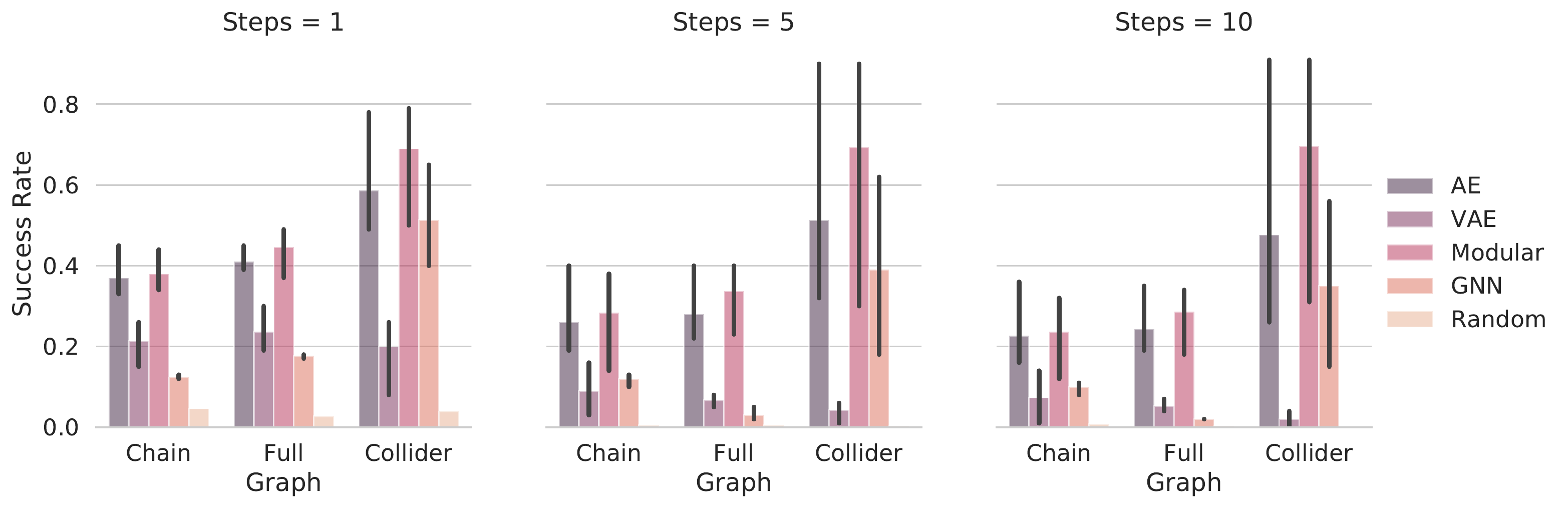}
    
    \caption{Mean reward and success rate for models trained on the chemistry environment with 5 objects and 5 colors. Modular models outperform all other models in almost all cases which shows that introducing structure in the form of modularity is an important inductive bias for learning causal models. }
    \label{fig:chem_reward}
    \vspace{-1\baselineskip}
\end{figure}

\begin{figure}
    \centering
    \includegraphics[width=15cm]{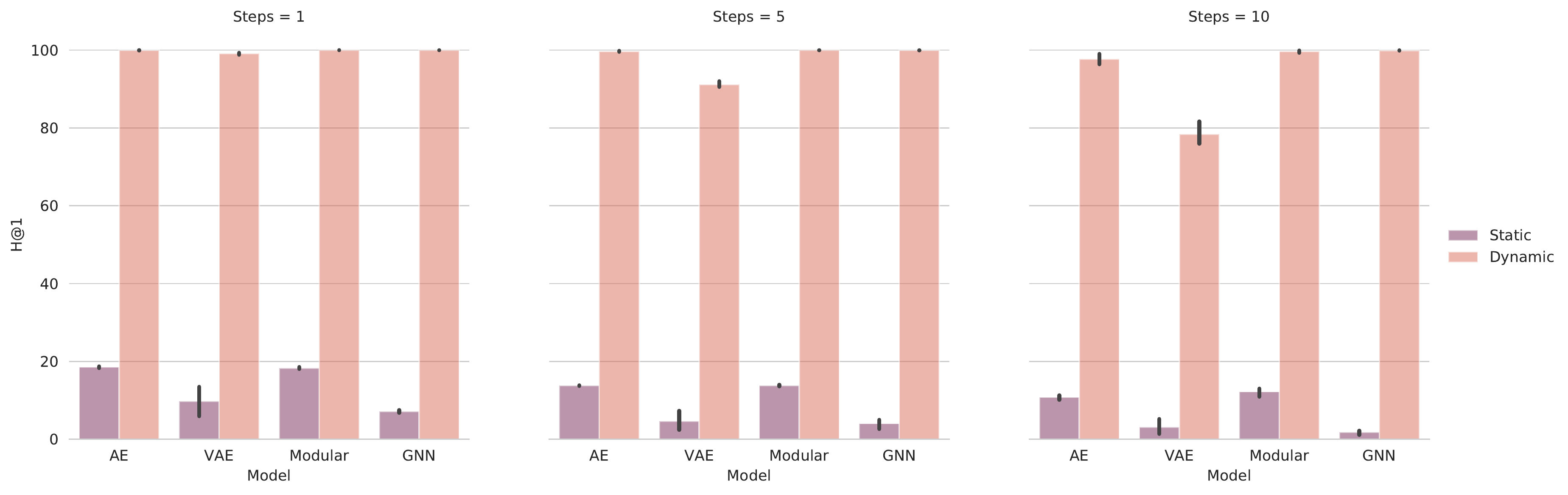}
    \includegraphics[width=15cm]{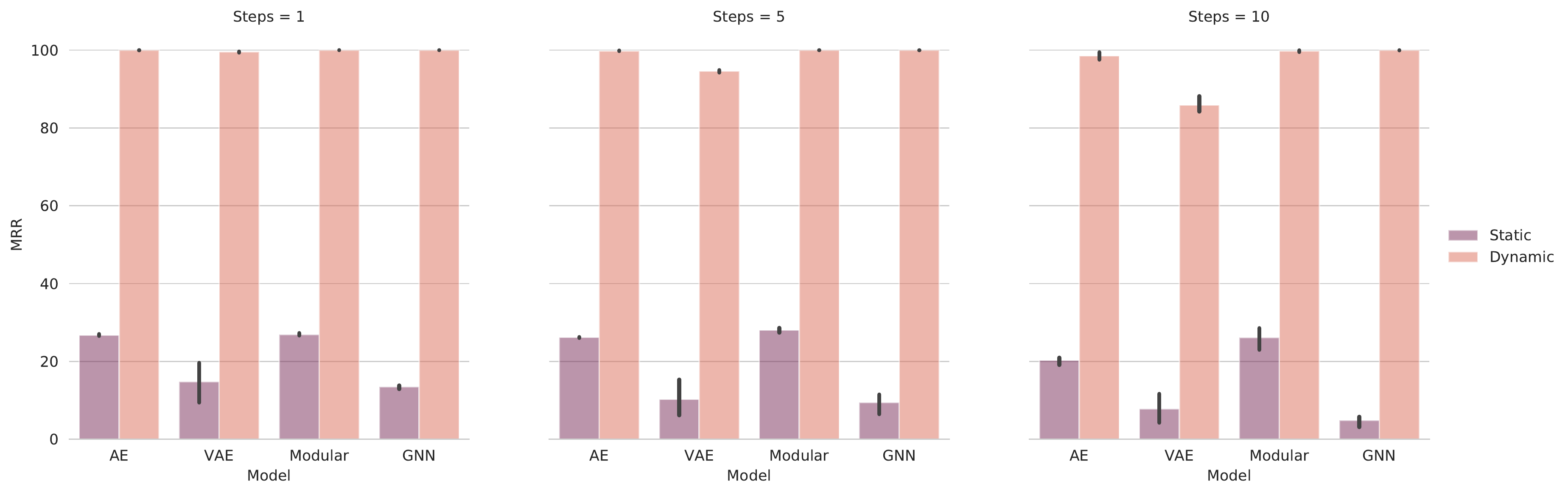}
    \includegraphics[width=15cm]{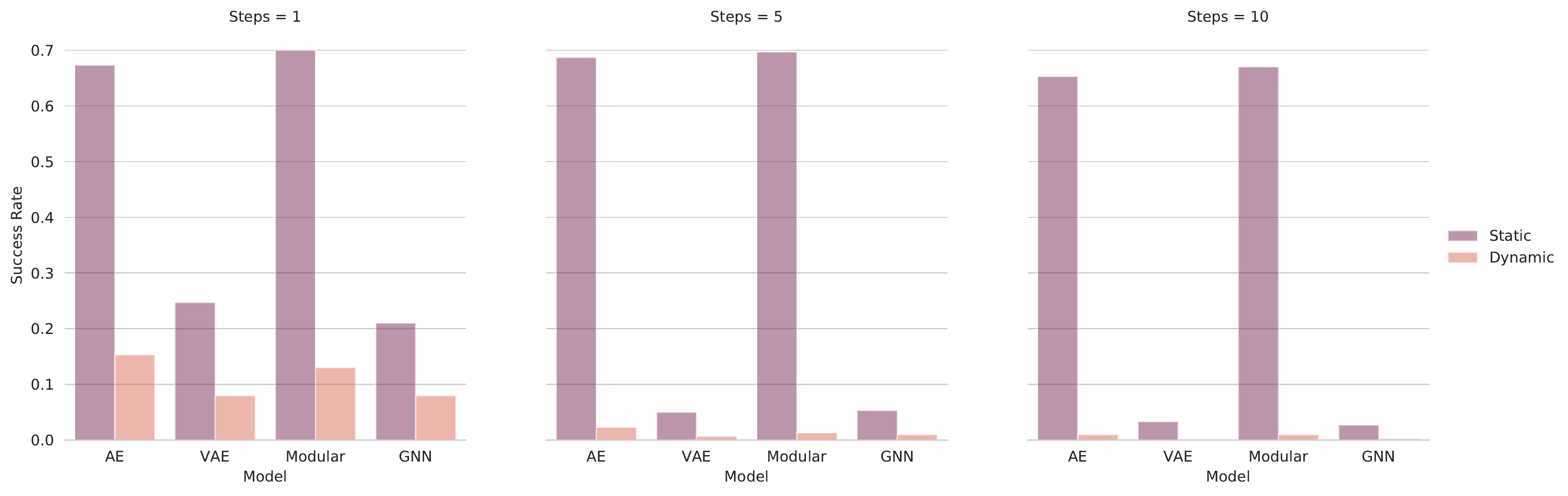}
    \caption{This figure compares the performance of \textit{static} and \textit{dynamic} setting of the chemistry environment. We can see that for the dynamic setting even though the models  achieve almost perfect performance on the ranking losses(H@1 and MRR) as compared to the static setting, their performance on the RL task is extremely low as compared to the static setting. This shows that the ranking losses are not an accurate indicator for model performance. For a description of static and dynamic setting see \Cref{appendix:chemistry_static_dynamic}. These experiments were run for collider graph.}
    \label{fig:static_dynamic}
\end{figure}

We perform ablation studies on the chemistry environment with varying factors in the underlying causal graph to study how these factors impact learning. We summarize our findings below -
\begin{itemize}
    \item It is easier for models to learn the right causal structure when the cause-effect chains are short. For eg., all models perform much better (under all metrics) on the \textit{collider} graph where cause-effect length can be at-most one as opposed to chain and full graph where the cause-effect length is longer (refer to \Cref{fig:chemistry_nll_finetune} and \Cref{tab:chem_nll_finetune_appendix})
    \item \textit{Modular Models} generally perform better than \textit{Graph Neural Networks (GNNs)} when trained using NLL loss because the former can encode \textit{higher-order interactions} while the latter only encodes \textit{pairwise interactions} (refer to \Cref{fig:chemistry_nll_finetune} and \Cref{tab:chem_nll_finetune_appendix}).
    \item While models trained on the \textit{dynamic} chemistry environment perform very well on ranking metrics, they don't do well on the downstream RL task. This is because these models don't actually learn the right causal structure but only encode the visual aspects of the particular episode such as shapes and positions. To further investigate this, we decided to keep the objects \textit{stationary}. We saw that the ranking metrics immediately suffer by a large margin because the models couldn't cheat by just encoding the visual details and not the causal structure (refer to \Cref{appendix:chemistry_static_dynamic} and \Cref{fig:static_dynamic} for details). 
    \item \textit{Increased stochasticity (entropy) }of the conditional probability tables (CPTs) make it harder for the models to learn (refer to \Cref{fig:skew_demo}). In the figure, we can see that almost all models generally perform better on less stochastic (more skewed) data as compared to more stochastic (less skewed) data. 
    \item Modular models outperform all other models on the downstream RL task (refer to \Cref{fig:chem_reward} and \Cref{tab:chem_nll_reward}) for all settings(i.e different graphs and number of steps) due to their ability to encode \textit{higher-order interaction} which monolithic models like AEs and VAEs cannot do while Graph Neural Networks(GNNs) only en \textit{pairwise interactions}. We also report 2 baselines \textit{random} and \textit{optimal} as described in \Cref{appendix:reward_prediction_implementation}
    
\end{itemize}

\begin{table}[hbt!]
\scriptsize	
\centering 
\renewcommand{\arraystretch}{1.2}
\setlength{\tabcolsep}{1pt}
{\begin{tabular}{c|c|ccc|ccc|ccc}
\toprule
& & \multicolumn{3}{c}{\underline{1 Step}} & \multicolumn{3}{c}{\underline{5 Steps}} & \multicolumn{3}{c}{\underline{10 Steps}} \\
Graph Type & Model & H@1 & MRR & Rec. & H@1 & MRR & Rec. & H@1 & MRR & Rec. \\
\hline
\multirowcell{3}{\textbf{Chain} \\ \textbf{NLL}} 
&AE&\highlight{\g{16.937}{0.386}}&\g{23.007}{0.133}&\g{0.07}{0.0}&\g{4.433}{0.023}&\g{8.187}{0.118}&\g{0.073}{0.0}&\g{1.48}{0.04}&\g{2.957}{0.063}&\g{0.076}{0.0} \\
&VAE&\g{10.293}{2.711}&\g{15.897}{2.927}&\g{0.071}{0.0}&\g{2.987}{0.282}&\g{6.983}{1.079}&\g{0.075}{0.0}&\highlight{\g{2.19}{0.184}}&\highlight{\g{5.78}{0.821}}&\g{0.076}{0.0} \\
&Modular&\g{16.863}{0.135}&\highlight{\g{23.047}{0.027}}&\g{0.07}{0.0}&\highlight{\g{5.317}{0.249}}&\highlight{\g{10.31}{1.343}}&\g{0.072}{0.0}&\g{2.04}{0.259}&\g{4.45}{1.043}&\g{0.074}{0.0} \\
&GNN&\g{3.587}{0.412}&\g{6.93}{0.91}&\g{0.07}{0.0}&\g{0.617}{0.05}&\g{1.9}{0.195}&\g{0.076}{0.0}&\g{0.257}{0.002}&\g{0.947}{0.023}&\g{0.079}{0.0} \\
\hline

\multirowcell{3}{\textbf{Full} \\ \textbf{NLL}} 
&AE&\highlight{\g{17.62}{0.192}}&\highlight{\g{23.85}{0.065}}&\g{0.071}{0.0}&\g{5.127}{0.058}&\g{9.707}{0.184}&\g{0.072}{0.0}&\g{2.527}{0.045}&\g{4.913}{0.177}&\g{0.073}{0.0} \\
&VAE&\g{9.847}{0.572}&\g{15.407}{0.559}&\g{0.071}{0.0}&\g{2.747}{0.104}&\g{6.363}{0.342}&\g{0.074}{0.0}&\g{1.957}{0.056}&\g{4.927}{0.289}&\g{0.076}{0.0} \\
&Modular&\g{15.977}{1.066}&\g{22.813}{0.374}&\g{0.071}{0.0}&\highlight{\g{6.493}{0.209}}&\highlight{\g{12.837}{0.62}}&\g{0.071}{0.0}&\highlight{\g{4.233}{0.848}}&\highlight{\g{9.157}{2.529}}&\g{0.071}{0.0} \\ 
&GNN&\g{2.68}{0.073}&\g{5.15}{0.069}&\g{0.071}{0.0}&\g{0.23}{0.001}&\g{0.913}{0.001}&\g{0.077}{0.0}&\g{0.103}{0.001}&\g{0.503}{0.002}&\g{0.084}{0.0} \\

\hline
\multirowcell{3}{\textbf{Collider} \\ \textbf{NLL}} 
&AE&\highlight{\g{20.993}{0.016}}&\highlight{\g{29.723}{0.014}}&\g{0.072}{0.0}&\g{14.84}{0.09}&\g{29.32}{0.135}&\g{0.069}{0.0}&\g{15.01}{0.829}&\g{29.657}{2.029}&\g{0.067}{0.0} \\
&VAE&\g{9.847}{0.572}&\g{15.407}{0.559}&\g{0.071}{0.0}&\g{2.747}{0.104}&\g{6.363}{0.342}&\g{0.074}{0.0}&\g{1.957}{0.056}&\g{4.927}{0.289}&\g{0.076}{0.0} \\
&Modular&\g{20.89}{0.16}&\g{29.563}{0.173}&\g{0.072}{0.0}&\highlight{\g{15.297}{0.063}}&\highlight{\g{29.99}{0.062}}&\g{0.068}{0.0}&\highlight{\g{15.78}{0.47}}&\highlight{\g{31.21}{0.515}}&\g{0.067}{0.0} \\
&GNN&\g{8.377}{2.358}&\g{15.737}{4.398}&\g{0.072}{0.0}&\g{5.443}{2.729}&\g{14.527}{15.714}&\g{0.073}{0.0}&\g{4.04}{3.073}&\g{10.607}{20.141}&\g{0.08}{0.0} \\
\hline
\end{tabular}
}
\caption{Hits at Rank 1 (H@1), Mean Reciprocal Rank (MRR) \textit{(higher is better)} and Reconstruction Error \textit{(lower is better)} for different models trained using \textbf{NLL loss} for 1, 5 and 10 step prediction for the vanilla chemistry environment with 5 objects and 5 colors. }
\label{tab:chem_nll_finetune_appendix}

\end{table}

\begin{table}[hbt!]
\scriptsize	
\centering 
\renewcommand{\arraystretch}{1.2}
\setlength{\tabcolsep}{2pt}
{\begin{tabular}{c|c|cc|cc|cc}
\toprule
& & \multicolumn{2}{c}{\underline{1 Step}} & \multicolumn{2}{c}{\underline{5 Steps}} & \multicolumn{2}{c}{\underline{10 Steps}} \\
Graph Type & Model & Mean Reward  & Success & Mean Reward & Success & Mean Reward  & Success \\
\hline
\multirowcell{3}{\textbf{Chain}} 
&Random&0.56&      0.046& 0.38      &0.005&0.36      0.007\\
&Optimal&0.86      &0.52&0.83     &0.39      &0.16&0.38 \\
&AE&\g{0.81}{0.001}&\g{0.37}{0.003}&\g{0.75}{0.003}&\g{0.26}{0.01}&\g{0.717}{0.004}&\g{0.227}{0.009} \\
&VAE&\g{0.74}{0.003}&\g{0.213}{0.002}&\g{0.583}{0.005}&\g{0.09}{0.003}&\g{0.557}{0.006}&\g{0.073}{0.003} \\
&Modular&\highlight{\g{0.82}{0.001}}&\highlight{\g{0.38}{0.002}}&\highlight{\g{0.763}{0.002}}&\highlight{\g{0.283}{0.011}}&\highlight{\g{0.743}{0.003}}&\highlight{\g{0.237}{0.007}} \\
&GNN&\g{0.673}{0.0}&\g{0.123}{0.0}&\g{0.6}{0.001}&\g{0.12}{0.0}&\g{0.563}{0.001}&\g{0.1}{0.0} \\

\hline
\multirowcell{3}{\textbf{Full}}
&Random&0.45 & 0.027 & 0.27    & 0.005 & 0.25    & 0.004 \\
&Optimal&0.79     &0.44 & 0.737   &0.275&0.72    & 0.24 \\
&AE&\g{0.8}{0.0}&\g{0.41}{0.001}&\g{0.773}{0.002}&\g{0.28}{0.007}&\g{0.747}{0.003}&\g{0.243}{0.006} \\
&VAE&\g{0.707}{0.001}&\g{0.237}{0.002}&\g{0.55}{0.001}&\g{0.067}{0.0}&\g{0.523}{0.001}&\g{0.053}{0.0}\\
&Modular&\highlight{\g{0.82}{0.0}}&\highlight{\g{0.447}{0.003}}&\highlight{\g{0.807}{0.002}}&\highlight{\g{0.337}{0.006}}&\highlight{\g{0.78}{0.002}}&\highlight{\g{0.287}{0.006}} \\
&GNN&\g{0.663}{0.0}&\g{0.177}{0.0}&\g{0.457}{0.0}&\g{0.03}{0.0}&\g{0.39}{0.0}&\g{0.02}{0.0} \\

\hline

\multirowcell{3}{\textbf{Collider}}
&Random&0.45&0.23  &0.27 & 0.005 &  0.25 &  0.004 \\
&Optimal&0.95&0.75&0.94&0.733& 0.96&  0.80 \\
&AE&\g{0.9}{0.002}&\g{0.587}{0.019}&\g{0.86}{0.007}&\g{0.513}{0.075}&\g{0.833}{0.011}&\g{0.477}{0.094} \\
&VAE&\g{0.747}{0.004}&\g{0.2}{0.007}&\g{0.543}{0.006}&\g{0.043}{0.001}&\g{0.45}{0.011}&\g{0.02}{0.0} \\
&Modular&\highlight{\g{0.93}{0.002}}&\highlight{\g{0.69}{0.018}}&\highlight{\g{0.91}{0.007}}&\highlight{\g{0.693}{0.077}}&\highlight{\g{0.907}{0.008}}&\highlight{\g{0.697}{0.075}} \\
&GNN&\g{0.887}{0.001}&\g{0.513}{0.011}&\g{0.827}{0.006}&\g{0.39}{0.032}&\g{0.807}{0.007}&\g{0.35}{0.028} \\

\hline

\hline
\end{tabular}
}
\caption{Mean reward and Success rate (\textit{higher is better}) for 1, 5 and 10 step for the vanilla setting of the chemistry environment with 5 objects and 5 colors. This table uses models trained using \textbf{NLL loss}. }
\label{tab:chem_nll_reward}
\end{table}

\begin{figure}
    \centering
    \includegraphics[width=15cm]{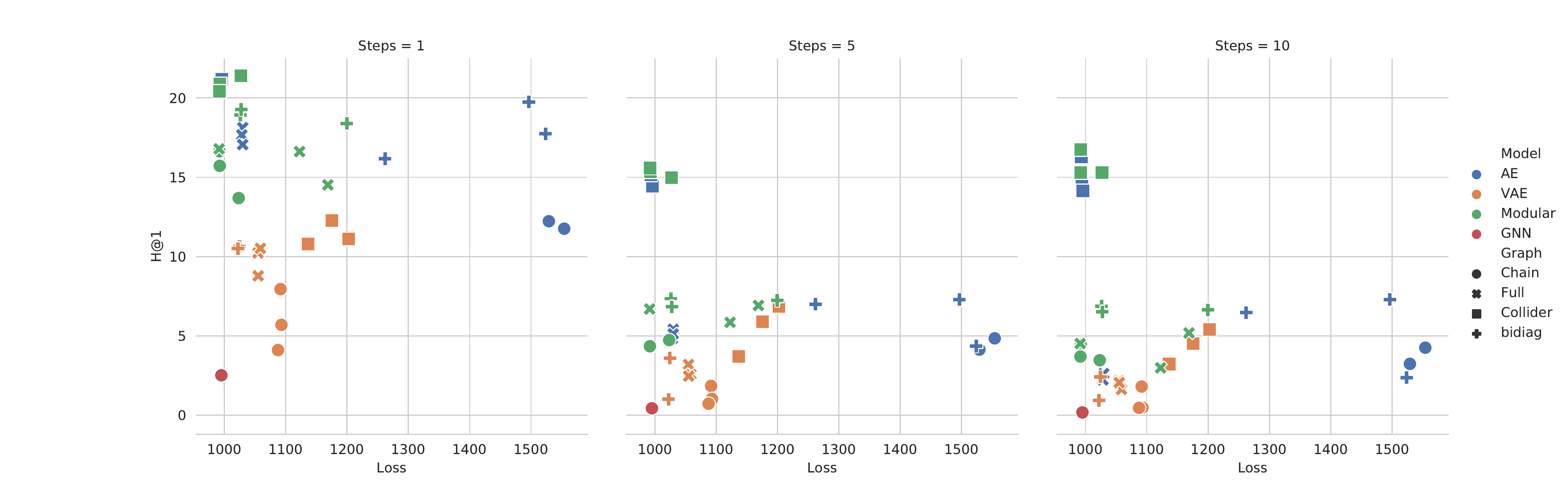}
    \includegraphics[width=15cm]{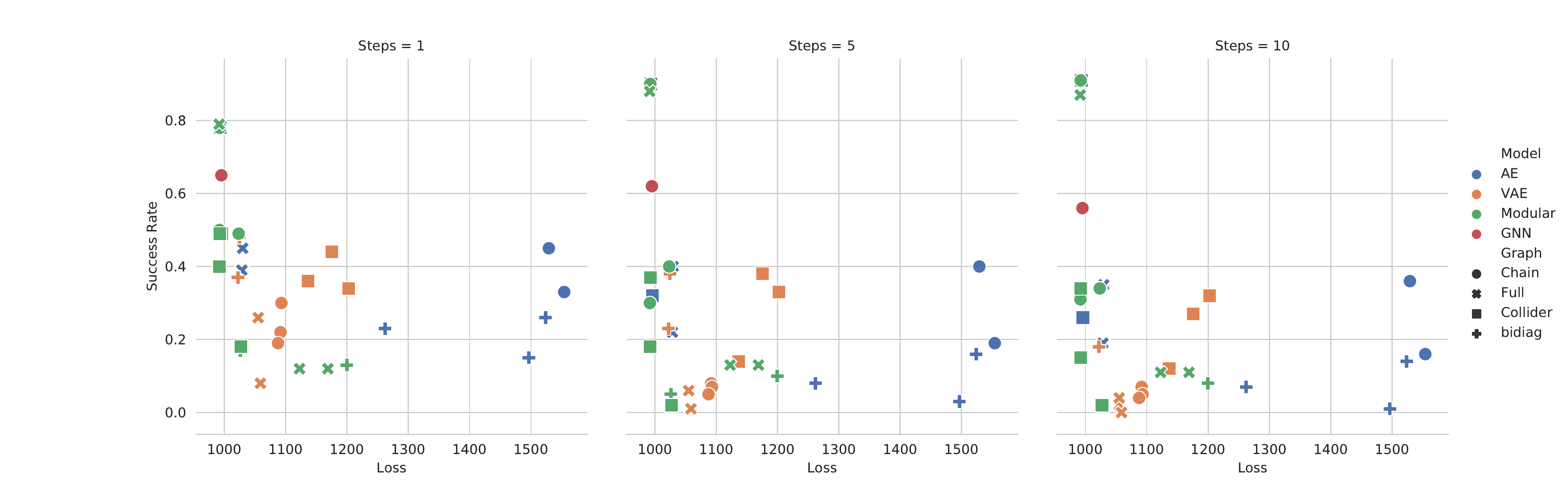}
    \includegraphics[width=15cm]{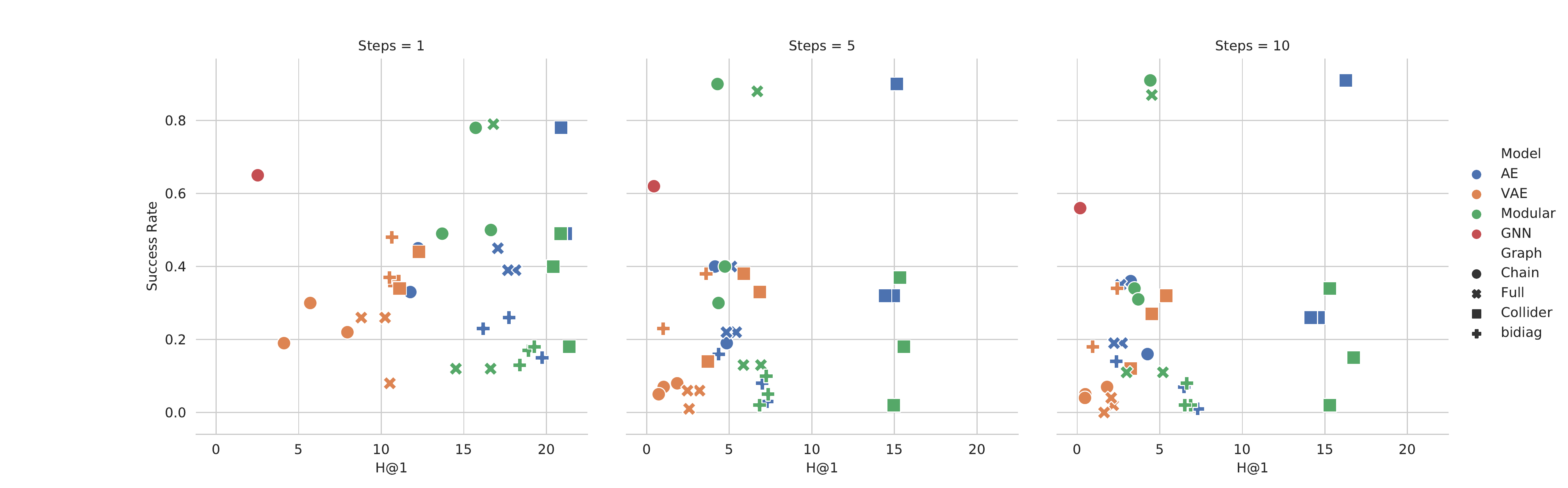}
    \caption{Plots for chemistry environment with 5 objects and  5 colors for models trained using NLL Loss. We see that there seems to be a positive correlation between H@1 and success rate for step 1 but this may not be true for longer steps. }
    \label{fig:chemistry_scatter_5}
\end{figure}

\end{document}